\documentclass[accepted]{uai2023} %
\usepackage[american]{babel}

\usepackage{natbib} %
\usepackage{mathtools} %
\usepackage{booktabs} %
\usepackage{tikz} %
\usepackage{multirow, booktabs} \usepackage{lipsum}
\usepackage{subcaption}
\usepackage[ruled,vlined]{algorithm2e} \usepackage{hyperref}
\usepackage{multirow}
\usepackage[acronym]{glossaries}
\newacronym{elbo}{ELBO}{Evidence Lower Bound}
\newacronym{pdmp}{PDMP}{Piecewise Deterministic Markov Process}
\newacronym{bps}{BPS}{Bouncy Particle Sampler}
\newacronym{sbps}{SBPS}{Stochastic Bouncy Particle Sampler}
\newacronym{esbps}{eSBPS}{Efficient Stochastic Bouncy Particle Sampler}
\newacronym{atsbps}{atSBPS}{Adaptive Thinning Stochastic Bouncy Particle Sampler}
\newacronym{ipp}{IPP}{Inhomogeneous Poisson Process}
\newacronym{vi}{VI}{Variational Inference}
\newacronym{hmc}{HMC}{Hamiltonian Monte Carlo}
\newacronym{sghmc}{SGHMC}{Stochastic Gradient Hamiltonian Monte Carlo}
\newacronym{bnn}{BNN}{Bayesian Neural Network}
\newacronym{cnn}{CNN}{Convolutional Neural Network}
\newacronym{sgld}{SGLD}{Stochastic Gradient Langevin Dynamics}
\newacronym{ood}{OOD}{Out of Distribution}
\usepackage{amsfonts}
\newcommand{\norm}[1]{\left\lVert#1\right\rVert}
 \newcommand{\taur}[0]{\tau_{\text{ref}}}
 
\newcommand{\myv}[0]{\mathbf{v}}
\newcommand{\myw}[0]{\mathbf{\omega}}
\newcommand{\sigbps}[0]{$\sigma$BPS }
\usepackage{tabularx}
\newcolumntype{L}[1]{>{\raggedright\let\newline\\\arraybackslash\hspace{0pt}}m{#1}}
\newcolumntype{C}[1]{>{\centering\let\newline\\\arraybackslash\hspace{0pt}}m{#1}}
\newcolumntype{R}[1]{>{\raggedleft\let\newline\\\arraybackslash\hspace{0pt}}m{#1}}

\newlength{\Oldarrayrulewidth}

\usepackage{float}

\title{Piecewise Deterministic Markov Processes for Bayesian Neural Networks}

\author[1]{\href{mailto:<ej.goan@qut.edu.au>?Subject=UAI2023 BNN Paper}{Ethan~Goan}{}}
\author[1]{Dimitri~Perrin}
\author[1]{Kerrie Mengersen}
\author[1]{Clinton Fookes}
\affil[1]{%
  Queensland University of Technolgy
}

\begin{document}
\maketitle

\begin{abstract}
  Inference on modern Bayesian Neural Networks (BNNs) often relies on a
  variational inference treatment, imposing violated assumptions of independence
  and the form of the posterior. Traditional MCMC
  approaches avoid these assumptions at the cost of increased computation due to
  its incompatibility to subsampling of the likelihood. New Piecewise Deterministic
  Markov Process (PDMP) samplers permit subsampling, though introduce a model-specific inhomogenous Poisson Process (IPPs) which is difficult to sample
  from. This work introduces a new generic and adaptive thinning scheme for approximate
  sampling from these IPPs, and demonstrates how this approach can
  accelerate the application of PDMPs for inference in BNNs. Experimentation
  illustrates how inference with these methods is computationally feasible, can
  improve predictive accuracy, MCMC mixing performance, and provide informative uncertainty measurements when
  compared against other approximate inference schemes.
\end{abstract}

\section{Introduction}
Since \Gls{hmc} was first developed for Bayesian inference
\cite{neal2012bayesian}, sampling methods have seen relatively little
application to \Glspl{bnn}. Flexibility, inference diagnostics and asymptotic
guarantees of HMC comes at the cost of computational complexity as each data
point needs to be used to compute the entire likelihood, and to perform Metropolis Hastings corrections. As models and data
sets have grown, this expense has not been offset by the considerable
performance increase in computational hardware. A recent study found that the
fitting of a HMC model for ResNet20 required a computational cost equivalent to
60 million SGD epochs to obtain only 240 samples from three chains \cite{izmailov2021bayesian}.
\par
\begin{figure}[t]
  \centering
  \subfloat[][]{\includegraphics[width=0.5\linewidth]{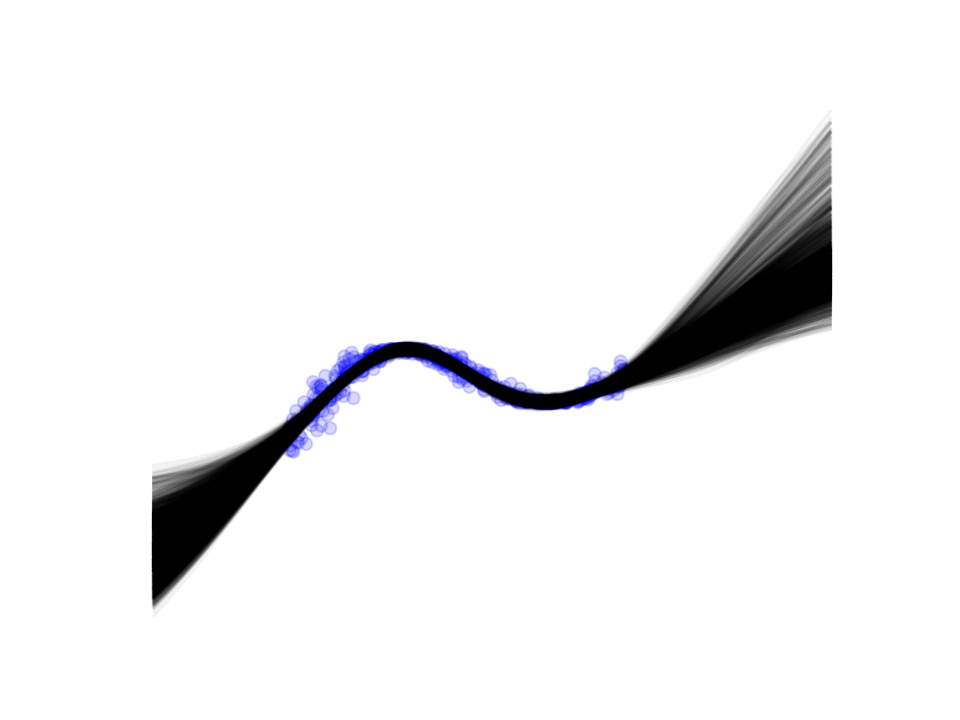}}
  \subfloat[][]{\includegraphics[width=0.5\linewidth]{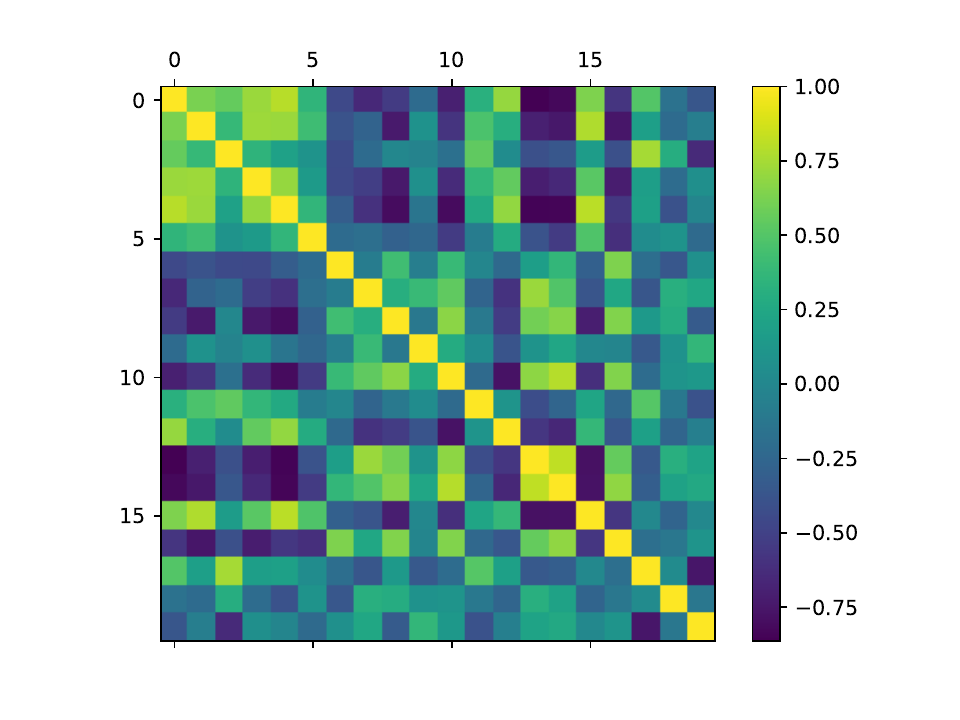}}
  \\
  \subfloat[][]{\includegraphics[width=0.8\linewidth]{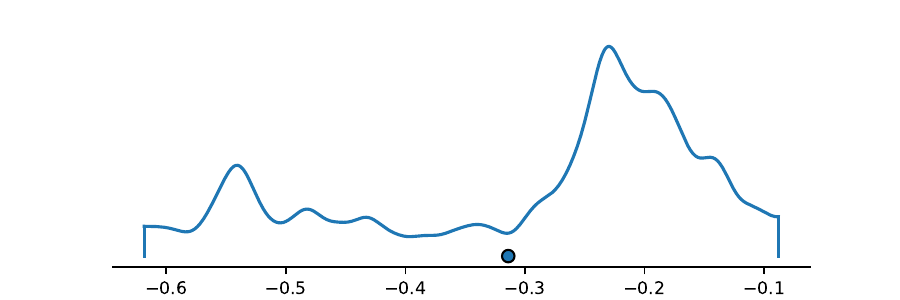}}
  \caption{Example of correlations between the parameters in the first layer of
    a BNN for a simple regression task. Plot (a) samples of predictive posterior
    from proposed method, (b) correlation between all parameters on the first
    layer, (c) kernel density estimate for a single parameter.}
  \label{fig:corr}
\end{figure}
To circumnavigate the computational expense, much research has explored the
application of approximate inference through the lens of \Gls{vi}
\cite{jordan1999introduction, wainwright2008graphical, blei2017variational} or
through exploiting properties of SGD \cite{mandt2017stochastic}. \Gls{vi}
replaces the true target distribution with an approximate distribution that can
be easily manipulated, typically using a mean-field approach
where independence between parameters is assumed. These methods are attractive due to
their reduced computational complexity and their amenability to stochastic
optimisation. However, the suitability of these methods relies heavily on the
expressiveness of the approximate posterior to accurately model the true
distribution. Given the known correlations and frequent multi-modal structure
amongst parameters within BNNs \cite{barber1998ensemble, mackay1995probable}, a
mean-field approximation can be unsuitable for accurate inference. Figure
\ref{fig:corr} illustrates these properties for a simple BNN. Stochastic gradient MCMC methods such as
\Gls{sgld} aim to address this issue but requires prohibitively small and decreasing learning
rates to target the posterior that limits their applicability~\cite{nagapetyan2017true}.
\par
This work explores a new set of ``exact'' inference methods based on
\Glspl{pdmp} \cite{davis1984piecewise} to perform Bayesian inference. \Gls{pdmp} methods can maintain the
true posterior as its invariant distribution during inference whilst permitting
sub-sampling of the likelihood at each update. This property is attractive for
\Glspl{bnn} which typically are of large dimension in terms of parameters and
data sets. Furthermore, previous research has highlighted PDMP methods for favourable
performance in terms of mixing and sampling efficiency
\cite{bouchard2018bouncy, bierkens2019zig, wu2017generalized,
  bierkens2020boomerang}. The dynamics of these samplers are simple to simulate, though simulating the times
to update these dynamics is controlled by an \Gls{ipp} which can be difficult
to sample. This work explores an adaptive procedure to approximate
sampling of these event times to allow for approximate inference within the
context of \Glspl{bnn}. The contributions of this paper are the following,
\begin{itemize}
  \item Propose a novel adaptive thinning method for approximate sampling from
    \Glspl{ipp} events
  \item Develop a GPU-accelerated package for applying these methods to
    general models
  \item Evaluate the performance of these methods applied to computer vision
    tasks using \Glspl{bnn};
  \item Evaluate the suitability of \Gls{pdmp} samplers for
    \Glspl{bnn} and investigate how they can improve predictive accuracy, calibration and
    posterior exploration when compared against \Gls{sgld}.
\end{itemize}
MCMC
methods have often been seen as computationally prohibitive for models with
many parameters or where modern large data sets are used. It is hoped
that this work will demonstrate that approximate inference using MCMC
approaches for BNNs can be practically feasible, offer insightful
results, and to show how we can leverage exact methods for approximate
inference to more accurately target posterior distributions in \Glspl{bnn}.
\section{Preliminaries}
\label{sec:preliminaries}
Following the description from \cite{fearnhead2018piecewise}, \Gls{pdmp} are
defined by three key components: piecewise deterministic dynamics, an event rate,
and the transition kernel. For inference, the goal is to design these three key
components such that we can use the properties of a \Gls{pdmp} to sample from the
posterior distributions of our parameters $\mathbf{\omega}$. We represent the
deterministic dynamics as $\Psi(\myw, \myv, t)$, where $\myv$ is an auxiliary
velocity variable to guide posterior exploration with known distribution $\Phi(\myv)$ and $t$ represents time. At random events,
these dynamics are updated in accordance to a specified transition kernel.
Upon an update event, the piecewise deterministic dynamics of the system update according to the kernel,
and the state $\myw$ at the time of the update event serves as the starting
position for the next segment such that they are all connected.
\par

An \Gls{ipp} with rate function $\lambda(\mathbf{\omega}(t), \mathbf{v}(t))$ governs the
update times for the dynamics. All rate functions in this work rely upon the
negative joint log probability of the model,
\begin{equation}
  \label{eq:energy}
  U(\omega) = -\log \Big(p(\omega) p(\mathcal{D}|\omega)\Big),
\end{equation}
where $p(\omega)$ is a prior or reference measure and $p(\mathcal{D}|\omega)$ is our likelihood,
If these three components are suitably defined,
these processes can sample from a given posterior distribution. For derivations on how
to design these components to target a posterior distribution, the reader can
refer to \cite{fearnhead2018piecewise, vanetti2017piecewise, davis1993markov}. We now introduce the
samplers used within this work.
\par
\subsection{Bouncy Particle Sampler}
\label{sec:bps}
The dynamics of the \Gls{bps} \cite{bouchard2018bouncy} are given
by $\Psi(\myw, \myv, t) = \myw^{i} + \myv^{i} t$, where the superscripts indicate a
deterministic segment. The velocity remains constant within these segments and the
parameter space is explored linearly. The velocity is updated at event times
given by $ \tau \sim \text{IPP}( \lambda(\mathbf{\omega}(t)), \mathbf{v})$, where,
\begin{equation}
  \label{eq:bps_rate}
  \lambda(\mathbf{\omega}(t), \mathbf{v}) = \max \{0, \nabla U(\mathbf{\omega}) \cdot \mathbf{v}^{i} \}.
\end{equation}
Once an event time is sampled, the state of our variable ``bounces'' according
to a lossless inelastic Newtonian collision,
\begin{equation}
  \label{eq:bps_bounce}
  \mathbf{v}^{i+1}= \mathbf{v}^{i} - 2 \dfrac{ \nabla U(\mathbf{\omega}^{i+1}) \cdot \mathbf{v}^{i}}{ \norm{\nabla U(\mathbf{\omega}^{i+1}) }^{2}} \nabla U(\mathbf{\omega}^{i+1})
\end{equation}
where $\myw^{i+1}$ represents the end of the previous segment at time $\tau$, and serves as the starting position for the following segment.
The \Gls{bps} provides linear dynamics that are simple to simulate,
though relies only on local gradient information, which can lead to inefficient
exploration for \Glspl{bnn}. Preconditioning can allow us to address this.
\subsection{Preconditioned BPS}
To accelerate posterior exploration in directions of interest, we can
precondition the gradients to include more information about the structure of our
posterior space. Introduction of a
preconditioning matrix $A$ results in new dynamics of
$\Psi(\myw, \myv, t) = \myw^{i} + A\myv^{i} t$, and a new event rate,  $\lambda(\mathbf{\omega}(t), \mathbf{v}) = \max\{0, \mathbf{v} \cdot A \nabla U(\mathbf{\omega} + \mathbf{v}t) \}$.
Upon events, the velocity is updated according to,
\begin{equation}
  \label{eq:bps_bounce}
  \mathbf{v}^{i+1}= \mathbf{v}^{i} - 2 \dfrac{A \nabla U(\mathbf{\omega}^{i+1}) \cdot \mathbf{v}^{i}}{ \norm{A\nabla U(\mathbf{\omega}^{i+1}) }^{2}} A\nabla U(\mathbf{\omega}^{i+1}).
\end{equation}
With careful choice of $A$, exploration along certain axis can be appropriately
scaled. \cite{pakman2017stochastic} propose a preconditioner similar to
\cite{li2015preconditioned}, though our preliminary experimentation found
inconsistent results when applied to \Glspl{bnn}. Instead, we opt to build on the
approach of \cite{bertazzi2020adaptive}, where we use variance information of
our samples to precondition our dynamics. We choose the preconditioner such
that $A =\text{diag}\big(\Sigma^{\frac{1}{2}}\big)$, where $\Sigma$ is the estimated
covariance in our sample found during a warm-up period. As such, we refer to this
sampler as the \sigbps.

\subsection{Boomerang Sampler}
The Boomerang Sampler \cite{bierkens2020boomerang} introduces non-linear
dynamics for both parameter and velocity terms, and the inclusion of a Gaussian
reference measure for the parameters and
velocity $ \mathcal{N} (\myw_{\star}, \Sigma_{\star}) \otimes \mathcal{N}(0, \Sigma_{\star})$. The first term in this reference
measure will appear in the target distribution similar to a  prior in the joint probability over
parameters, and the second as the known distribution for the velocity component.
The parameters $\myw_{\star}$ and $\Sigma_{\star}$ can be specified as traditional prior, or
can be specified in an empirical approach where they are learnt from the data.
Within this work, we will set $\myw_{\star}$ to the MAP estimate. In the original
paper, $\Sigma_{\star}$ is set to the inverse of the Hessian, however, this can be
computationally prohibitive for \Glspl{bnn}. Instead, we approximate the
diagonal of the Hessian matrix as,
\begin{equation}
  \label{eq:boomerang_sigma}
  \Sigma_{\star } = \gamma \Big[ \sum_{i=0}^{N}\frac{\partial}{\partial \myw^{2}} -\log p(\mathbf{y}_{i}|\mathbf{x}_{i}, \myw)\Big] ^{-1}
\end{equation}
where $N$ is the number of mini-batches present, $\mathbf{x}_{i}, \mathbf{y}_{i}$ represents
data from the $i$'th mini-batch, and $\gamma$ is a hyperparameter to
adjust the scale as needed.
\par
Unlike the \Gls{bps} samplers, the velocity does not
remain constant between events. The dynamics of the Boomerang sampler for
$\myw$ and $\myv$ within events are given
by
$\Psi(\myw, \myv, t)_{\myw} = \myw_{\star} + (\myw^{i} - \myw_{\star}) \cos(t) + \myv^{i}\sin (t)$,
$\Psi(\myw, \myv, t)_{\myv}= -(\myw^{i} - \myw_{\star}) \sin (t) + \myv^{i}\cos (t)$,
where the subscripts denote the parameter and velocity trajectory within the
deterministic segment. The event rate is the same as the \Gls{bps}, and the
starting velocity for the next segment is updated upon events as,
\begin{equation}
  \label{eq:bps_bounce}
  \mathbf{v}= \mathbf{v} - 2 \dfrac{\nabla U(\mathbf{\omega}) \cdot \mathbf{v}}{ \norm{\Sigma_{\star}^{\frac{1}{2}}\nabla U(\mathbf{\omega}) }^{2}} \Sigma_{\star}\nabla U(\mathbf{\omega}).
\end{equation}
\subsection{Velocity Refreshment}
\label{sec:refresh}
All of the samplers introduced fail to target the posterior explicitly when using
the above dynamics alone. Introduction of a refreshment step
rectifies this, which is governed by a homogeneous PP  $\tau_{ref} \sim \lambda(\lambda_{ref})$.
When $\tau_{ref} < \tau$, the velocity is instead randomly sampled from the known
reference distribution $\Phi(\myv)$, and $\tau_{ref}$ is used for the update event
time. For BPS samplers in this work, we use a refreshment distribution of the
form $\mathcal{N}(0, \sigma^{2})$, where for the BPS $\sigma$ is a hyper-parameter to be set and for
for the \sigbps is used as the standard deviation learnt during the warmup stage, and the Boomerang
sampler requires $\Phi(\myv) = \mathcal{N}(0, \Sigma_{\star})$. A summary of PDMP algorithms for
inference is described in Algorithm \ref{alg:pdmp}.
\begin{algorithm}
  \SetAlgoLined \KwResult{Samples from posterior distribution}
  \While{Sampling}{
    \tcp{Simulate event time}
    \tcp{event times in this work simulated with Algorithm \ref{alg:ipp}}
    $\tau \sim \text{PP}(\lambda(\myw, \myv))$\;
    \tcp{Simulate time of refresh event}
    $\taur \sim \text{PP}(\lambda_{\text{ref}})$\;
    $\tau^i = \text{min}(\tau, \taur)$\;
    \tcp{find end of current piecewise-deterministic segment, which will form start of next segment}
    $\omega^{i+1}  = \Psi(\myw, \myv, \tau^{i})_{\myw}$\;
    \eIf{$\tau^{i} = \tau$}{ \tcp{update according kernel}
      $\mathbf{v}^{i+1} = R(\mathbf{\omega}^{i+1}, \mathbf{v}^{i})$\; }{
      \tcp{refresh velocity}
      $\mathbf{v}^{i+1} \sim \Phi(\mathbf{v})$; } }
  \caption{Application of PDMP samplers for Inference}
  \label{alg:pdmp}
\end{algorithm}
\subsection{Problems with the event rate}
With the deterministic dynamics illustrated in these samplers, the main challenge in
implementation of these methods is due to the sampling of the event times.
Analytic sampling from $\text{IPP}\big(\lambda(t)\big)$ requires being able to invert
the integral of the event rate w.r.t. time,
\begin{equation}
  \label{eq:event_analytic}
  \Lambda(t) = \int_0^\tau \lambda(t) dt = \int_0^\tau \max\{0, \mathbf{v} \cdot A \nabla U(\mathbf{\omega}(\myv, t) \} dt,
\end{equation}
where $A = \mathbf{I}$ for the \Gls{bps} and Boomerang samplers.
Inverting the above integral is feasible only for simple models. A general case
for sampling from IPPs is available through thinning \cite{lewis1979simulation}.
This requires introducing an additional rate function that we can sample from
$\mu(t)$ that is also a strict upper bound on the event rate function of
interest such that $\mu(t) \geq \lambda(t)$ for all $t \geq 0$.
\par
\begin{figure*}[!htpb]
  \begin{subfigure}[t]{0.25\textwidth}
    \centering \includegraphics[width=1\linewidth]{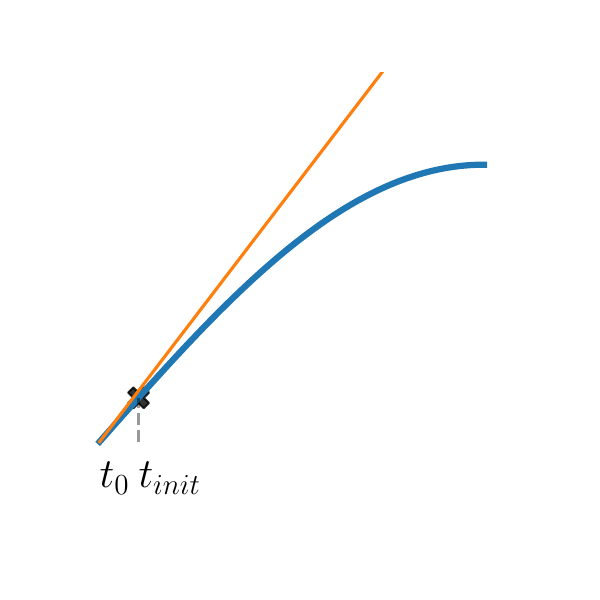}
  \end{subfigure}%
  ~
  \begin{subfigure}[t]{0.25\textwidth}
    \centering \includegraphics[width=1\linewidth]{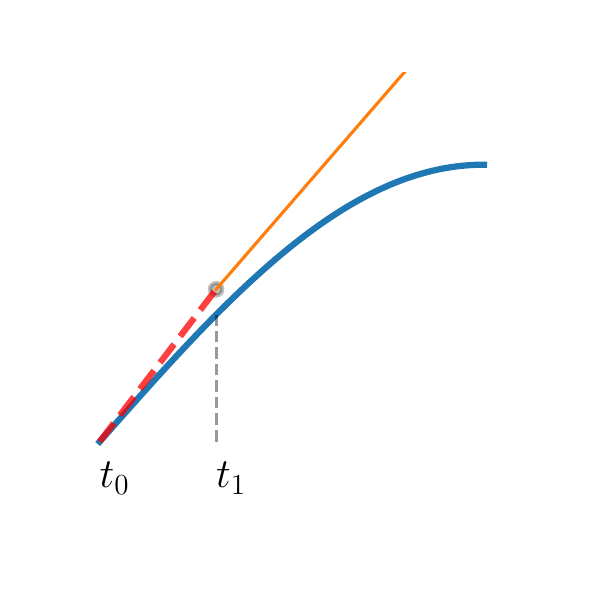}
  \end{subfigure}
  \begin{subfigure}[t]{0.25\textwidth}
    \centering \includegraphics[width=1\linewidth]{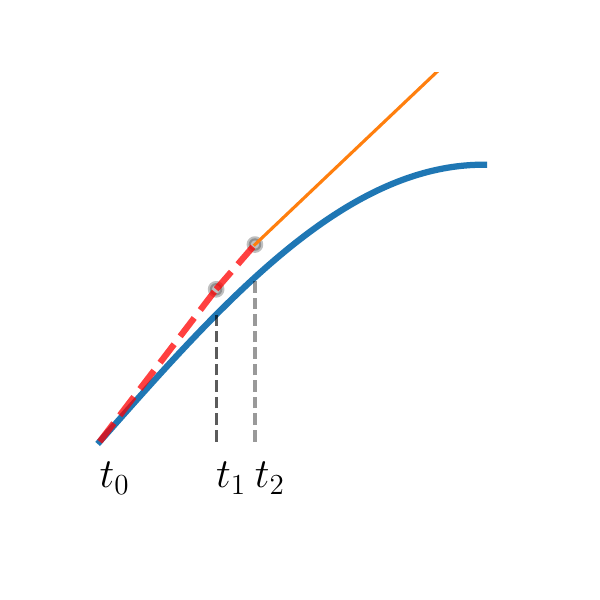}
  \end{subfigure}%
  \begin{subfigure}[t]{0.25\textwidth}
    \centering \includegraphics[width=1\linewidth]{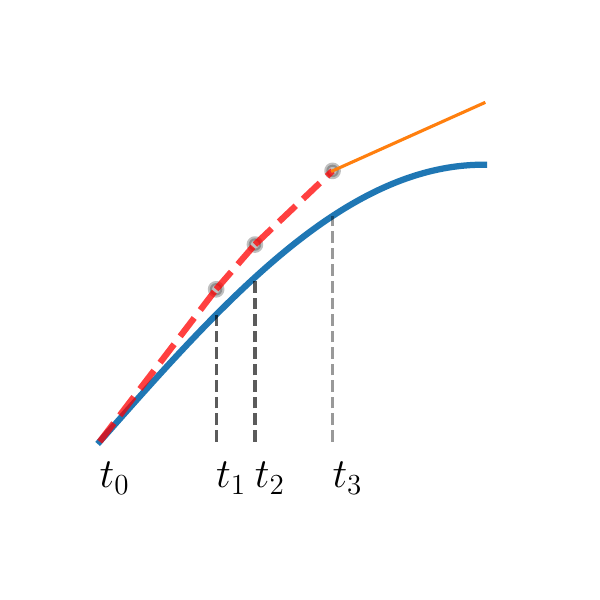}
  \end{subfigure}
  \caption{Example of the progression of the proposed envelope scheme used for
    thinning. The blue line represents the true event rate, orange section
    depicts the active regions for which we sample a new proposal time, and the
    red section depicts previous segments in the envelope. Starting from the
    left, an initial segment is found through interpolation between time
    points $t_{0}$ and $t_{init}$. In the next segment, active regions of the
    envelope are found by interpolating between the two prior points, which
    extends to create a new segment to propose times. This process continues
    until a proposed time is accepted from thinning.}
  \label{fig:envelope}
\end{figure*}
The efficiency of any thinning scheme relies on the tightness of the upper
bound; the greater the difference between the upper bound and the true rate, the
more likely a proposed time will be rejected when sampling.
\cite{pakman2017stochastic} propose a Bayesian linear regression method to
generate an upper bound suitable for thinning, though requires the calculation of
variance within gradients to formulate a suitable upper bound. They calculate
this variance empirically, which requires computing the gradient for each data
point individually within a mini-batch. This computation prohibits use for \Glspl{bnn}
where automatic differentiation
software is used. Furthermore, the selection of a suitable prior distribution for this is
non-trivial for neural networks. In the next
section, we address this issue by instead introducing an interpolation-based
scheme for creating efficient and adaptive approximate upper bounds that avoids excessive gradient
computations and the need to specify a prior on the approximate bound.
\section{Adaptive Bounds for samplers}
\subsection{Sampling from IPPs with Linear Event Rates}
Our goal is to create a piecewise-linear envelope $h(t)$ that will serve as an
approximate upper bound of our true event rate, where each segment in $h(t)$ is
represented by $a_{i}t + b_{i}$. This envelope will serve as the event rate for
a proposal IPP that will be suitable for use with the
thinning method of \cite{lewis1979simulation}. Acceptance of an event time $t$
is given by,
\begin{equation}
  \label{eq:thinning}
  U \le \frac{\lambda(t)}{h(t)},
\end{equation}
where $U \sim \text{Uniform}[0, 1]$. We begin by
building on the work of \cite{klein1984time} to demonstrate how to sample times
from an IPP with a piecewise-linear event rate which we can use with thinning.
\par
Within our proposal \Gls{ipp} with rate $h(t)$, we wish to generate the next event time $t_{i}$
given the previous event $t_{i-1}$. The probability of events occuring
within the range of $[t_{i-1}, t_{i}]$ is given by \cite{devroye2006nonuniform},
\begin{equation}
  \label{eq:2}
  F(x) = 1 - \text{exp}\{-(\Lambda(t_{i}) - \Lambda(t_{i-1}) )\}.
\end{equation}
We can solve this expression for $t_{i}$ by,
\begin{equation}
  \label{eq:3}
  t_{i} = \Lambda^{-1}(\Lambda(t_{i-1}) - \ln U),
\end{equation}
where $U \sim \text{Uniform[0,1]}$. For linear segments, the solution to this
system can be written as \cite{klein1984time},
\begin{align}
  \label{eq:linear_time}
  t_{i} = & -b_{i} / a_{i} \nonumber
  \\
          & + \sqrt{b_{i}^{2} +  a_{i}^{2}t_{i-1}^{2} + 2a_{i}b_{i}t_{i-1} - 2a_{i}log(1 - U)}\big ) /a_{i}.
\end{align}
This provides a framework for sampling from \Glspl{ipp} with a linear event
rate. We now describe how we create a piecewise-linear envelope for a proposal
process that can be used for thinning.
\subsection{Piecewise Interpolation for Event Thinning}
\label{sec:interpolation}
We begin by introducing a modified event rate for which we will form our envelope,
\begin{equation}
  \label{eq:rate_adjust}
  \hat{\lambda}(\mathbf{\omega}(t), \mathbf{v}) = \max \{0, \alpha\nabla U(\mathbf{\omega}) \cdot \mathbf{v}^{i} \},
\end{equation}
where $\alpha \geq 1$ is a positive scaling
factor to control the tightness of the approximate bound on the rate. The use of $\hat{\lambda}$ for
creating our envelope is valid, since for values
of $\alpha \geq 1$, $\hat{\lambda}(t) \geq \lambda(t)$. The scaling factor included in this
event rate is designed to provide flexibility to end users with respect to
computational time and bias that will be introduced during inference. The closer
$\alpha$ is to one, the lower the probability for rejection of proposed event
times, but the greater the probability that the generated event rate will not be
a strict upper bound.
\par
Our goal is to create a piecewise-linear upper bound suitable for proposing
event times using Equation \ref{eq:linear_time}. To achieve this we have two growing sets, one for proposed
event times $T = \{t_{0}, ..., t_{n}\}$ and the value of the adjusted event rates
at these times $L = \{\hat{\lambda}(t_{0}), \dots, \hat{\lambda}(t_{n})\}$ for which we can
create a set of functions,
\begin{equation}
  \label{eq:h(t)}
  h(t) =  a_{i}t + b_{i}, \hspace{0.5cm} t \ge t_{i}.
\end{equation}
The values for $a_{i}$ and $b_{i}$ are found by interpolating between
the points $(t_{i-1}, \hat{\lambda}(t_{i-1}))$ and $(t_{i}, \hat{\lambda}(t_{i}))$.
\par
At the beginning of every deterministic \Gls{pdmp} segment, the sets $T$ and $L$
will be empty. To initialise the sets and create our first linear segment, we
evaluate the event rate at two points, $t_{0}=0$ and $t=t_{init}$,
where $t_{init} > t_{0}$. To evaluate the values for $a_{0}$ and $b_{0}$, we
interpolate between these two segments. Once the values for the first linear
segment are found, $t_{0}$ and $\hat{\lambda}(t_{0})$ are appended to their
corresponding sets, and $t_{init}$ and $\hat{\lambda}(t_{init})$ are discarded. With
this initial linear segment, we can propose a time $t_{i}$ through Equation
\ref{eq:linear_time}. This proposed time is either accepted or rejected from
Equation \ref{eq:thinning}.
\par
If the proposed time is accepted, then the dynamics of the \Gls{pdmp} sampler
are updated at the given event time and the sets $T$ and $L$ are cleared, ready
to be re-initialised for the new dynamics. If the time is rejected, the proposed
time $t_{i}$ and envelope evaluation $\hat{\lambda}(t_{i})$ are appended to their
respective sets, and a new linear segment is calculated to interpolate between
this rejected proposal and the previous elements in the sets $T$ and $L$. The
rejected proposal time will serve as the new starting point $(t_{i-1})$ for the
new linear segment to propose the next time using Equation \ref{eq:linear_time}.
This will continue until the proposed event time is accepted. This process depicted visually in Figure \ref{fig:envelope}
and summarised in Algorithm \ref{alg:ipp}.
\par
Within this work, we limit ourselves to models where the envelope provided
by $h(t)$ will only be an approximate upper bound, meaning bias will likely be
introduced during inference. Diagnosis and correction of this can be identified
through the acceptance ratio $\lambda(t) / h(t)$; if this value is greater than one,
the condition of $h(t)$ being a local upper bound is violated. The amount of
potential bias introduced can be mitigated by increasing the scaling factor $\alpha$ in Equation \ref{eq:rate_adjust} at the
expense of increasing computation load. We also introduce an additional
rejection condition such that if $\lambda(t) / h(t) > R$, then we will reject the
the proposed time. For this work, $R = 2$. These properties are investigated in Supp.
Material A. In the following sections, we evaluate the proposed event thinning
scheme for \Glspl{bnn} to identify the suitability of different samplers for inference
in these challenging models, and how they can outperform other stochastic approximation
methods in terms of calibration, posterior exploration, sampling efficiency and
predictive performance.
\begin{algorithm}[!h]
  \SetAlgoLined \KwResult{Proposed PDMP Event Time $\tau$}
  Initialize $T, L$\;
  Evaluate $(0, \lambda(0)), (t_{init}, \lambda(t_{init}))$\;
  $i = 1$\;
  Compute $a_{i}, b_{i}$\;
  $T_{0} \leftarrow 0, L_{0} \leftarrow \lambda(t)$\;
  Discard $t_{init}, \lambda(t_{init}) $\;
  \While{not accepted}{
    \tcp{propose event time}
    $t_{i} \sim PP \Big(h(t)$\Big)\;
    $u \sim \text{Uniform}[0, 1]$\;
    $r = \lambda(t_{i}) / h(t)$\;
    \tcp{if valid proposal}
    \If{$u \leq r$ \textbf{and} $r < R$}{
      \tcp{sample is accepted}
      $\tau = t_{i}$\;
      accepted = True\;
    }
    \Else{
      $i += 1$\;
      $T_{i} \leftarrow t_{i}, L_{i} \leftarrow \lambda(t_{i})$\;
      \tcp{update linear segment}
      $a_{i}, b_{i}$ = update($L, T$)\;
    }
  }
  \caption{Sampling event rate using proposed adaptive thinning method.}
  \label{alg:ipp}
\end{algorithm}

\vspace*{1cm}
\section{Related Work}
\label{sec:related}
The samplers used within this work require the use of an additional reference
process to provide velocity refreshments. The Generalised BPS
\cite{wu2017generalized} is an updated variant of the BPS algorithm that
incorporates a stochastic update of the velocity which alleviates the
need for a refreshment process. Simulations have shown comparable performance
to the BPS for simple models and how it can reduce the need for fine-tuning the reference
parameter $\tau_{ref}$.
\par
Another prominent sampler is the Zig-Zag Process (ZZP)
\cite[]{bierkens2019zig}, where at events the dynamics of a single
parameter are updated. For the one-dimensional case, this sampler
represents the same process as the BPS. This sampler has shown
favourable results in terms of mixing performance and can achieve
ergodicity for certain models where the BPS cannot. A key characteristic
of this method is that each parameter is assigned an individual event
rate, making implementation for high-dimensional \Gls{bnn} models challenging.
\par
Another class of algorithms designed for subsampling are discrete
stochastic MCMC methods \cite{wenzel2020good, chen2014stochastic, chen2015complete,
  welling2011bayesian, li2015preconditioned}. These models have shown
favourable performance, with a recent variant achieving comparable
predictive accuracy on the ImageNet data set
\cite{heek2019bayesian}. Compared to algorithms related to PDMPs, it
has been shown that high variance related to naive subsampling limits
these methods to provide only an approximation to the posterior
\cite{betancourt2015fundamental}. The bias that is introduced due to subsampling can be controlled by
reducing the step-size for these methods at the expense of mixing performance
and posterior exploration\cite{nagapetyan2017true,brosse2018promises,teh2016consistency}.
We investigate the effect of this property for \Gls{sgld} and compare performance with \Gls{pdmp}
samplers in the following section.
\section{Experiments}
\label{sec:experiments}
We now validate the performance of \Glspl{pdmp} using the proposed event
sampling method on a number of synthetic and real-world data sets for regression
and classification. To analyse performance for predictive tasks, the predictive
posterior needs to be evaluated. In this work, we discretise samples from the
trajectory to allow for Monte Carlo integration,
\begin{multline}
  \label{eq:discretise}
  p(y^* | x^*, \mathcal{D}) = \int \pi(\mathbf{\omega}) p(y^* | \mathbf{\omega}, x^*) d\mathbf{\omega} \\
  \approx \dfrac{1}{N} \sum_{i=1}^N p(y^* | \mathbf{\omega}_i, x^*)
  \hspace*{0.5cm} \mathbf{\omega}_i \sim \pi(\mathbf{\omega}),
\end{multline}
where parameter samples of $\mathbf{\omega}^{(i)}$ are taken from the values
encountered at event times. Experimentation is first conducted on synthetic data
sets to allow us to easily visualise predictive performance and uncertainty in
our models, followed by more difficult classification tasks with Bayesian
\Glspl{cnn} on real data sets. For all experimentation, we set our scaling
factor from Equation \ref{eq:rate_adjust} to $\alpha=1.0$ to promote computational
efficiency. To enable these experiments, we deliver a Python package titled
Tensorflow PDMP (TPDMP). This package utilises the Tensorflow Probability
library \cite{dillon2017tensorflow}, allowing for hardware acceleration and
graph construction of all our models to accelerate computation. We deliver
kernels to implement the \Gls{bps}, \sigbps, and Boomerang sampler with our
proposed event thinning scheme. Code is available
at
\href{https://github.com/egstatsml/tpdmp.git}{https://github.com/egstatsml/tpdmp.git}.
\begin{figure*}[!hbt]
  \begin{subfigure}[t]{0.18\textwidth}
    \centering
    \includegraphics[width=1\linewidth]{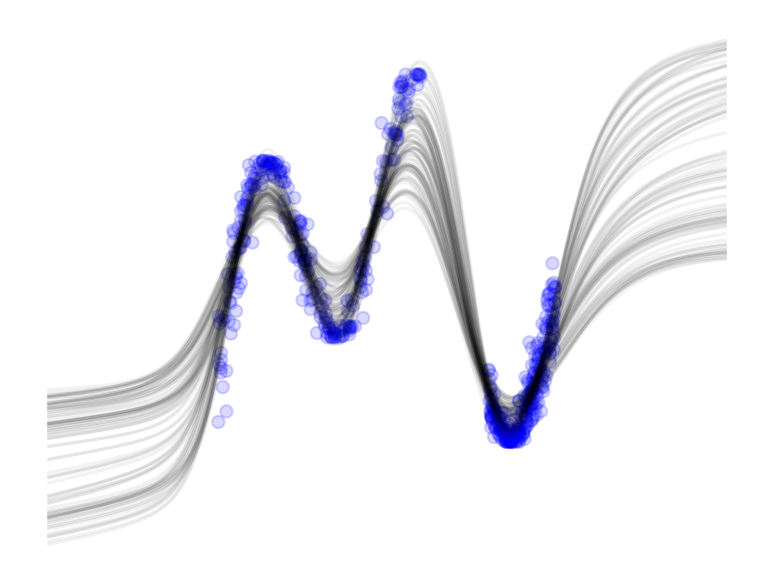}
    \caption{BPS}
  \end{subfigure}%
  ~
  \begin{subfigure}[t]{0.18\textwidth}
    \centering
    \includegraphics[width=1\linewidth]{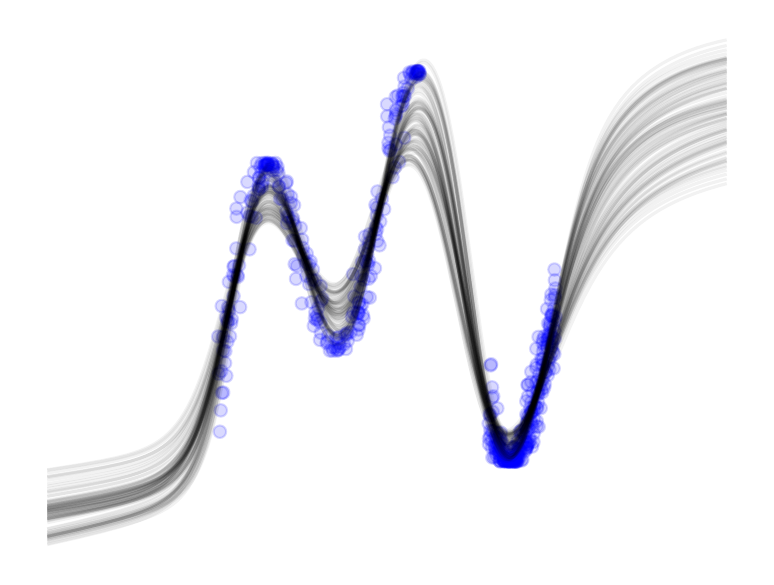}
    \caption{\sigbps}
  \end{subfigure}
  ~
  \begin{subfigure}[t]{0.18\textwidth}
    \centering
    \includegraphics[width=1\linewidth]{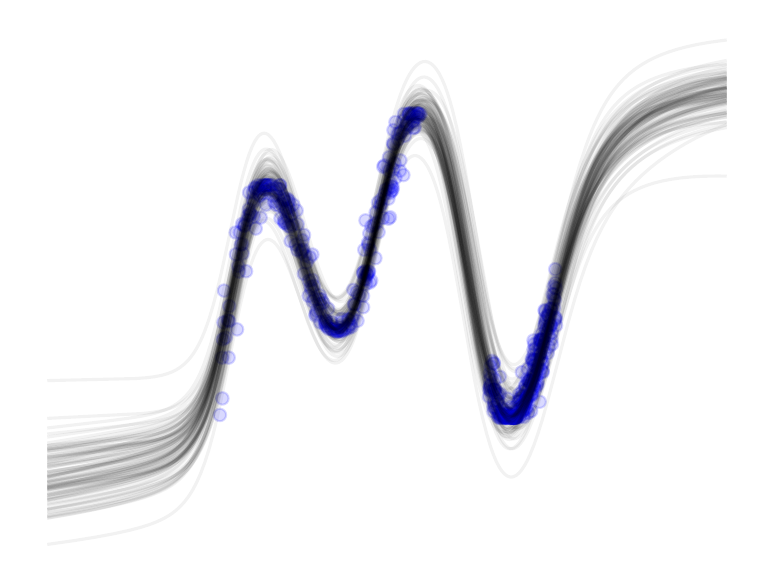}
    \caption{Boomerang}
  \end{subfigure}%
  ~
  \begin{subfigure}[t]{0.18\textwidth}
    \centering
    \includegraphics[width=1\linewidth]{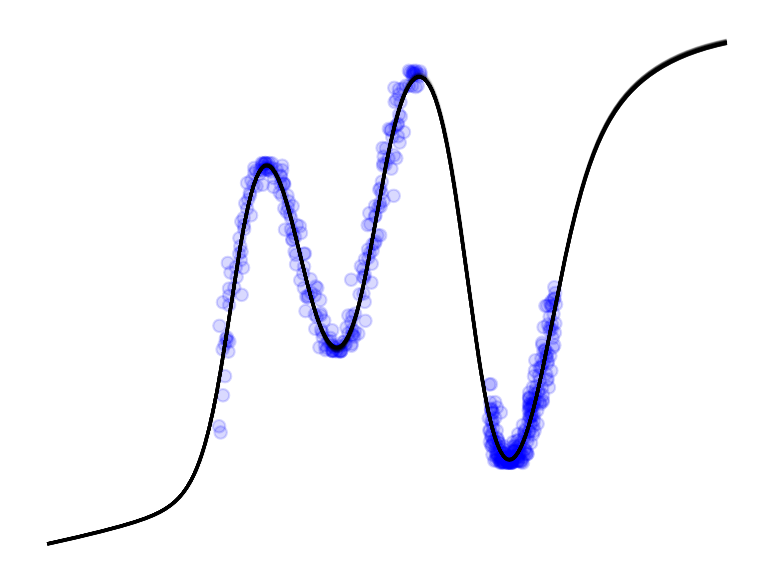}
    \caption{SGLD}
  \end{subfigure}
  ~
  \begin{subfigure}[t]{0.18\textwidth}
    \centering
    \includegraphics[width=1\linewidth]{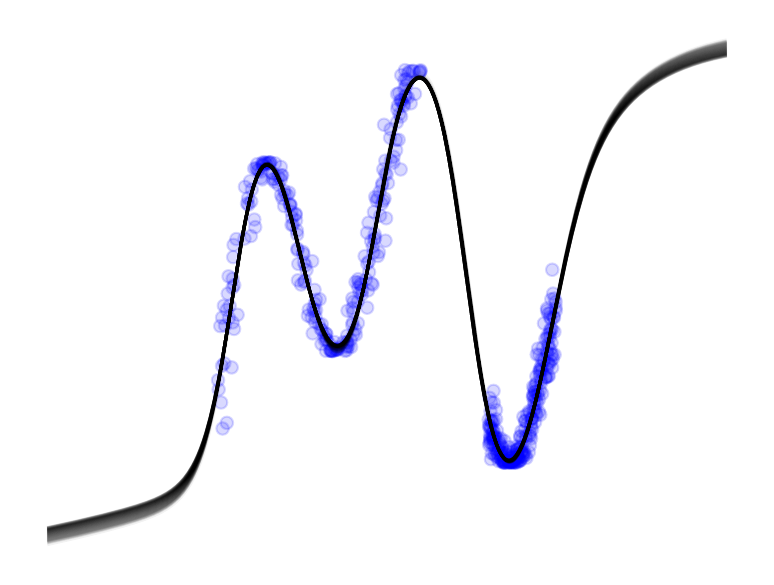}
    \caption{SGLD-ND}
  \end{subfigure}
  \caption{Examples of the different PDMP samplers using the proposed event
    thinning procedure on synthetic regression task compared against SGLD with
    decaying learning rate and constant learning rate (SGLD-ND).}
  \label{fig:regression}
\end{figure*}
\subsection{Regression and Binary Classification with Synthetic Data}
\label{sec:reg_bin}
\begin{figure}[!h]
  \begin{subfigure}[t]{0.25\textwidth}
    \centering
    \includegraphics[width=1\linewidth]{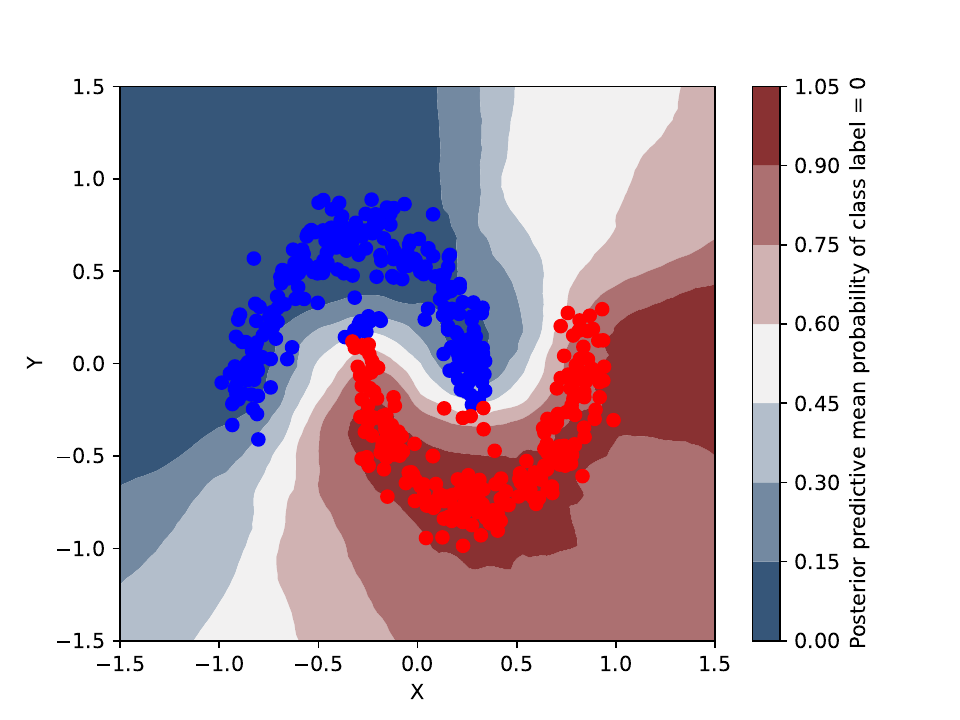}
  \end{subfigure}%
  \begin{subfigure}[t]{0.25\textwidth}
    \centering
    \includegraphics[width=1\linewidth]{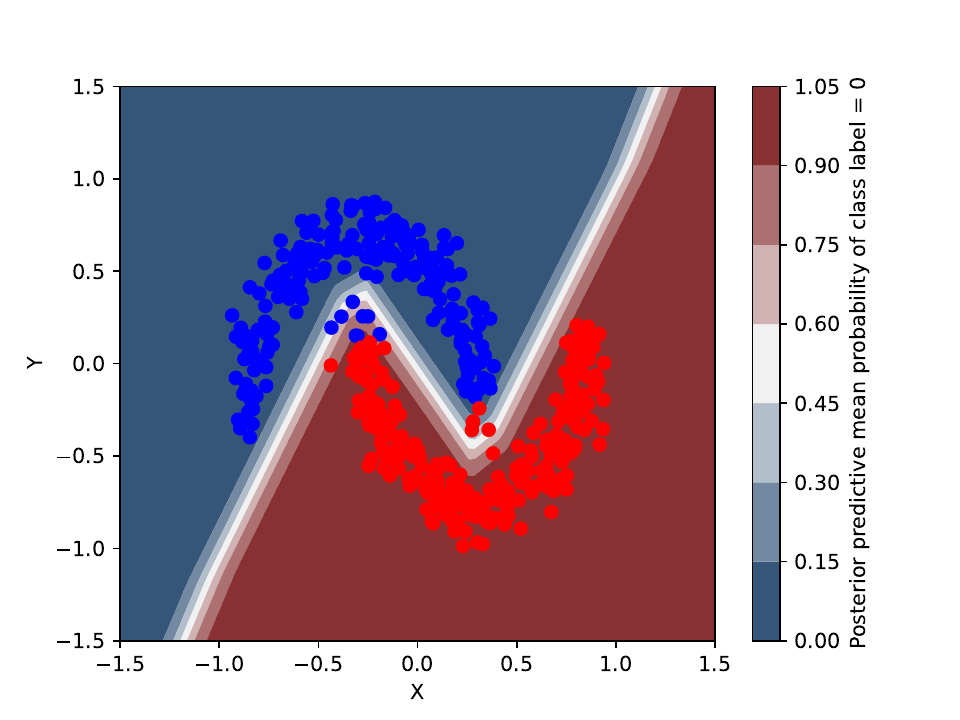}
  \end{subfigure}
  \\
  \begin{subfigure}[t]{0.25\textwidth}
    \centering
    \includegraphics[width=1\linewidth]{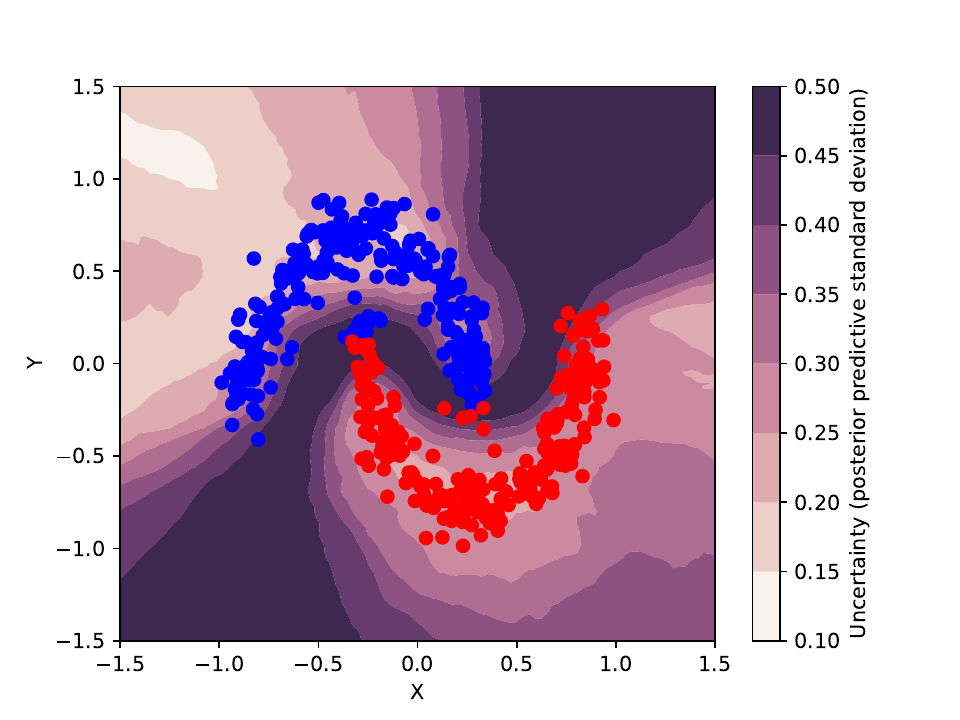}
    \caption{BPS}
  \end{subfigure}%
  \begin{subfigure}[t]{0.25\textwidth}
    \centering
    \includegraphics[width=1\linewidth]{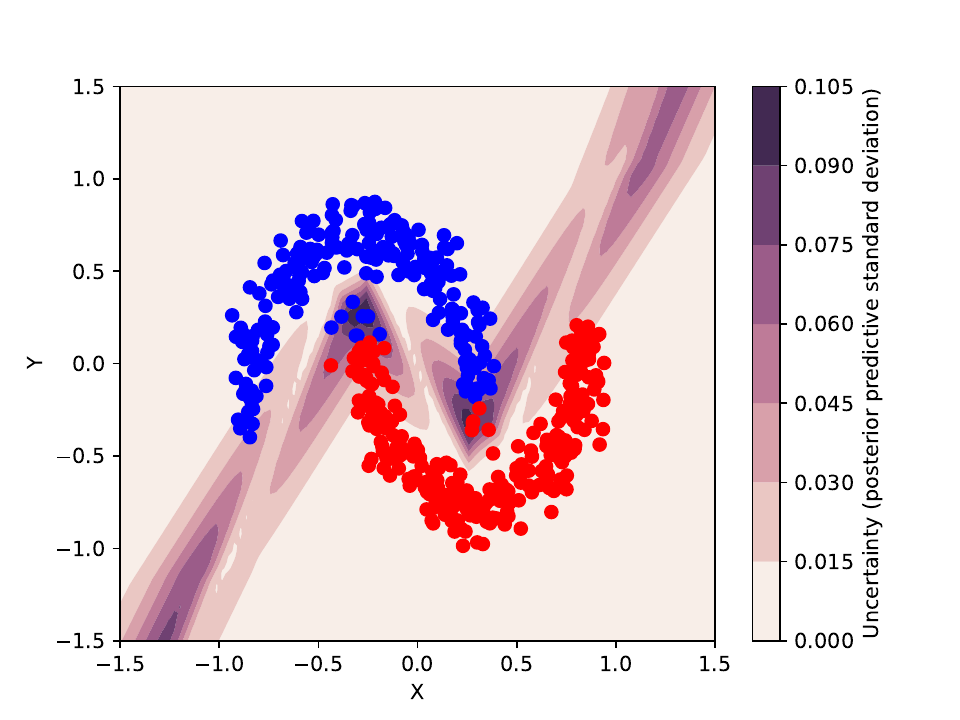}
    \caption{SGLD}
  \end{subfigure}
  \caption{Examples of the predictive mean and variance for synthetic
    classification task. Left column illustrates results using the BPS and the
    right using SGLD. We see increased uncertainty for the BPS outside the range
    of observed data, whilst SGLD shows greater certainty.}
  \label{fig:logistic}
\end{figure}

To visualise predictive performance and uncertainty estimation, regression and
binary classification tasks are formed on synthetic data sets. Description of
the networks used for these tasks is described in Supp. Material E. Before
sampling, a MAP estimate was first found using stochastic optimisation, and was
used to initialise each sampler. 2,000 samples were generated using each
sampling method, with each sampler initialised from the same MAP estimate. The
\sigbps requires an additional warmup period to identify suitable values for the
preconditioner. We achieve this by performing 1,000 initial samples using the
BPS, and standard deviation parameters used for the preconditioner are estimated
from these samples using the Welford algorithm \cite{welford1962note}. These
preconditioner values are then fixed throughout the sampling process. The \Gls{pdmp} methods are
compared against \Gls{sgld} which starts with a learning rate that decays to
zero as required \cite{welling2011bayesian, nagapetyan2017true}, and with no
decay of the learning rate as is commonly done in practice (SGLD-ND). Examples
of the predictive posterior distribution for regression and binary
classification are shown in Figures \ref{fig:regression} and \ref{fig:logistic}
respectively, with full analysis in Supp. Material B. All \Gls{pdmp} models are
fit with the proposed adaptive event thinning procedure.
\par
Results from these experiments affirm that inference from the PDMP models is
suitable for predictive reasoning, with low variance seen within the range of
observed data and greater variance as distance from observed samples increases.
We similarly see an increase in uncertainty along the decision boundary, which
is a desireable property. This is in contrast to \Gls{sgld}, which is able to
provide low variance predictions for in-distribution data, but offers less
predictive uncertainty whilst extrapolating, for  even in the case for larger
non-decreasing learning rates. This highlights the known limitations of SGLD,
that with a decaying learning rate it can fail to explore the posterior, and
with a larger non-decreasing learning rate will converge to dynamics offered by
traditional SGD \cite{brosse2018promises,nagapetyan2017true}.
\par
These tests indicate promising performance in terms of predictive accuracy and
uncertainty estimates. To further demonstrate classification performance, we
move to larger and more complicated models for performing regression and classification on
real-world data sets.
\subsubsection{UCI-Datasets}
\label{sec:uci}
We further evaluate the performance of the PDMP samplers enabled by the proposed
event sampling scheme on datasets from the UCI repository \cite{ucidatasets}. In
Table \ref{tab:uci}, we show performance metrics on the Boston houses dataset,
with the Naval, Energy, Yacht, and Concrete datasets evaluated in Supp. Material
E.2. Each model is fit with 1,000 samples. For these experiments, we further
include the naive \Gls{sghmc} \cite{chen2014stochastic}. Predictive performance
of these models is measured with Root Mean Squared Error (RMSE) and Negative
Log-Likelihood (NLL). Sampling efficiency is evaluated with Effective Sample
Size (ESS) \cite{robert1999monte}. Due to the high dimension of our models, we
perform PCA on returned samples and project them onto the first principal
component to report ESS on the direction of greatest variance within samples.
\par
From these results, we see \Gls{sgld} and \Gls{sghmc} provide a slight
improvement in terms of RMSE, though we see that the Boomerang Sampler
considerably outperforms both of these methods in terms of sample efficiency.
This result follows from the previous sections where we see that SGLD frequently
converges to the SGD solution space, whilst the PDMP samplers can explore the
posterior space. Additional results in Supp. Material E.2 further validate these
results, and show how predictive performance can be improved with PDMP samplers.
\begin{table}[t!]
  \caption{Summary of predictive performance using PDMP samplers with the
    proposed event time sampling methods on the Boston Houses dataset. Negative
    log-likelihood (NLL) and Root Mean Squared Error (RMSE) are reported. Effective sample size (ESS) is measured over the first
    principal component of samples. Results are shown over 5 independent runs
    with standard deviations reported.}
  \label{tab:uci}
  \begin{center}
    \begin{small}
      \scalebox{0.85}{
        \begin{tabular}{l l l l l}
          \toprule
          {\bfseries Inference} & {\bfseries NLL $\downarrow$}
                                & {\bfseries RMSE $\downarrow$}
                                & {\bfseries ESS $\uparrow$}
          \\
          \midrule\midrule[.1em]
          BPS                   & \textbf{0.96 $\pm$ 0.00}      & 2.76 $\pm$ 0.11 & 2.71 $\pm$ 0.02    \\
          \sigbps               & \textbf{0.96 $\pm$ 0.01}      & 2.73 $\pm$ 0.12 & 2.70 $\pm$ 0.01    \\
          Boomerang & \textbf{0.96 $\pm$ 0.01} & 2.77 $\pm$ 0.13 & \textbf{901.07 $\pm$ 105.03} \\
          SGLD                  & \textbf{0.96 $\pm$ 0.00}      & \textbf{2.68 $\pm$ 0.01} & 2.89 $\pm$ 0.00    \\
          SGHMC                 & \textbf{0.96 $\pm$ 0.00}      & 2.69 $\pm$ 0.05 & 2.71 $\pm$ 0.00    \\
          \bottomrule
        \end{tabular}
      }
    \end{small}
  \end{center}
\end{table}

\subsection{Multi-Class Classification}
\label{sec:classification}
We now evaluate the performance of the proposed sampling procedures on the
popular MNIST \cite{lecun1998gradient}, Fashion MNIST \cite{xiao2017fashion},
SVHN \cite{netzer2011reading}, CIFAR-10 and CIFAR-100
\cite{krizhevsky2009learning} data sets using \Glspl{cnn}. For MNIST and
Fashion-MNIST, the LeNet5 architecture was used whilst for SVHN, CIFAR-10, and
CIFAR-100 the modified ResNet20 architecture from \cite{wenzel2020good} was
used. For the Boomerang Sampler, we follow the original
\cite{bierkens2020boomerang} and implement a flat prior, and for other samplers
a normal prior is used.
\begin{table}[t!]
  \caption{Summary of predictive performance using PDMP samplers with the
    proposed event time sampling methods. Negative log-likelihood (NLL) is reported, along with calibration measured using the expected
    calibration error (ECE) \cite{guo2017calibration}. Effective sample size (ESS) is measured over the first
    principal component of samples. Mean and standard deviation in results
    presented over 5 independent runs.}
  \label{tab:conv}
  \begin{center}
    \begin{smaller}
      \scalebox{0.85}{
        \begin{tabular}{l l l l l}
          \toprule
          {\bfseries Inference} & {\bfseries ACC $\uparrow$}
                                & {\bfseries NLL $\downarrow$}
                                & {\bfseries ECE $\downarrow$}
                                & {\bfseries ESS $\uparrow$}
          \\
          \midrule\midrule[.1em]
          \multicolumn{5}{c}{MNIST}
          \\ \midrule
          BPS                   & \textbf{0.99 $\pm$ 0.00}     & 0.08 $\pm$ 0.04          & 5.22 $\pm$ 3.44          & 2.88 $\pm$ 0.07             \\
          $\sigma$BPS           & \textbf{0.99 $\pm$ 0.00}              & \textbf{0.02 $\pm$ 0.00}          & 0.48 $\pm$ 0.22          & 2.72 $\pm$ 0.02             \\
          Boomerang             & \textbf{0.99 $\pm$ 0.00}     & 0.03 $\pm$ 0.00          & 0.85 $\pm$ 0.37          & \textbf{175.26 $\pm$ 29.38} \\
          SGLD                  & \textbf{0.99 $\pm$ 0.00}     & \textbf{0.02 $\pm$ 0.00} & 0.19 $\pm$ 0.02          & 22.74 $\pm$ 0.18            \\
          SGLD-ND               & \textbf{0.99 $\pm$ 0.00}     & \textbf{0.02 $\pm$ 0.00} & \textbf{0.15 $\pm$ 0.01} & 3.00 $\pm$ 0.00             \\
          SGHMC                 & \textbf{0.99 $\pm$ 0.00}     & \textbf{0.02 $\pm$ 0.00} & 0.17 $\pm$ 0.02          & 2.71 $\pm$ 0.00             \\
          \midrule
          \multicolumn{5}{c}{Fashon-MNIST}                                                                                                         \\
          \midrule
          BPS                   & \textbf{0.91 $\pm$ 0.00}     & \textbf{0.25 $\pm$ 0.00} & \textbf{1.00 $\pm$ 0.38} & 2.73 $\pm$ 0.04             \\
          $\sigma$BPS           & \textbf{0.91 $\pm$ 0.00}     & 0.30 $\pm$ 0.01          & 3.94 $\pm$ 0.30          & 2.74 $\pm$ 0.02             \\
          Boomerang             & \textbf{0.91 $\pm$ 0.00}     & 0.28 $\pm$ 0.01          & 3.82 $\pm$ 2.24          & \textbf{143.05 $\pm$ 31.32} \\
          SGLD                  & 0.90 $\pm$ 0.01              & 0.30 $\pm$ 0.01          & 3.09 $\pm$ 1.49          & 102.52 $\pm$ 84.14          \\
          SGLD-ND               & \textbf{0.91 $\pm$ 0.00}     & 0.33 $\pm$ 0.00          & 4.71 $\pm$ 0.01          & 2.97 $\pm$ 0.00             \\
          SGHMC                 & \textbf{0.91 $\pm$ 0.00}     & 0.34 $\pm$ 0.01          & 4.88 $\pm$ 0.13          & 2.71 $\pm$ 0.00             \\
          \midrule
          \multicolumn{5}{c}{SVHN}                                                                                                                 \\
          \midrule
          BPS                   & 0.95 $\pm$ 0.00              & 0.18 $\pm$ 0.00          & \textbf{0.61 $\pm$ 0.09} & 2.71 $\pm$ 0.02             \\
          $\sigma$BPS           & 0.95 $\pm$ 0.00              & 0.19 $\pm$ 0.01          & 0.63 $\pm$ 0.18          & 2.75 $\pm$ 0.02             \\
          Boomerang             & 0.95 $\pm$ 0.00              & 0.21 $\pm$ 0.01          & 3.58 $\pm$ 0.82          & \textbf{172.02 $\pm$ 40.73} \\
          SGLD                  & \textbf{0.96 $\pm$ 0.00}     & \textbf{0.16 $\pm$ 0.00} & 1.33 $\pm$ 0.03          & 21.84 $\pm$ 0.25            \\
          SGLD-ND               & \textbf{0.96 $\pm$ 0.00}     & \textbf{0.16 $\pm$ 0.00} & 1.34 $\pm$ 0.00          & 2.97 $\pm$ 0.00             \\
          SGHMC                 & 0.95 $\pm$ 0.00              & 0.19 $\pm$ 0.00          & 0.83 $\pm$ 0.27          & 2.71 $\pm$ 0.00             \\
          \midrule
          \multicolumn{5}{c}{CIFAR-10}                                                                                                             \\
          \midrule
          BPS                   & 0.80 $\pm$ 0.00              & 0.67 $\pm$ 0.00          & 4.91 $\pm$ 0.65          & 2.70 $\pm$ 0.02             \\
          $\sigma$BPS           & 0.79 $\pm$ 0.01              & 0.72 $\pm$ 0.01          & 8.55 $\pm$ 0.64          & 2.73 $\pm$ 0.02             \\
          Boomerang             & \textbf{0.81 $\pm$ 0.00}     & \textbf{0.59 $\pm$ 0.02} & \textbf{4.17 $\pm$ 1.55} & \textbf{200.00 $\pm$ 0.00}  \\
          SGLD                  & \textbf{0.81 $\pm$ 0.00}     & 1.01 $\pm$ 0.00          & 15.32 $\pm$ 0.04         & 19.78 $\pm$ 0.01            \\
          SGLD-ND               & \textbf{0.81 $\pm$ 0.00}     & 0.99 $\pm$ 0.00          & 15.15 $\pm$ 0.01         & 2.85 $\pm$ 0.00             \\
          SGHMC                 & 0.80 $\pm$ 0.00              & 0.96 $\pm$ 0.00          & 14.64 $\pm$ 0.15         & 2.71 $\pm$ 0.00             \\
          \midrule
          \multicolumn{5}{c}{CIFAR-100}                                                                                                            \\
          \midrule
          BPS                   & 0.63 $\pm$ 0.00              & 1.43 $\pm$ 0.01          & 10.16 $\pm$ 0.21         & 2.72 $\pm$ 0.02             \\
          $\sigma$BPS           & 0.63 $\pm$ 0.00              & \textbf{1.42 $\pm$ 0.00} & \textbf{8.70 $\pm$ 0.00} & 2.80 $\pm$ 0.00             \\
          Boomerang             & 0.63 $\pm$ 0.01              & \textbf{1.42 $\pm$ 0.08} & 12.67 $\pm$ 3.26         & \textbf{54.96 $\pm$ 18.20}  \\
          SGLD                  & \textbf{0.64 $\pm$ 0.00}     & \textbf{1.42 $\pm$ 0.00} & 11.24 $\pm$ 0.32         & 26.79 $\pm$ 0.73            \\
          SGLD-ND               & \textbf{0.64 $\pm$ 0.00}     & 1.45 $\pm$ 0.00          & 12.22 $\pm$ 0.05         & 2.69 $\pm$ 0.00             \\
          SGHMC                 & 0.63 $\pm$ 0.00              & 1.46 $\pm$ 0.00          & 12.63 $\pm$ 0.12         & 2.71 $\pm$ 0.00             \\
          \bottomrule
        \end{tabular}
      }
    \end{smaller}
  \end{center}
\end{table}
\par
Similar to the experiments on regression, a MAP estimate is found and used to
initialise each sampler. 2,000 samples for each model are then generated, though
a thinning factor
of 10 is used to reduce the number of returned samples used for prediction to 200.
For these models, we measure predictive performance and calibration
through the Accuracy, NLL, and Expected Calibration Error (ECE)
\cite{guo2017calibration}, and similarly measure sampling efficiency using the
ESS with on samples after performing PCA.  A full description of the models
used, and experiment parameters is shown in Supp. Material E.3. Table
\ref{tab:conv} summarises the results of these experiments.
\begin{figure}[!t]
  \centering
  \begin{subfigure}[t]{0.18\textwidth}
    \centering \includegraphics[width=\linewidth]{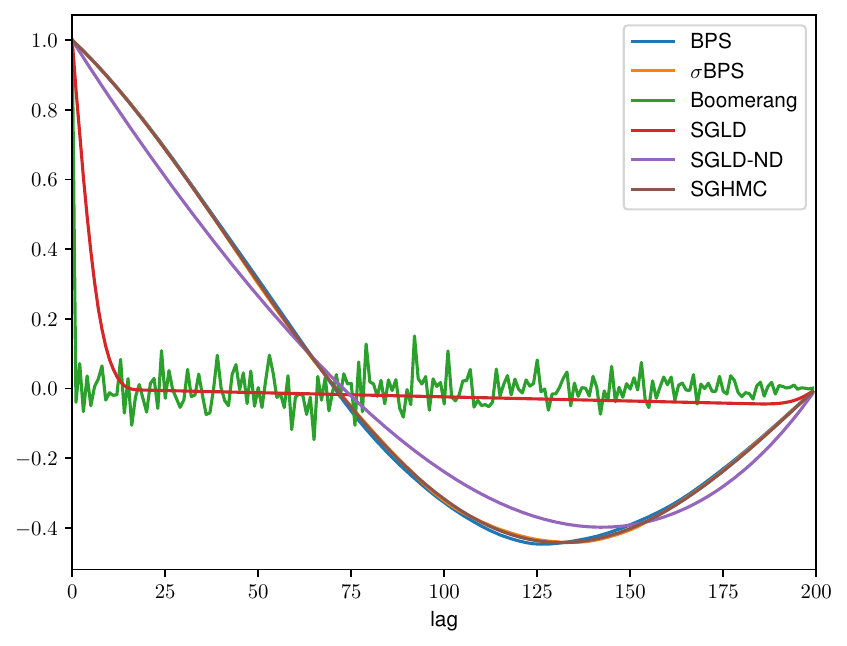}
    \caption{ACF first principal component}
  \end{subfigure}%
  ~
  \begin{subfigure}[t]{0.18\textwidth}
    \centering \includegraphics[width=\linewidth]{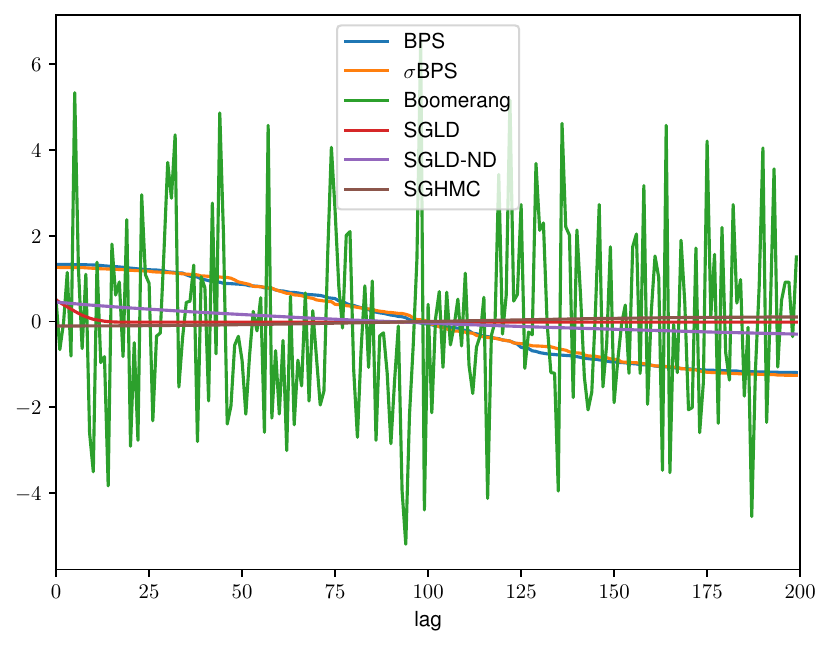}
    \caption{Trace first principal component}
  \end{subfigure}
  \caption{Example of ACF and trace plots for the first principal
    component of the samples from network fit on SVHN dataset.}
  \label{fig:autocorr}
\end{figure}
\begin{figure}[h]
  \centering
  \begin{subfigure}[t]{0.13\textwidth}
    \centering \includegraphics[width=\linewidth]{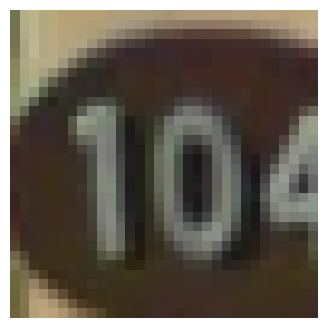}
    \caption{Image Class 0}
  \end{subfigure}%
  \\
  \begin{subfigure}[t]{0.20\textwidth}
    \centering \includegraphics[width=\linewidth]{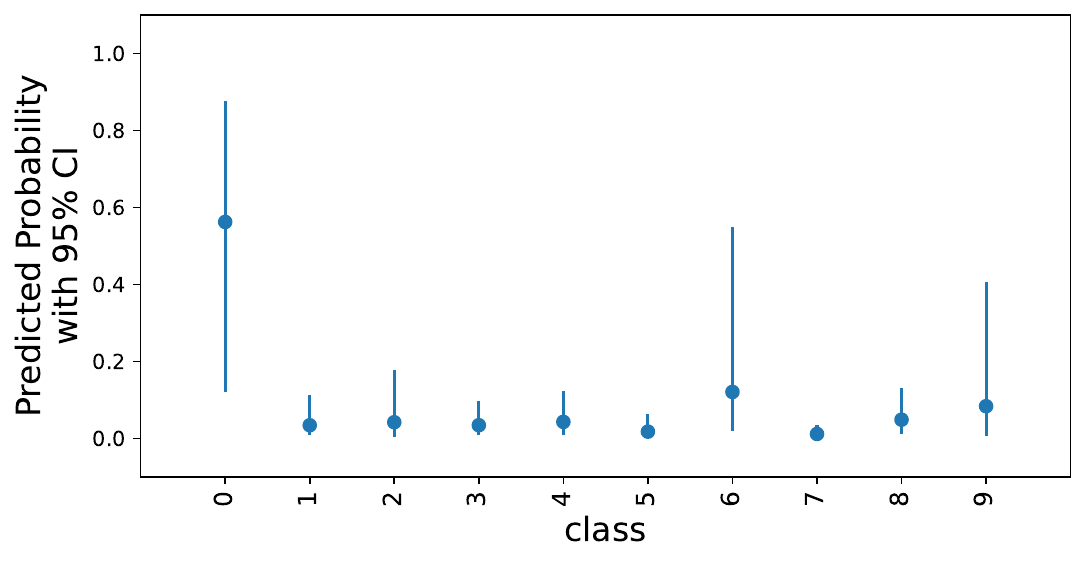}
    \caption{Boomerang Sampler}
  \end{subfigure}%
  ~
  \begin{subfigure}[t]{0.20\textwidth}
    \centering \includegraphics[width=\linewidth]{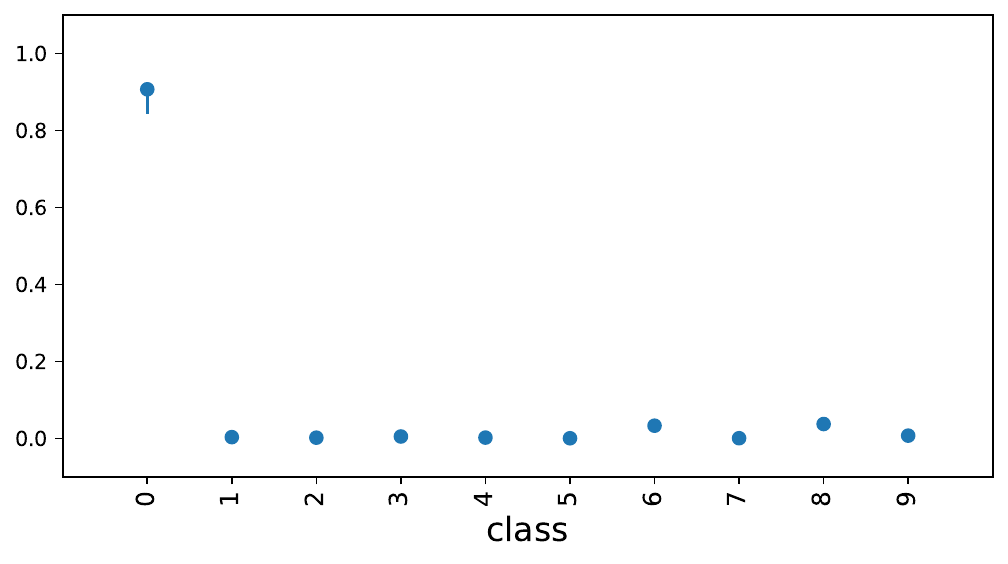}
    \caption{SGLD}
  \end{subfigure}%
  \caption{Examples from predictive posterior for difficult-to-classify samples
    from SVHN. Top row shows the original image and the bottom row shows the
    predictive distribution for the Boomerang sampler and SGLD. The mean for
    each class is represented by the dot, and the 95\% credible intervals shown with the error bars.}
  \label{fig:class_pred}
\end{figure}
\par
These results highlight favourable performance for PDMP samplers. The BPS
and \sigbps samplers frequently provide more calibrated predictions in terms of ECE,
whilst maintaining comparable predictive accuracy to SGLD methods.
Most importantly, we note the  Boomerang
sampler consistently outperforms other samplers in terms of effective
sample size, whilst also promoting competitive predictive accuracy and calibration.
This highlights the potential for these samplers for probabilistic inference within neural networks.
\par
With measures of predictive performance and ESS within our models, we wish to further investigate the mixing properties of the samplers
presented within to identify how well the posterior space is being explored. ESS only gives a measure to approximate the number of independent samples within our MCMC chain, though we are also interested in how well the support for the posterior is being explored.
Given the large number of parameters seen within a BNN, it is infeasible to
evaluate the coordinate trace and autocorrelation plots for individual parameters as is
typically done for MCMC models. Instead, we again perform PCA to reduce the
dimension of our data and investigate the trace plots of the first principal component as illustrated in Figure.
\ref{fig:autocorr}. From these figures, we can identify strong correlation
between samples from the \Gls{bps}, \sigbps, SGHMC, and SGLD-ND solutions. \Gls{sgld} offers reduced correlation in samples, however as seen in the trace plot, samples fail to explore the posterior and instead converge to a steady
state, whilst the Boomerang sampler provides considerably reduced correlation
and more favourable mixing, however we note in Supp. Material C that the
Boomerang sampler can exhibit a mode seeking behaviour due to the nature of the
reference measure.
Convergence of the \Gls{sgld} samples can be attributed to the reduction in learning rate required to target the posterior. We verify this result in Supp.
Material C, where we provide further analysis into results from all networks and remaining principal components. The effect of this convergence in terms of predictive uncertainty is illustrated within Figure \ref{fig:class_pred},
where the \Gls{pdmp} sampler is able to provide more meaningful uncertainty estimates for
difficult-to-classify samples, and the \Gls{sgld} predictive results converge to that
similar of a point estimate. Additional examples of the predictive distributions is
shown in Supp. Material H.
\par
\begin{figure}[!t]
  \centering
  \begin{subfigure}[t]{0.20\textwidth}
    \centering \includegraphics[width=\linewidth]{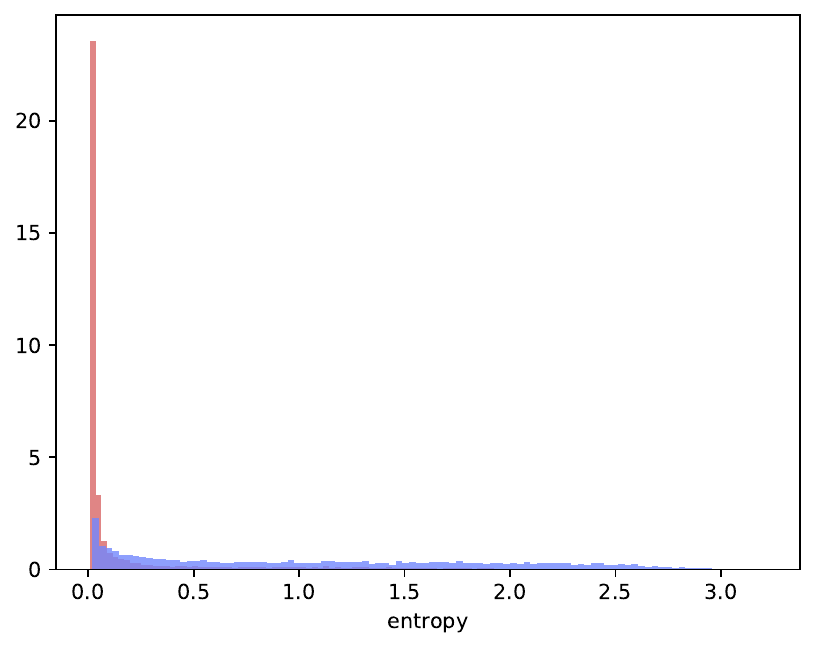}
    \caption{In-distribution}
  \end{subfigure}%
  ~
  \begin{subfigure}[t]{0.20\textwidth}
    \centering \includegraphics[width=\linewidth]{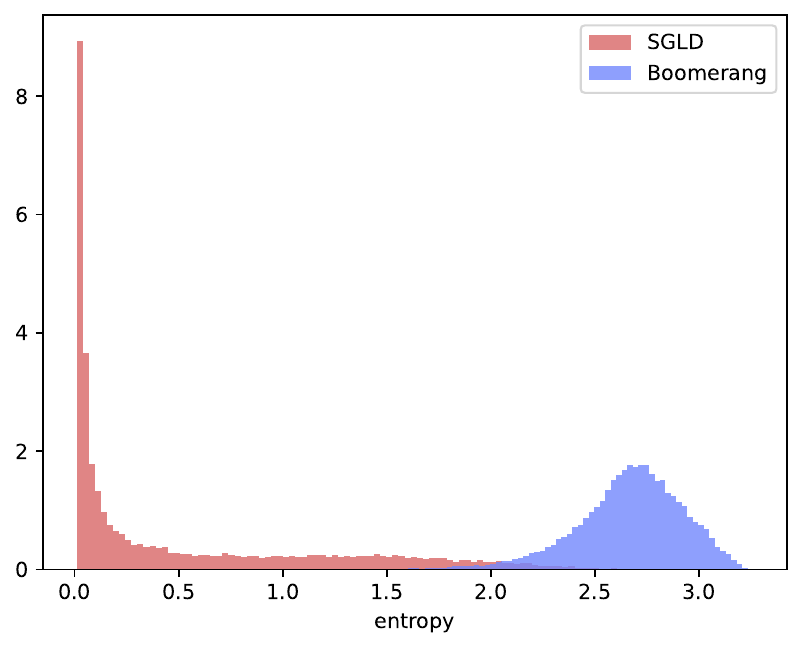}
    \caption{OOD}
  \end{subfigure}
  \caption{Entropy histograms comparing SGLD and Boomerang sampler fit on
    the CIFAR-10 dataset. OOD data represented by SVHN. We see the
    predictive entropy from the Boomerang sampler increases as desired for OOD
    data, whilst SGLD remains overly confident for erroneous samples.}
  \label{fig:entropy}
\end{figure}

Probabilistic methods have shown favourable performance in terms of \Gls{ood}
detection \cite{grathwohl2019your, maddox2019simple}. Given the point-estimate-like nature of the results returned by \Gls{sgld}, we wish to
compare with results from the Boomerang sampler to see if both can offer similar
performance for \Gls{ood} data. We see in Figure \ref{fig:entropy} that the
Boomerang sampler offers greater entropy for \Gls{ood} data, indicating a
desireable increase in aleatoric uncertainty. Additional analysis is provided in Supp. Material G.
\section{Discussion and Limitations}
\label{sec:discussion}
Whilst the \Gls{pdmp} methods have shown favourable performance in terms of predictive
accuracy, calibration and uncertainty in \Glspl{bnn}, there are certain
challenges with fitting them. The PDMP samplers used within this work are designed to target the joint
distribution,
\begin{equation}
  \label{eq:joint}
  p(\myw, \myv) = \pi(\myw) \Phi(\myv)
\end{equation}
where $\pi(\myw)$ is the target posterior and $\Phi(\myv)$ is the distribution of the
auxiliary velocity components which must be set by users in the form of the
refreshment distribution. For the \Gls{bps} and \sigbps samplers, it has been
shown that with a reference distribution may be a Gaussian or restricted to the
unit hypersphere \cite{bouchard2018bouncy,durmus2020geometric}. For the Boomerang sampler, the
velocity distribution is designed with respect to a reference measure to ensure
invariance to the target distribution, such that $\Phi(\myv) = \mathcal{N}(0, \Sigma_{\star})$,
where $\Sigma_{\star}$ is the same factor used to precondition the dynamics. The choice
in distribution used for the velocity component has an explicit effect on the
mixing capabilities of the models when applied to \Glspl{bnn}. We demonstrate
this effect in Supp. Material D.1. We find that a velocity distribution with too
much variance can cause effects similar to that of divergences seen in HMC and
NUTS. Furthermore, we see that with variance set too low, the samplers can fail
to explore the posterior sufficiently to provide the desired meaningful
uncertainty estimates. A similar effect can be seen for the choice of
refreshment rate, which we investigate in Supp. Material D.2. We highlight these
limitations as areas for future research to enable robust application of
\Gls{pdmp} methods for \Glspl{bnn}.
\par
The Boomerang sampler as implemented within this work and the original paper is
probabilistic, though is not purely Bayesian. This is due to the reference
measure being identified through the data itself. A strictly
Bayesian approach can be recovered by setting the reference measure and
associated preconditioner matrix from a prior distribution, though we would lose
some favourable sampling performance offered by this sampler. We can view the
approach implemented within similar to an empirical Bayes, where we are gleaning
information about the reference measure for the Boomerang sampler from
the data itself. Given the difficulty of specifying a meaningful and informative
prior, and the success seen when using empirical priors for \Glspl{bnn}
\cite{krishnan2020specifying}, we believe the use of such an approach for the
Boomerang sampler is justified.
\section{Conclusion}
Within this work, we demonstrate how \Glspl{pdmp} can be used for \Glspl{bnn}.
We provide a flexible piecewise linear bound to enable sampling of event times
within these frameworks that permits inference in \Glspl{bnn}. A
GPU-accelerated software package is offered to increase the availability of
PDMPs for a wide array of models. Experimentation on BNNs for regression and
classification indicates comparable or improved predictive performance and
calibration, though were able to consistently improve sampling efficiency
and uncertainty estimation when compared against existing stochastic inference
methods.
\section{Corrigendum}
\label{sec:corrigendum}
The published version of this paper contained a missing term in
Equation \ref{eq:linear_time} that has is corrected and an incorrect negative sign in the dynamics of the Boomerang Sampler. Furthermore an error in the
software for this paper was identified where the scaling term to correct for
mini-batch sizes for the unbiased gradient estimates was not being
called. Experiments have been rerun to with software corrected and additional
information provided in the supplementary material highlighting how samplers can
be explore the posterior space for \Glspl{bnn}. Additional information for the thinning algorithm has also been included and discussed in the supplementary material.
\bibliography{ref.bib}
 \clearpage

\end{document}

\onecolumn %
\maketitle

\appendix
\section{Tightness of Proposed Approximate Upper Bound}
The key contribution within this work is the proposal of a new generic and
data-dependent thinning method to approximately sample event times from within
PDMP samplers. The quality of this thinning method relies on two key components;
the tightness of the envelope and its ability to provide a strict upper bound.
We want the envelope used to be able to be as close to the true event rate as
possible without reducing below it. This enables maximum efficiency of thinning
methods by reducing the likelihood of a proposed event time will be rejected.
\par
\begin{figure}[!h]
  \centering
  \begin{subfigure}[t]{0.3\textwidth}
    \centering
    \includegraphics[width=1\linewidth]{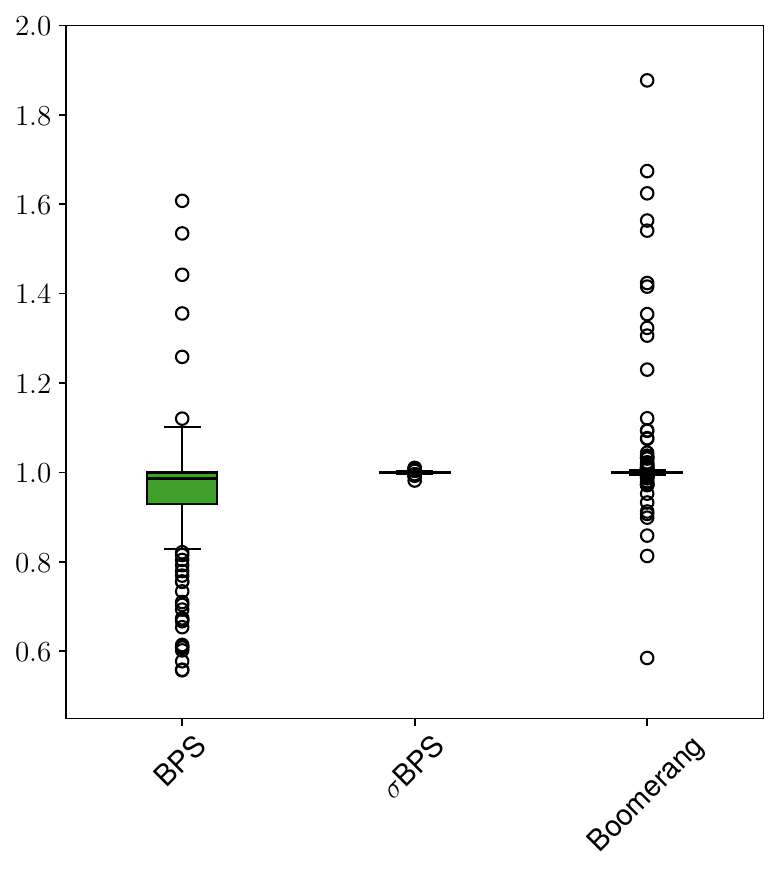}
    \caption{$\alpha=1.0$}
  \end{subfigure}%
  ~
  \begin{subfigure}[t]{0.3\textwidth}
    \centering
    \includegraphics[width=1\linewidth]{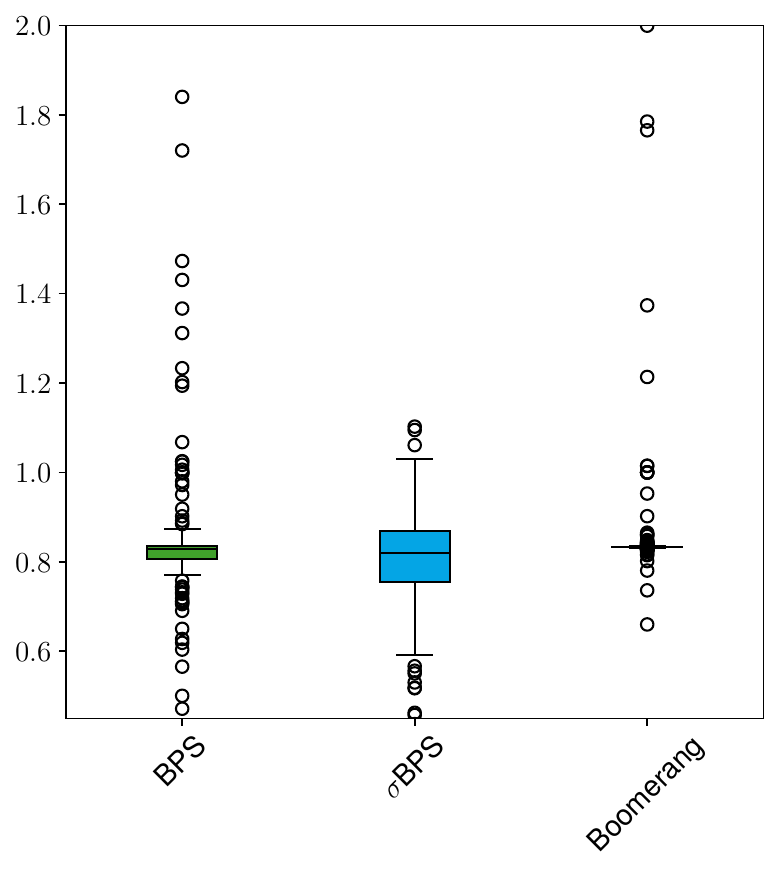}
    \caption{\(\alpha=1.2\)}
  \end{subfigure}%
  ~
  \begin{subfigure}[t]{0.3\textwidth}
    \centering
    \includegraphics[width=1\linewidth]{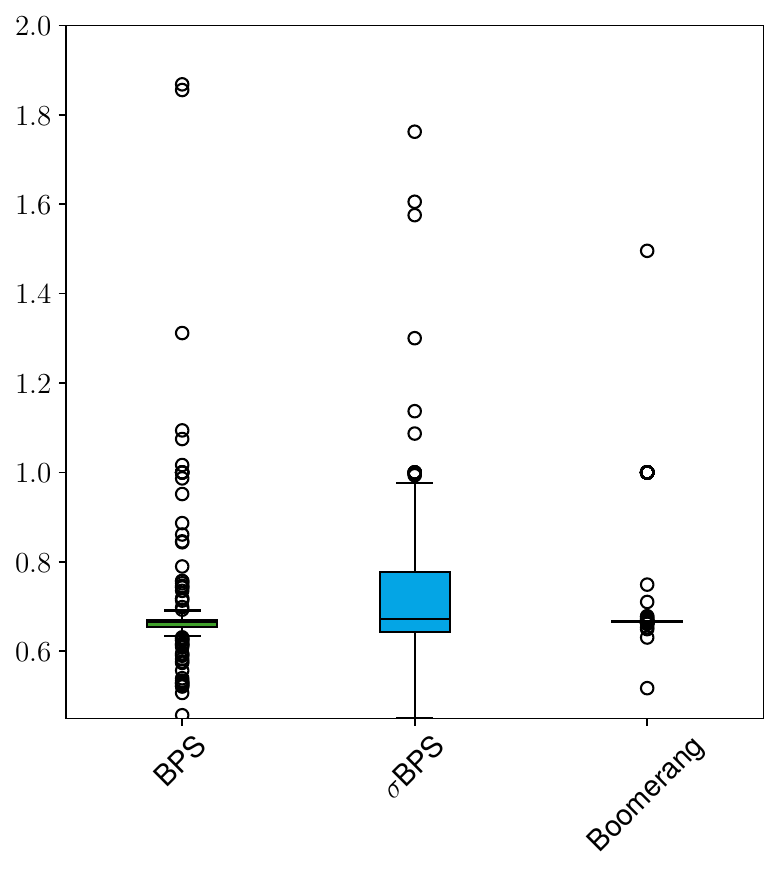}
    \caption{\(\alpha=1.5\)}
  \end{subfigure}%
  \caption{Distribution of acceptance ratios for event thinning across the
    different PDMP samplers used within this work for varying levels
    of $\alpha$. All models are fit on the MNIST data set as described in
    Section 5.2.}
  \label{fig:acceptance}
\end{figure}

Previous works have relied on performing experimentation on simple well-defined
models where derivation of a strict and exact upper
bound\cite{bouchard2018bouncy,bierkens2019zig,bierkens2020boomerang,wu2017generalized}.
Derivation for a strict upper bound is infeasible for neural networks, though we
can assess the quality of our event thinning method by analysing the acceptance
ratio in Equation 8 from the body of the paper. We want this
ratio to be as close to one as possible without exceeding it, otherwise, the
envelope section used to propose the time is below the true event rate. We
assess the distribution of these acceptance ratios for varying values
of $\alpha$ from Equation 12 in the body of the paper in
Figure \ref{fig:acceptance} and we illustrate the result on predictive
performance and computational load in Table \ref{tab:acceptance}.
\begin{table}[!h]
  \caption{Summary of predictive performance with and timings as the scaling value of $\alpha$ is increased for the PDMP samplers demonstrated within. All models are fit to the MNIST dataset using the Lenet5 architecture.}
  \label{tab:acceptance}
  \begin{center}
    \begin{small}
      \scalebox{0.85}{
        \begin{tabular}{c l l l l l l }
          \toprule
          $\mathbf{\alpha}$ & {\bfseries Inference} & {\bfseries ACC}
                            & {\bfseries NLL}
                            & {\bfseries ECC}
                            & {\bfseries Time}
          \\
          \midrule\midrule[.1em]
          \multirow{3}{2cm}{$\alpha=1.0$}
                            & BPS                   & 0.9896          & 0.0536 & 2.66    & 71  \\
                            & \sigbps               & 0.9923          & 0.0227 & 0.4127  & 121 \\
                            & Boomerang             & 0.9919          & 0.0230 & 0.139   & 78  \\
          \midrule
          \multirow{3}{2cm}{$\alpha=1.2$}
                            & BPS                   & 0.9902          & 0.0638 & 3.74    & 74  \\
                            & \sigbps               & 0.9922          & 0.0247 & 2.7380  & 118 \\
                            & Boomerang             & 0.9861          & 0.0456 & 33.2281 & 135 \\
          \midrule
          \multirow{3}{2cm}{$\alpha=1.5$}
                            & BPS                   & 0.987           & 0.0614 & 2.909   & 79  \\
                            & \sigbps               & 0.9925          & 0.0244 & 0.3858  & 126 \\
                            & Boomerang             & 0.9922          & 0.0232 & 0.1651  & 160 \\
          \bottomrule
        \end{tabular}
      }
    \end{small}
  \end{center}
\end{table}
\par
From this experiment, we note that for $\alpha=1$ we see tight clustering of
acceptance ratios around one for the \sigbps and Boomerang Sampler. Whilst this
may initially indicate a tight bound being found, it will in fact indicate that
our approximate upper bound was initially violated and then corrected. This
occurs if our rate for the event switching $\lambda(t)$ is zero and then rapidly
increases. If our event rate changes such that the acceptance
ratio $\lambda(t) / h(t)$ is greater than our rejection condition factor $R$,
the proposal will be rejected with hopes to correct it on subsequent samples.
Whilst the proposed method is capable of correcting the event rate on the next
proposal, bias will be introduced during inference due to the initial proposal
exceeding $h(t)$. We see that as $\alpha$ increases, the distribution of
acceptance ratio for the \sigbps broadens, and from Table~\ref{tab:acceptance}, the
run-time decreases as fewer rejections due to this process occurs, whilst
as \(\alpha\) increases further, the run-time begins to increase.
We can see that with $\alpha=1.0$, we see frequent occurrences of the proposed
envelope being below that of the true event rate, though as we increase the
value of $\alpha$, the likelihood of the approximate envelope being a strict
upper bound increases. In practice, setting this scaling parameter can be
achieved through the use of a small warm-up phase at the start of sampling to
find a ratio that satisfies a users willingness to mitigate bias that may be
induced due to the violation of the upper bound assumption. To mitigate
potential bias, the value of \(\alpha\) may be increased at the expense of a
small increase in the computational demands of the thinning method as seen in
Table \ref{tab:acceptance}. The violation of the bounds within the Boomerang
sampler can be attributed to the non-linear dynamics of the process. We
highlight this phenomenon as an direction for future research.
\section{Additional Regression and Binary Classification Examples}
\begin{figure}[!h]
  \centering
  \begin{subfigure}[t]{0.18\textwidth}
    \centering \includegraphics[width=1\linewidth]{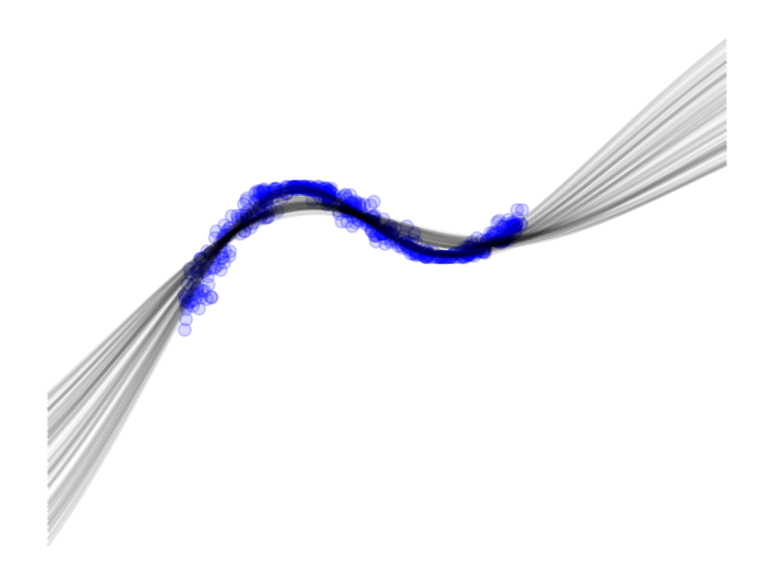}
  \end{subfigure}%
  ~
  \begin{subfigure}[t]{0.18\textwidth}
    \centering \includegraphics[width=1\linewidth]{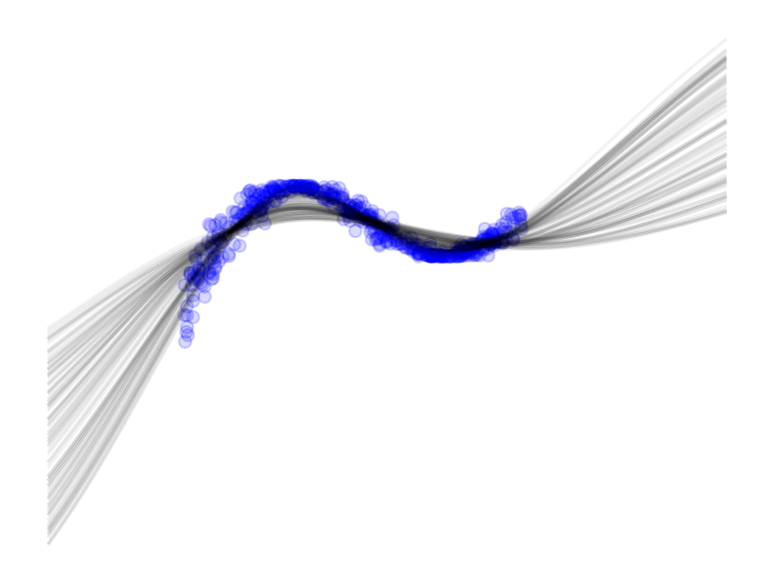}
  \end{subfigure}%
  ~
  \begin{subfigure}[t]{0.18\textwidth}
    \centering \includegraphics[width=1\linewidth]{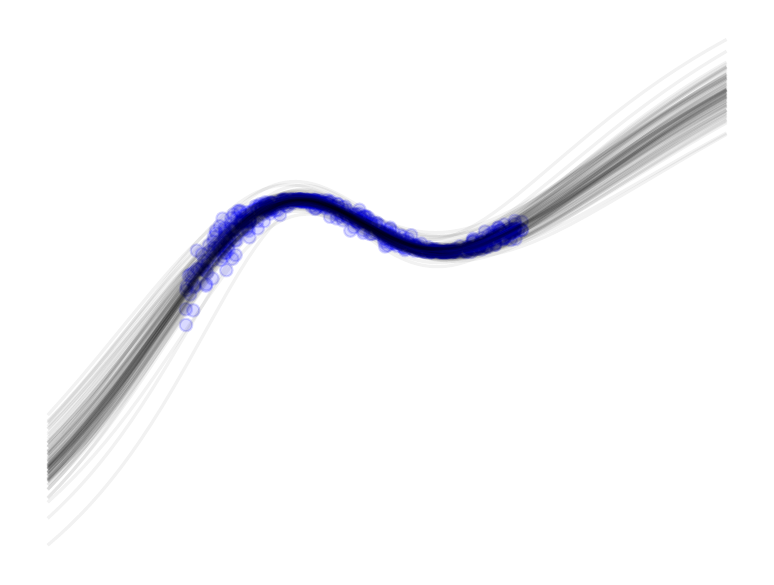}
  \end{subfigure}%
  ~
  \begin{subfigure}[t]{0.18\textwidth}
    \centering \includegraphics[width=1\linewidth]{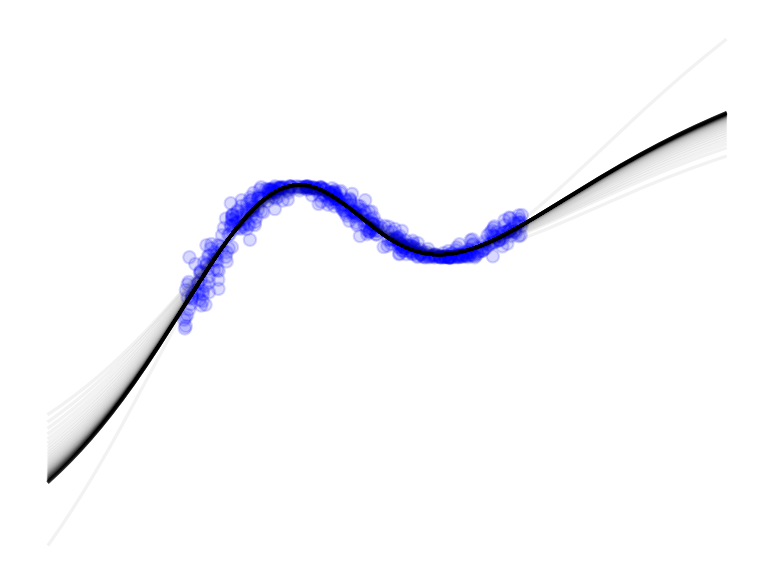}
  \end{subfigure}%
  ~
  \begin{subfigure}[t]{0.18\textwidth}
    \centering \includegraphics[width=1\linewidth]{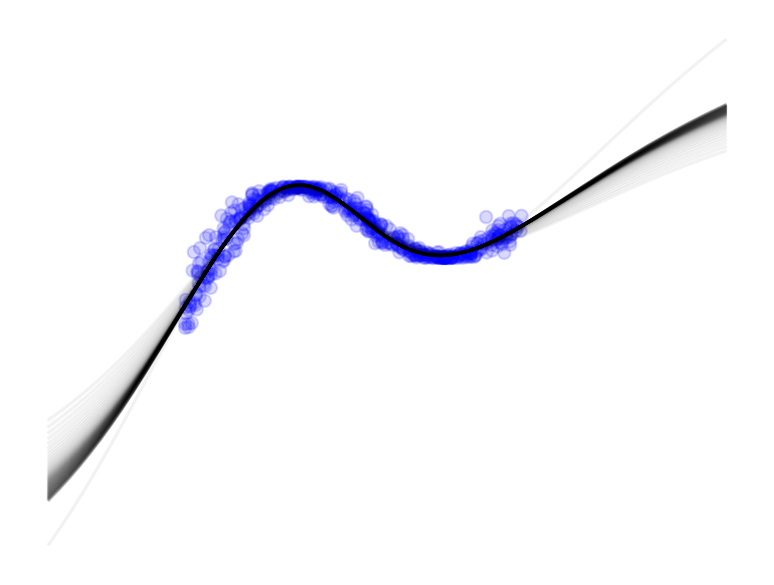}
  \end{subfigure}%
  \\
  \begin{subfigure}[t]{0.18\textwidth}
    \centering \includegraphics[width=1\linewidth]{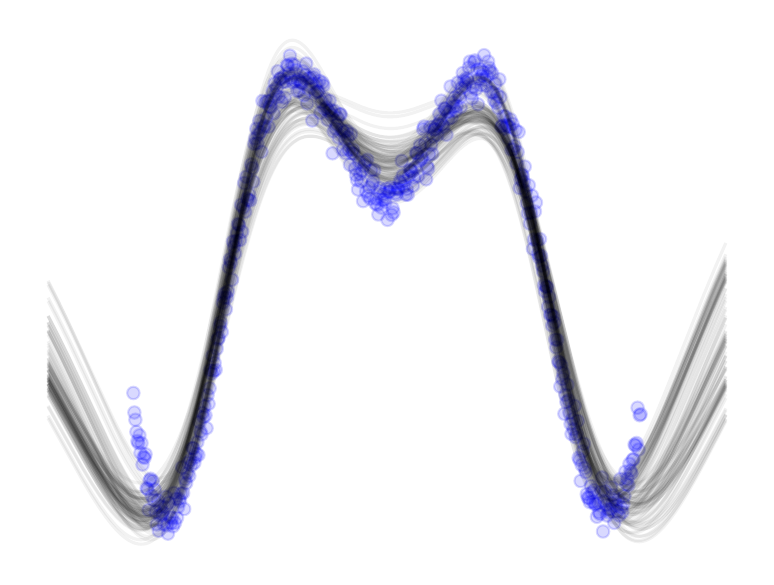}
    \caption{BPS}
  \end{subfigure}%
  ~
  \begin{subfigure}[t]{0.18\textwidth}
    \centering \includegraphics[width=1\linewidth]{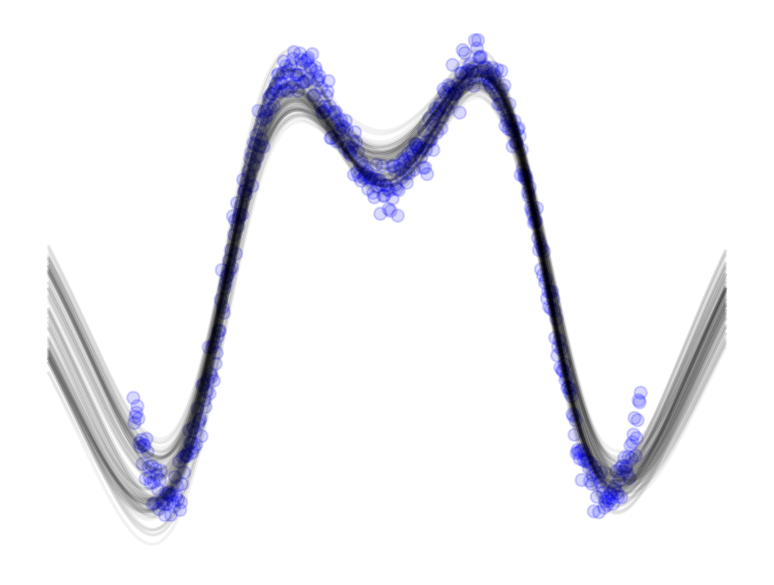}
    \caption{\sigbps}
  \end{subfigure}%
  ~
  \begin{subfigure}[t]{0.18\textwidth}
    \centering \includegraphics[width=1\linewidth]{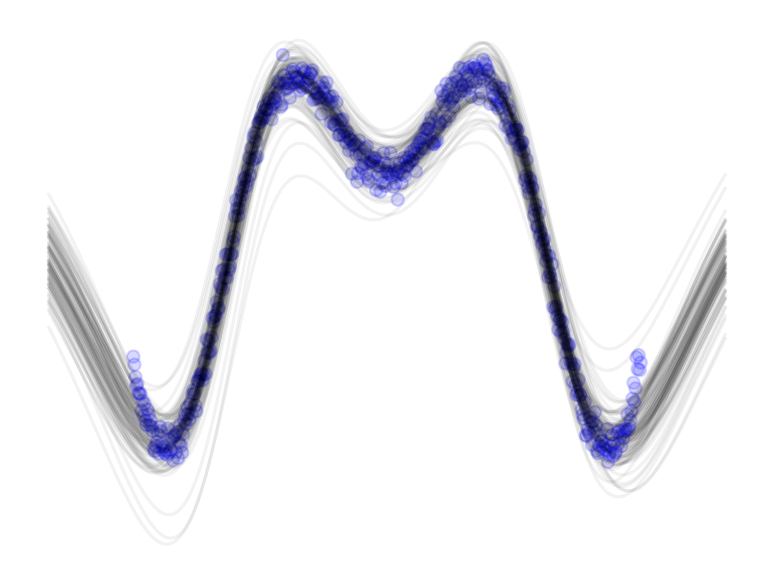}
    \caption{Boomerang}
  \end{subfigure}%
  ~
  \begin{subfigure}[t]{0.18\textwidth}
    \centering \includegraphics[width=1\linewidth]{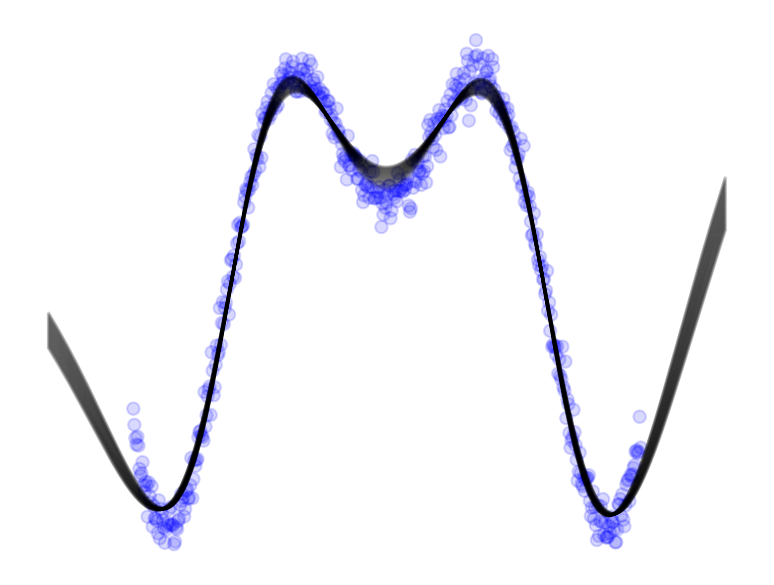}
    \caption{SGLD}
  \end{subfigure}%
  ~
  \begin{subfigure}[t]{0.18\textwidth}
    \centering \includegraphics[width=1\linewidth]{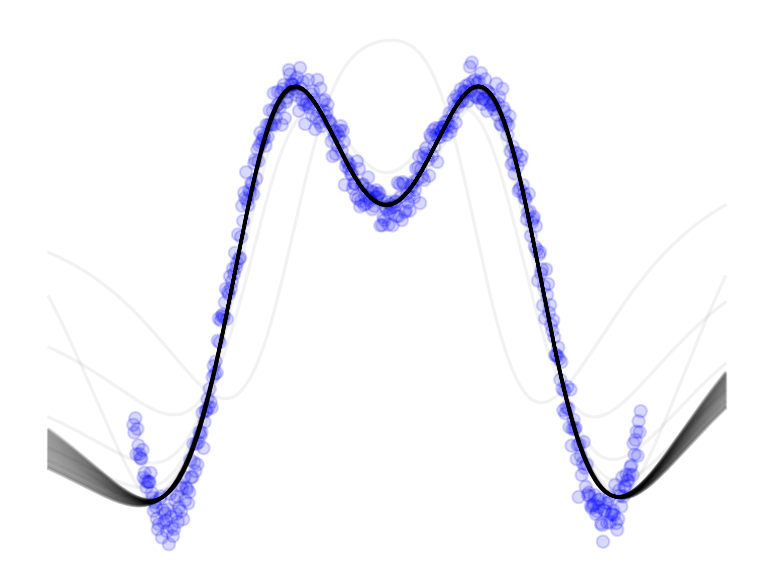}
    \caption{SGLD-ND}
  \end{subfigure}%

  \caption{Exampled of predictive posteriors for BNN regression models across synthetic data sets. Training samples are shown in blue dots, and draws from the predictive distrubtion shown with black lines.}
  \label{fig:regression}
\end{figure}

\begin{figure}[!h]
  \centering
  \begin{subfigure}[t]{0.18\textwidth}
    \centering \includegraphics[width=1\linewidth]{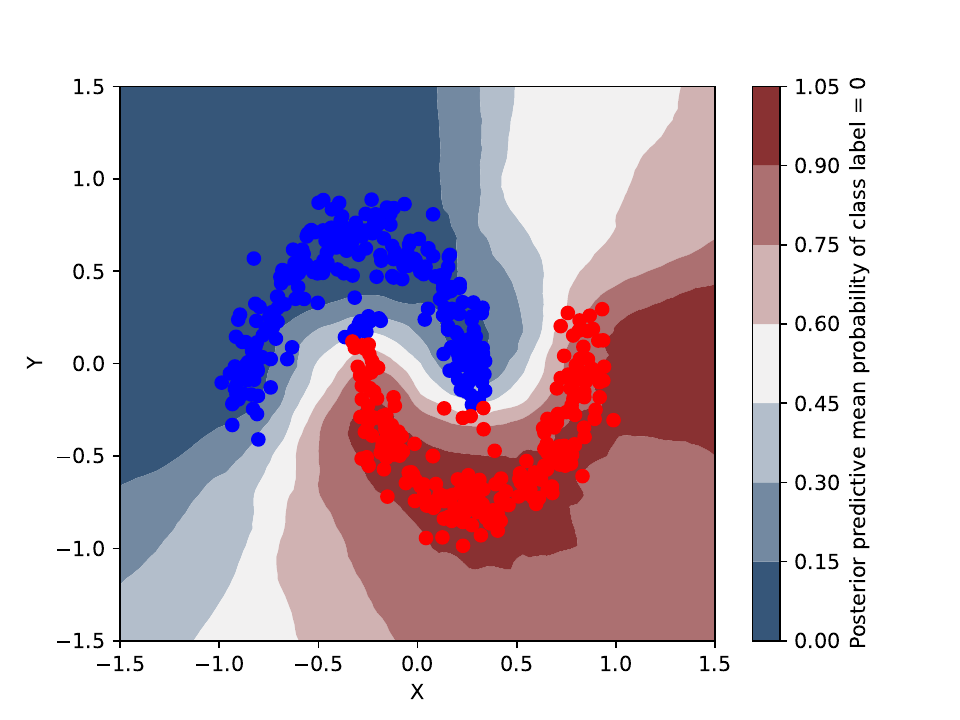}
  \end{subfigure}
  ~
  \begin{subfigure}[t]{0.18\textwidth}
    \centering \includegraphics[width=1\linewidth]{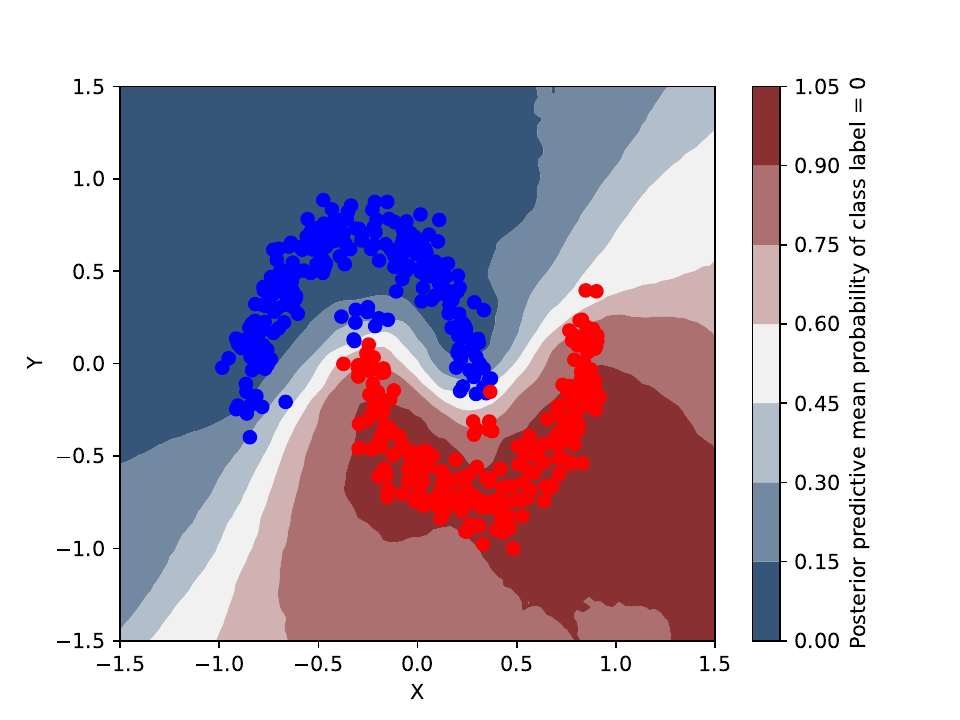}
  \end{subfigure}
  ~
  \begin{subfigure}[t]{0.18\textwidth}
    \centering \includegraphics[width=1\linewidth]{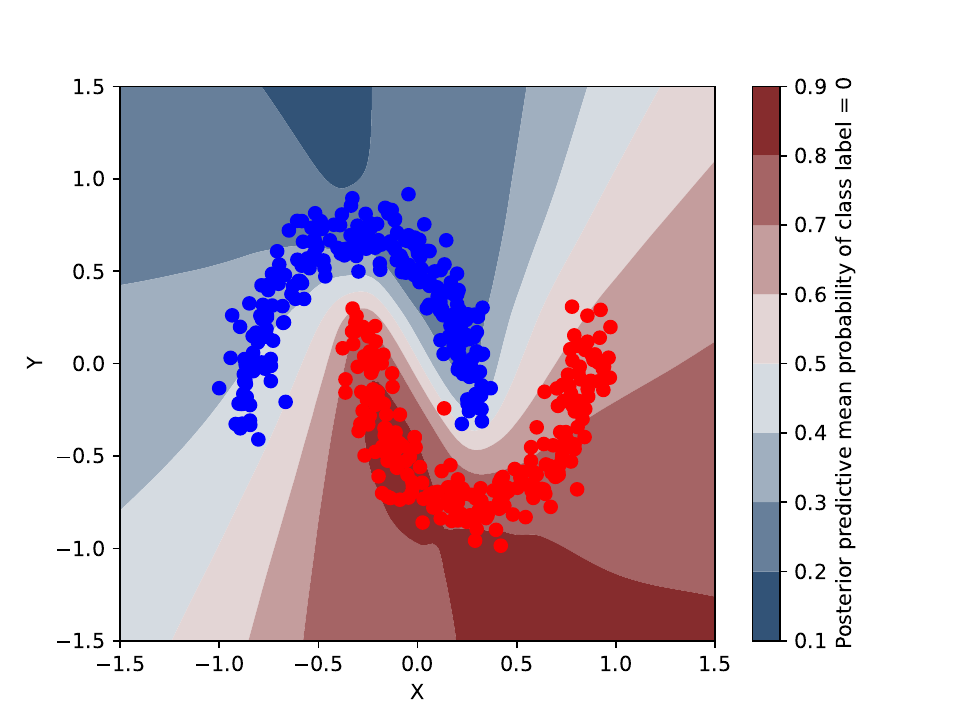}
  \end{subfigure}
  ~
  \begin{subfigure}[t]{0.18\textwidth}
    \centering \includegraphics[width=1\linewidth]{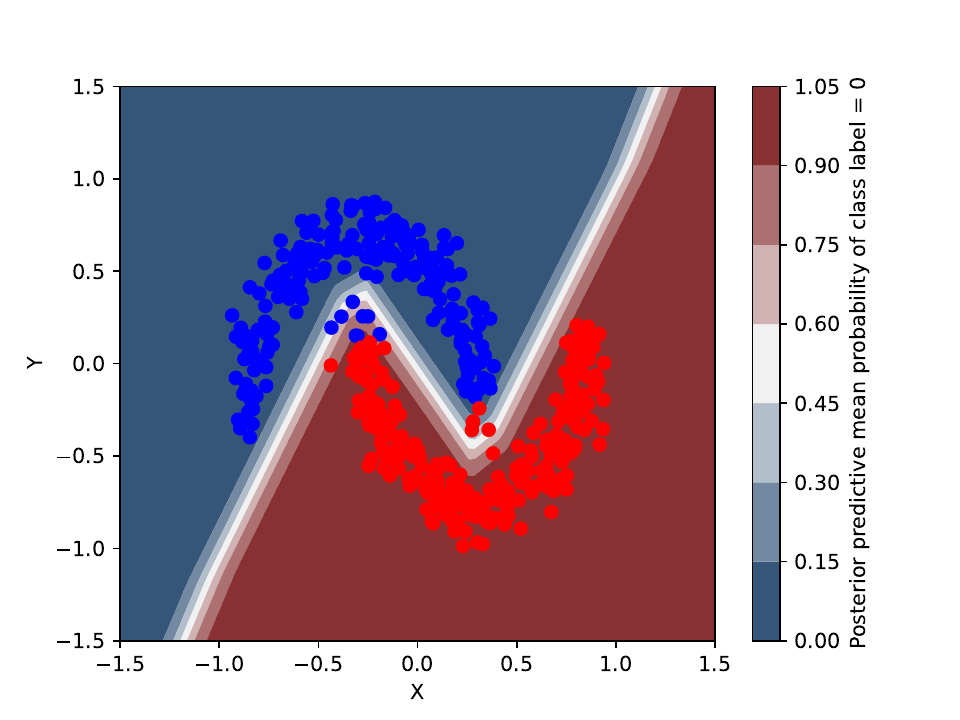}
  \end{subfigure}
  ~
  \begin{subfigure}[t]{0.18\textwidth}
    \centering \includegraphics[width=1\linewidth]{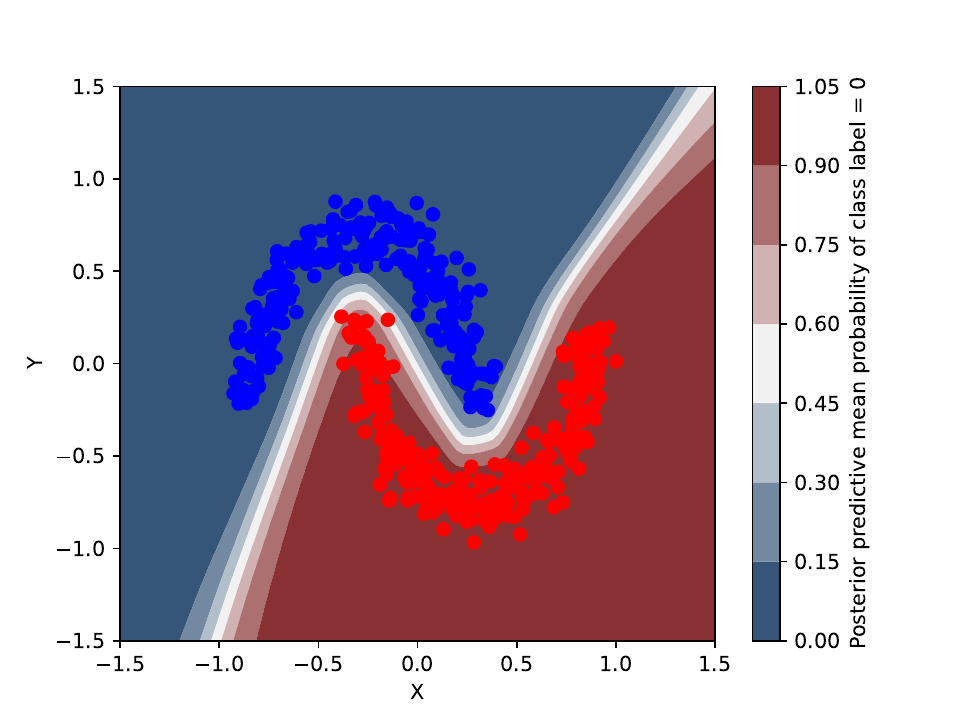}
  \end{subfigure}
  \\
  \begin{subfigure}[t]{0.18\textwidth}
    \centering \includegraphics[width=1\linewidth]{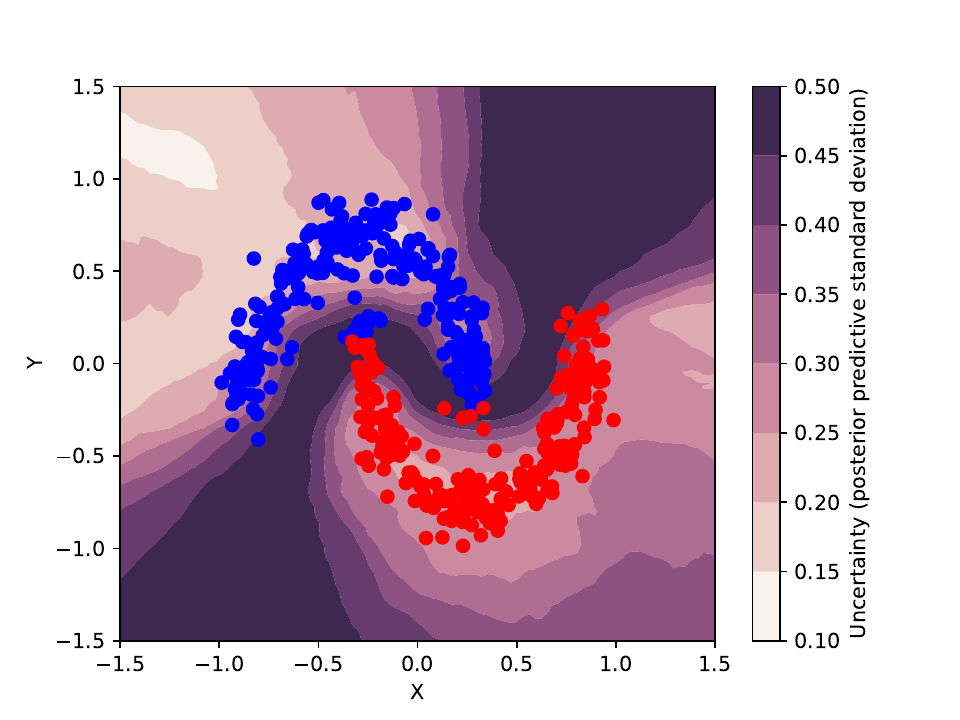}
    \caption{BPS}
  \end{subfigure}
  ~
  \begin{subfigure}[t]{0.18\textwidth}
    \centering \includegraphics[width=1\linewidth]{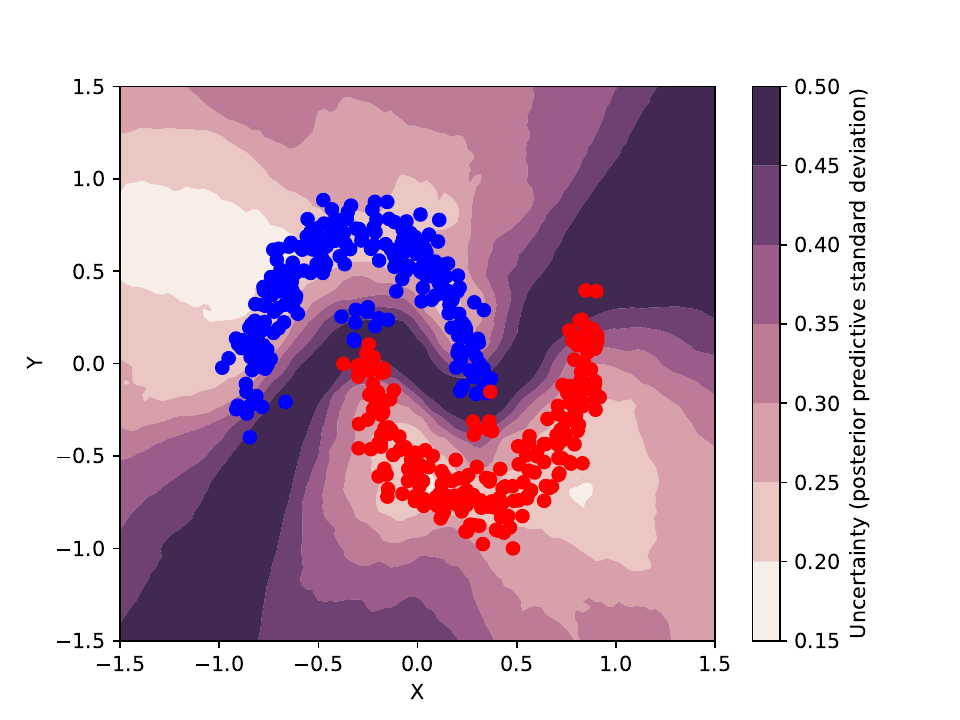}
    \caption{\sigbps}
  \end{subfigure}
  ~
  \begin{subfigure}[t]{0.18\textwidth}
    \centering \includegraphics[width=1\linewidth]{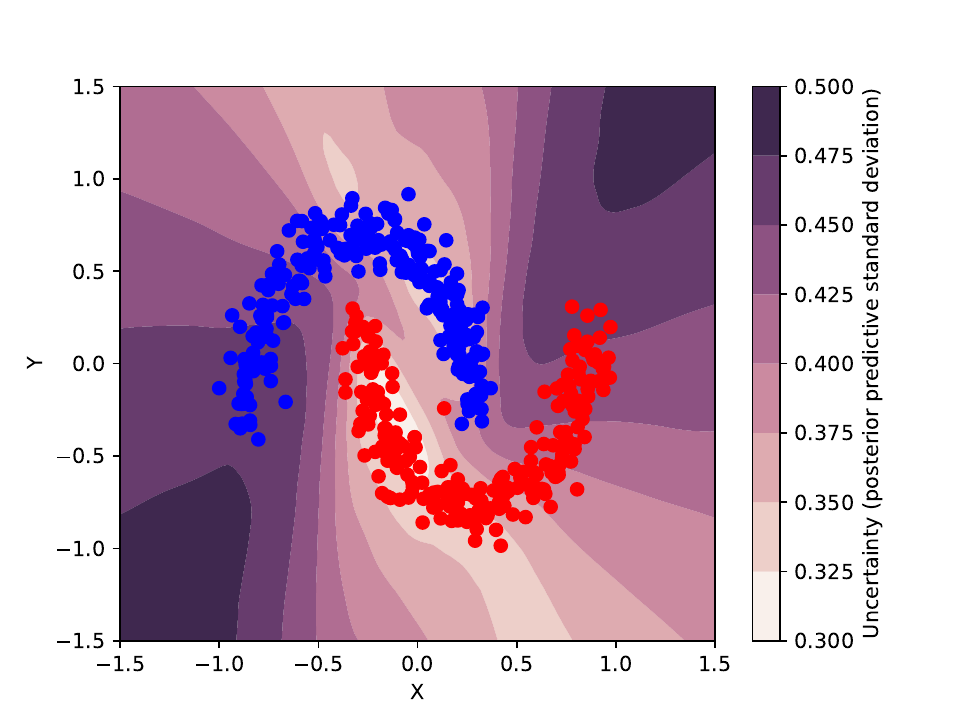}
    \caption{Boomerang}
  \end{subfigure}
  ~
  \begin{subfigure}[t]{0.18\textwidth}
    \centering \includegraphics[width=1\linewidth]{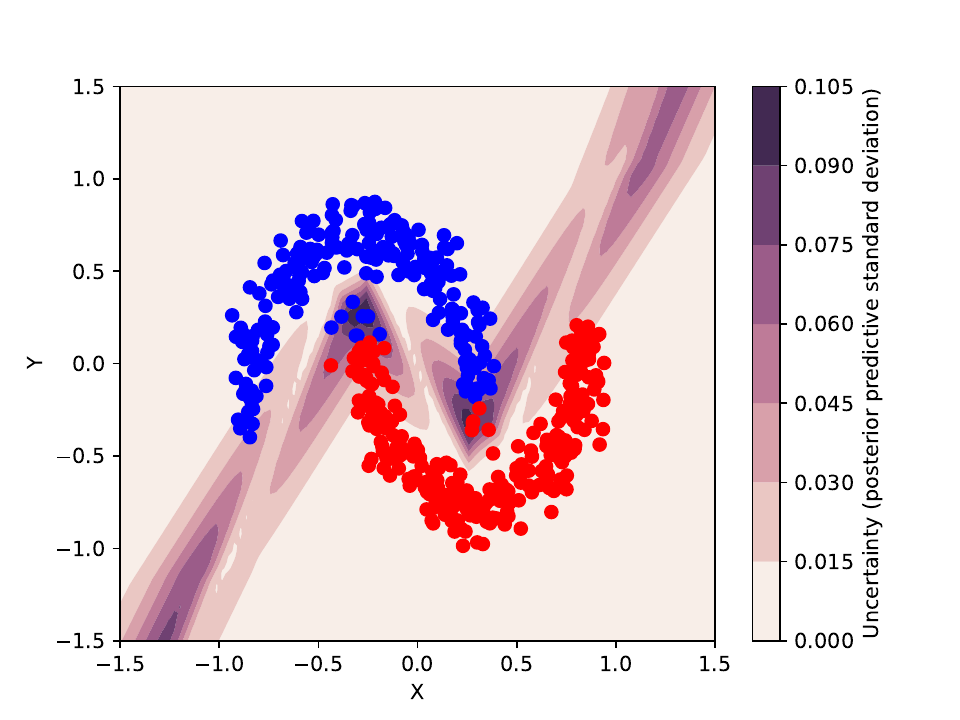}
    \caption{SGLD}
  \end{subfigure}
  ~
  \begin{subfigure}[t]{0.18\textwidth}
    \centering \includegraphics[width=1\linewidth]{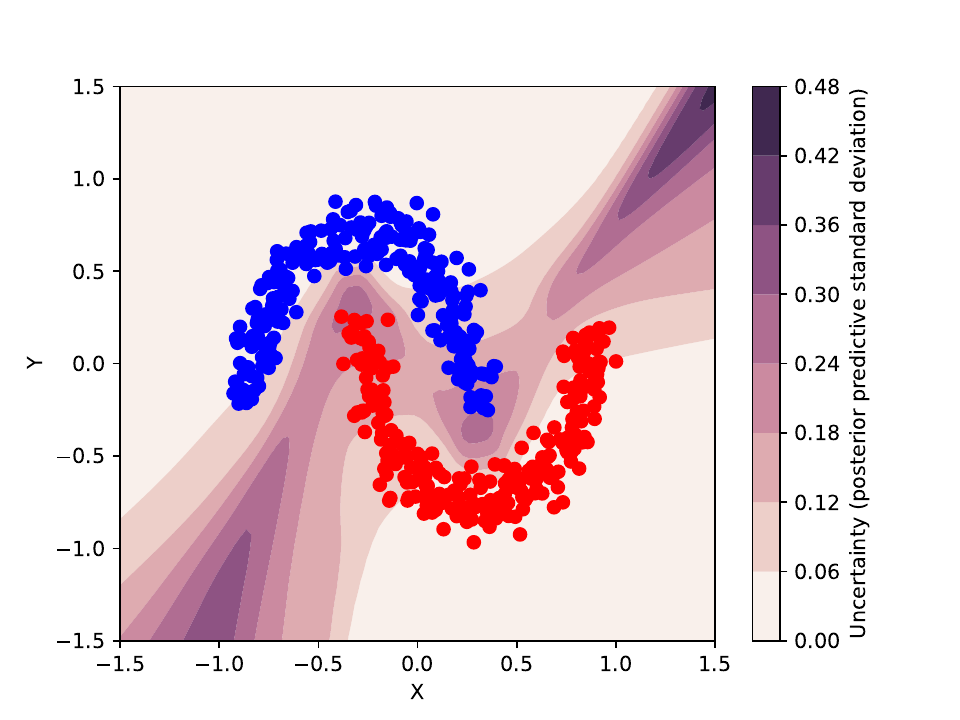}
    \caption{SGLD-ND}
  \end{subfigure}
  \caption{Examples of predictive distributions for synthetic binary classification task. Top row indicates predictive mean and bottom row illustrates variance in predictions. Best viewed on a computer screen in colour.}
  \label{fig:supp_logistic}
\end{figure}

To further validate the predictive performance of PDMP samplers using the
proposed event thinning method, we provide additional examples on easy to
visualise regression tasks in Figure. \ref{fig:regression} and Figure
\ref{fig:supp_logistic} which are compared with \Gls{sgld} with a decreasing
learning rate as required, and a constant learning rate with no decay as is
typically done in practice (SGLD-ND). For regression models, a learning rate
of $10^{-5}$ is used, and for binary
classification models the learning rate is set to $10^{-2}$.  The learning rate
for \Gls{sgld} experiments decays to zero linearly.
\par
From these results, we further validate the predictive performance of these
samplers and their ability to yield informative uncertainty information for
out-of-distribution data when compared to SGLD with a decaying learning rate and
a constant learning rate. We find that even with a larger value learning rate
used for SGLD-ND that the sampler is unable to explore the posterior
sufficiently to provide meaningful uncertainty estimates. This phenomenon has
been reported in \cite{brosse2018promises}, where they identify that even with a
larger and constant learning rate, \Gls{sgld} dynamics converge to that of regular SGD.
\section{Mixing performance}
\begin{figure*}[!hbt]
  \begin{subfigure}[t]{0.25\textwidth}
    \centering
    \includegraphics[width=1\linewidth]{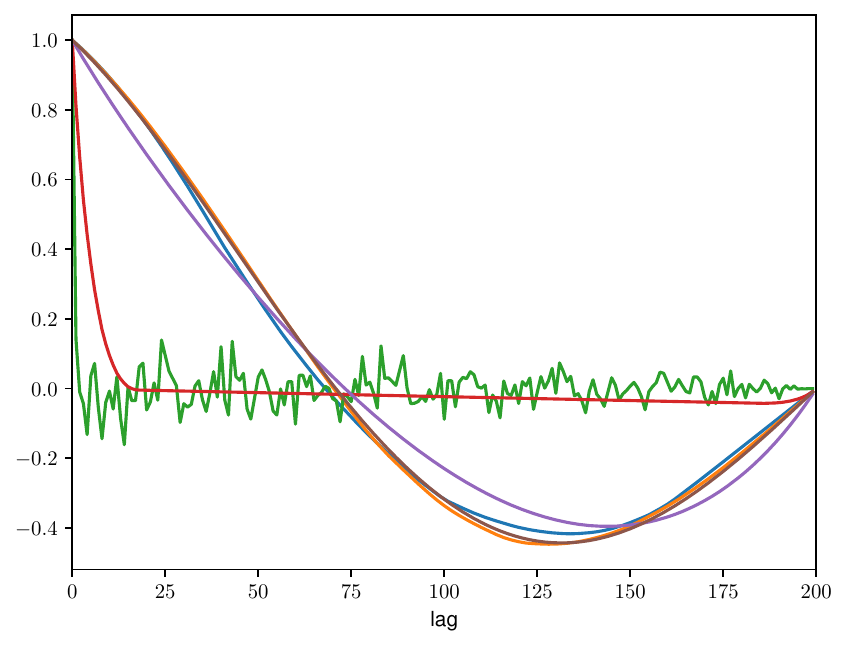}
  \end{subfigure}%
  \begin{subfigure}[t]{0.25\textwidth}
    \centering
    \includegraphics[width=1\linewidth]{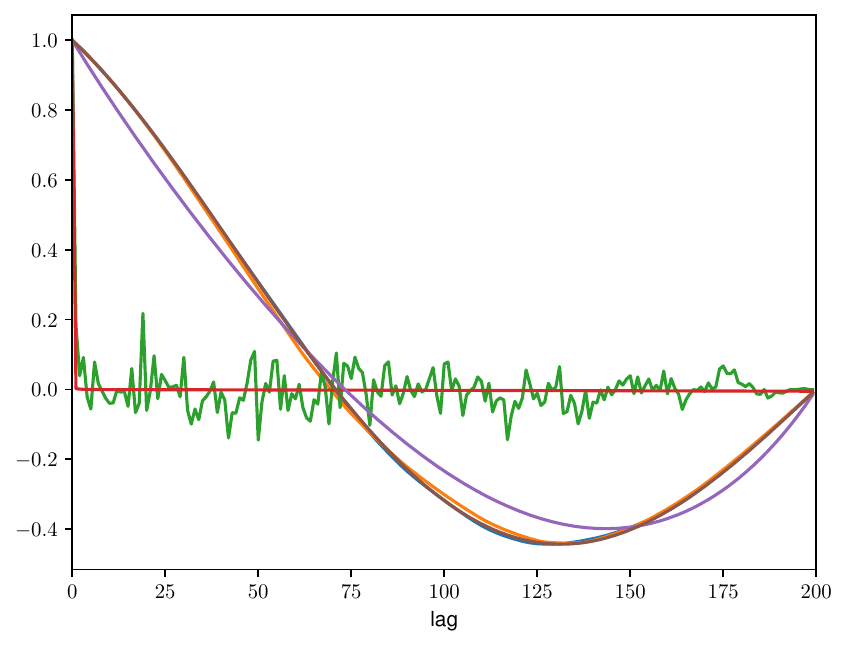}
  \end{subfigure}
  \begin{subfigure}[t]{0.25\textwidth}
    \centering
    \includegraphics[width=1\linewidth]{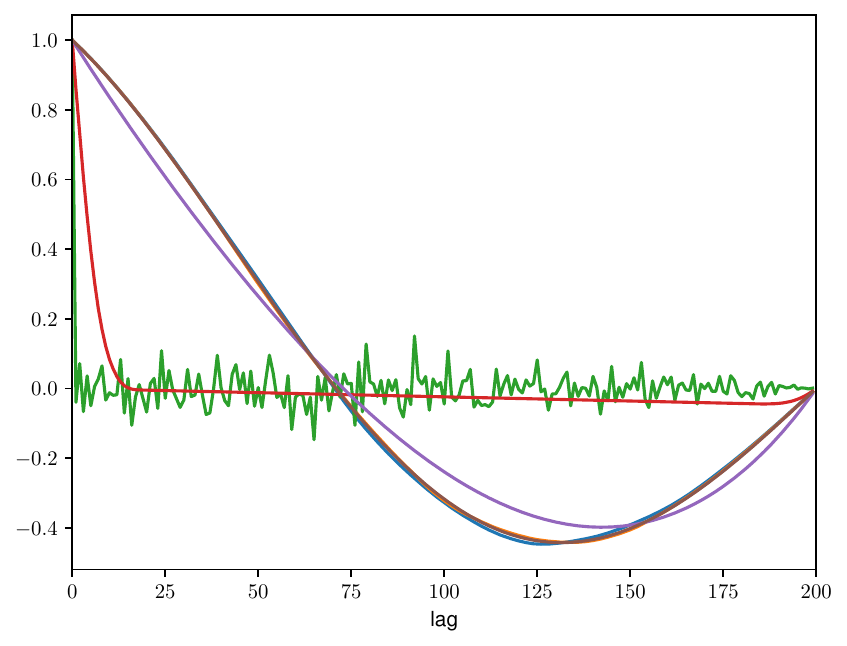}
  \end{subfigure}%
  \begin{subfigure}[t]{0.25\textwidth}
    \centering
    \includegraphics[width=1\linewidth]{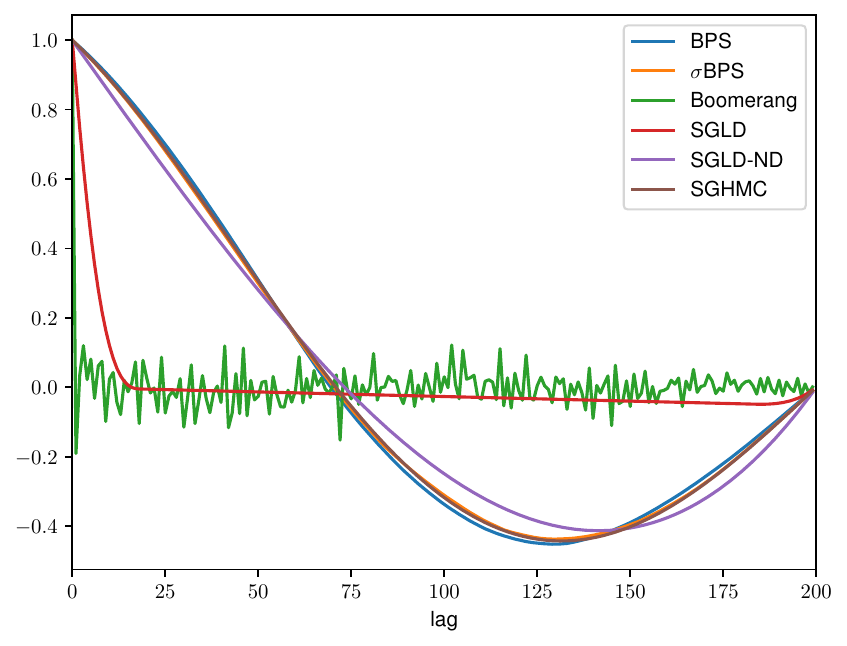}
  \end{subfigure}
  \\
  \begin{subfigure}[t]{0.25\textwidth}
    \centering
    \includegraphics[width=1\linewidth]{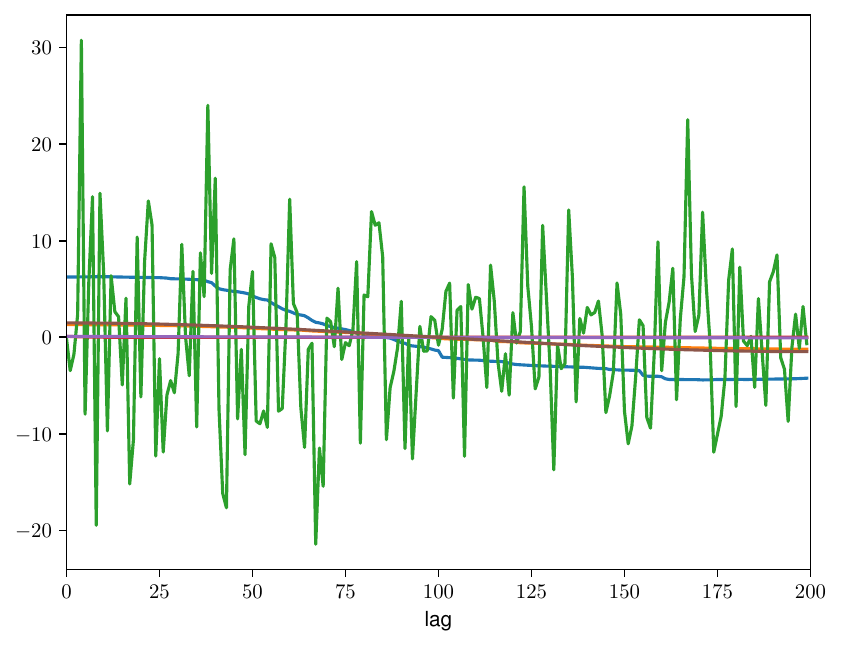}
    \caption{MNIST}
  \end{subfigure}%
  \begin{subfigure}[t]{0.25\textwidth}
    \centering
    \includegraphics[width=1\linewidth]{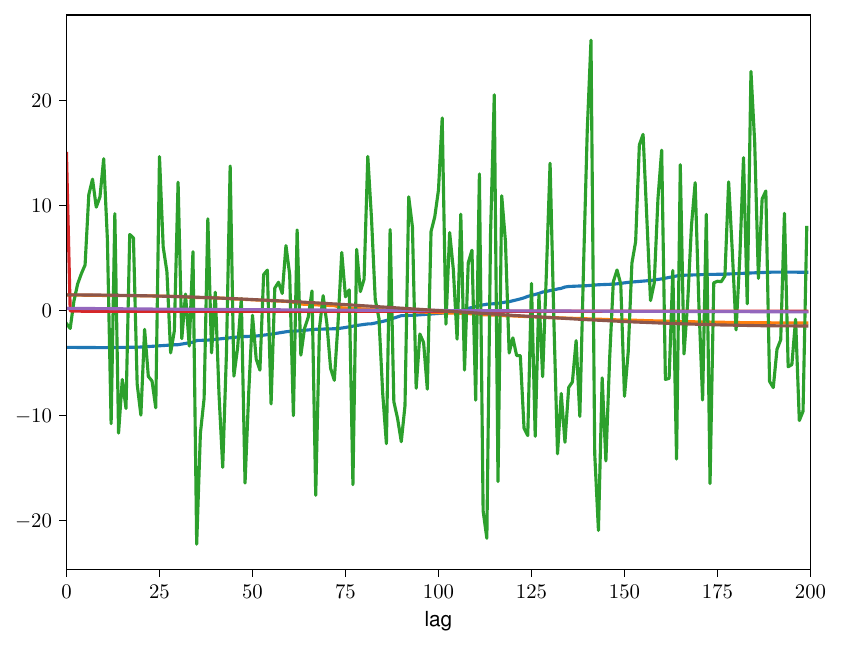}
    \caption{Fashion-MNIST}
  \end{subfigure}
  \begin{subfigure}[t]{0.25\textwidth}
    \centering
    \includegraphics[width=1\linewidth]{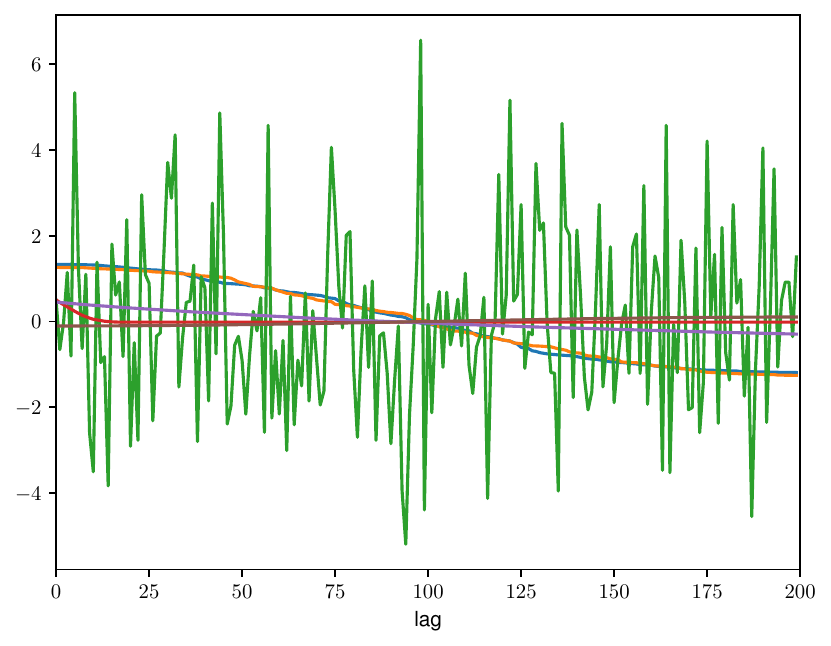}
    \caption{SVHN}
  \end{subfigure}%
  \begin{subfigure}[t]{0.25\textwidth}
    \centering
    \includegraphics[width=1\linewidth]{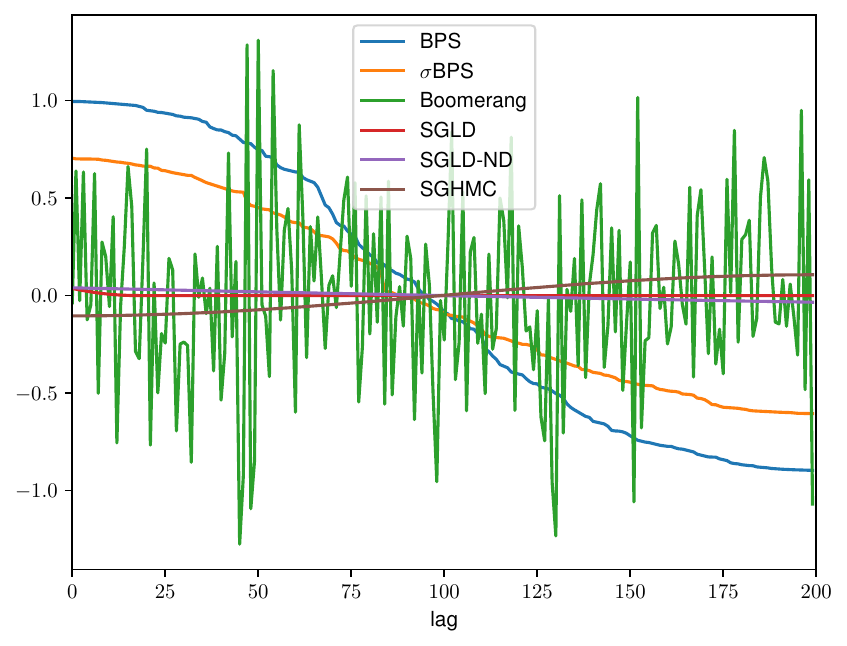}
    \caption{CIFAR-10}
  \end{subfigure}
  \caption{Plots summarising samples from tested samples projected onto first principal component. Top row represents the ACF plot, and the bottom shows the coordinate trace plot for the first principal component. Best viewed on a computer screen.}
  \label{fig:supp_pca_most}
\end{figure*}

\begin{figure*}[!hbt]
  \begin{subfigure}[t]{0.25\textwidth}
    \centering
    \includegraphics[width=1\linewidth]{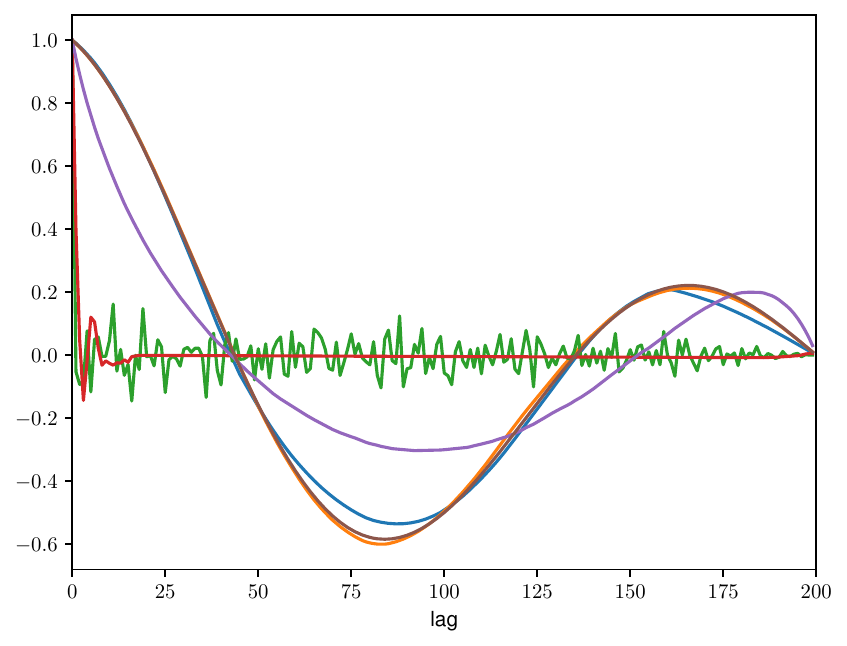}
  \end{subfigure}%
  \begin{subfigure}[t]{0.25\textwidth}
    \centering
    \includegraphics[width=1\linewidth]{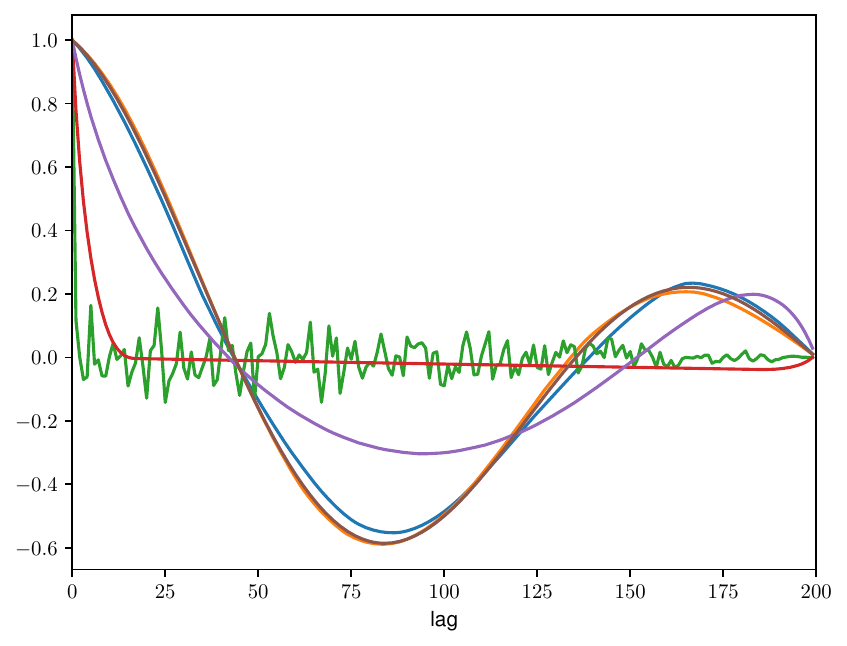}
  \end{subfigure}
  \begin{subfigure}[t]{0.25\textwidth}
    \centering
    \includegraphics[width=1\linewidth]{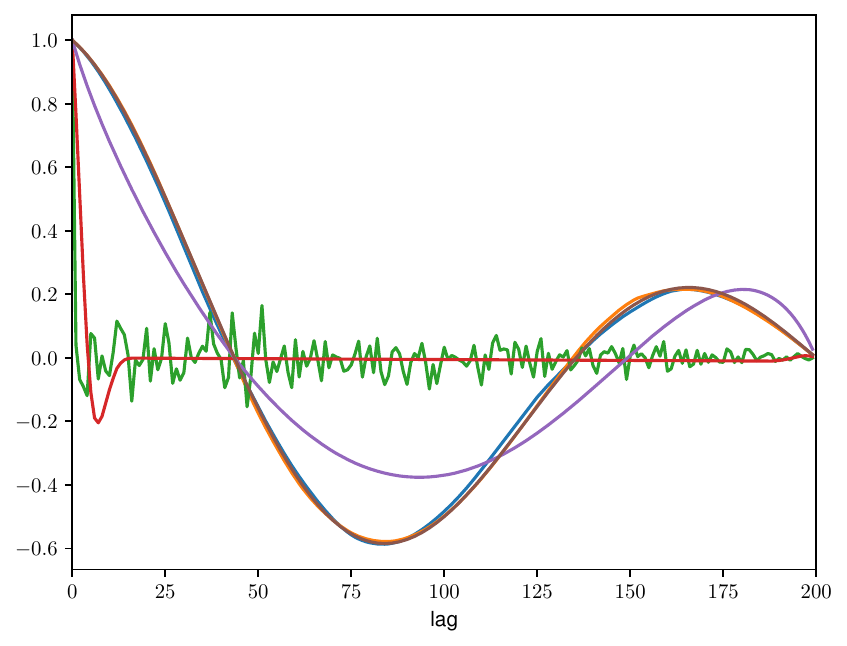}
  \end{subfigure}%
  \begin{subfigure}[t]{0.25\textwidth}
    \centering
    \includegraphics[width=1\linewidth]{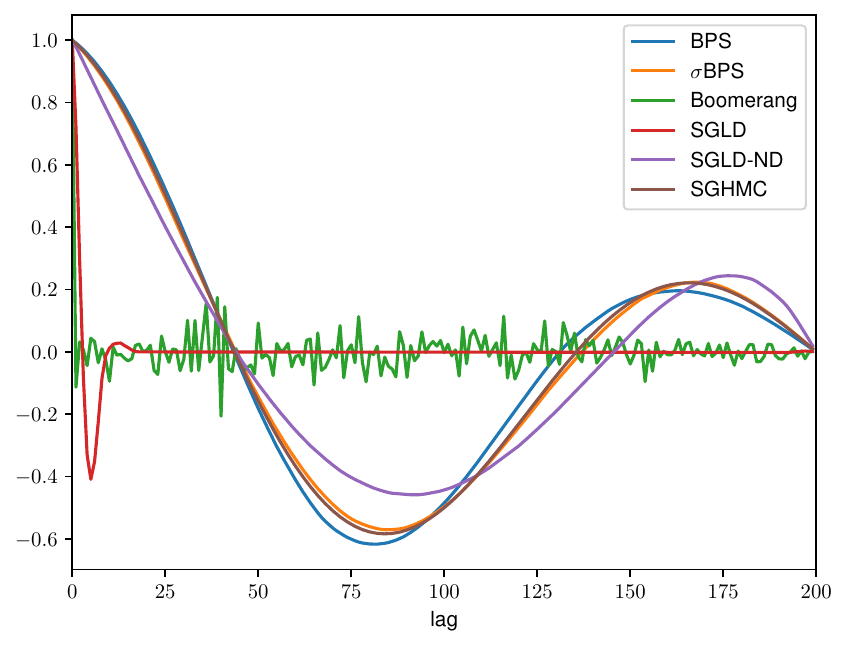}
  \end{subfigure}
  \\
  \begin{subfigure}[t]{0.25\textwidth}
    \centering
    \includegraphics[width=1\linewidth]{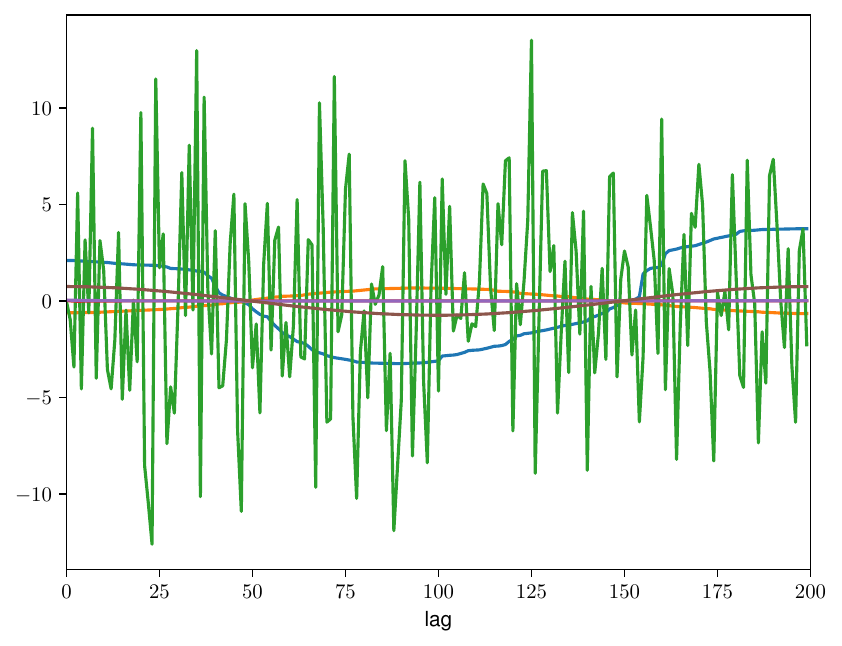}
    \caption{MNIST}
  \end{subfigure}%
  \begin{subfigure}[t]{0.25\textwidth}
    \centering
    \includegraphics[width=1\linewidth]{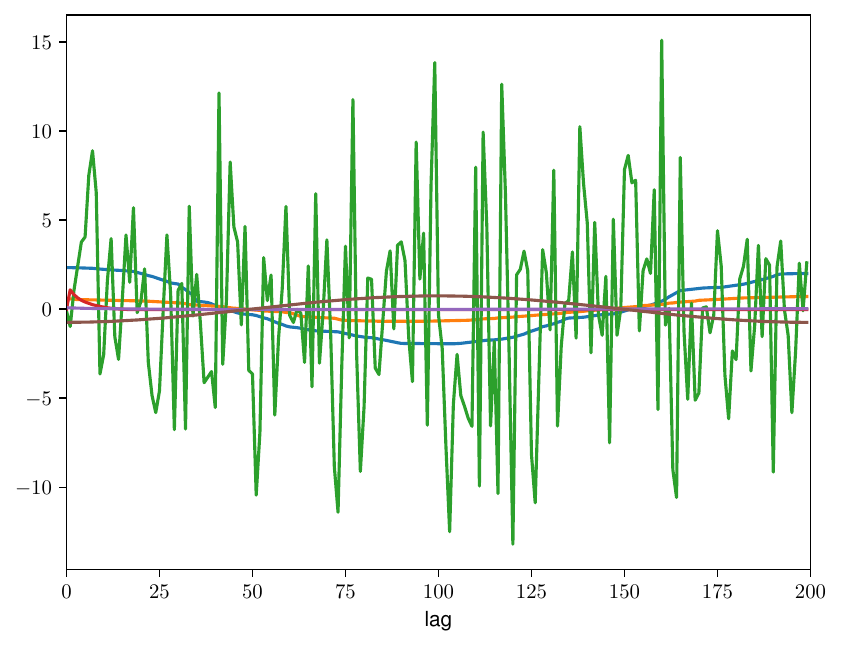}
    \caption{Fashion-MNIST}
  \end{subfigure}
  \begin{subfigure}[t]{0.25\textwidth}
    \centering
    \includegraphics[width=1\linewidth]{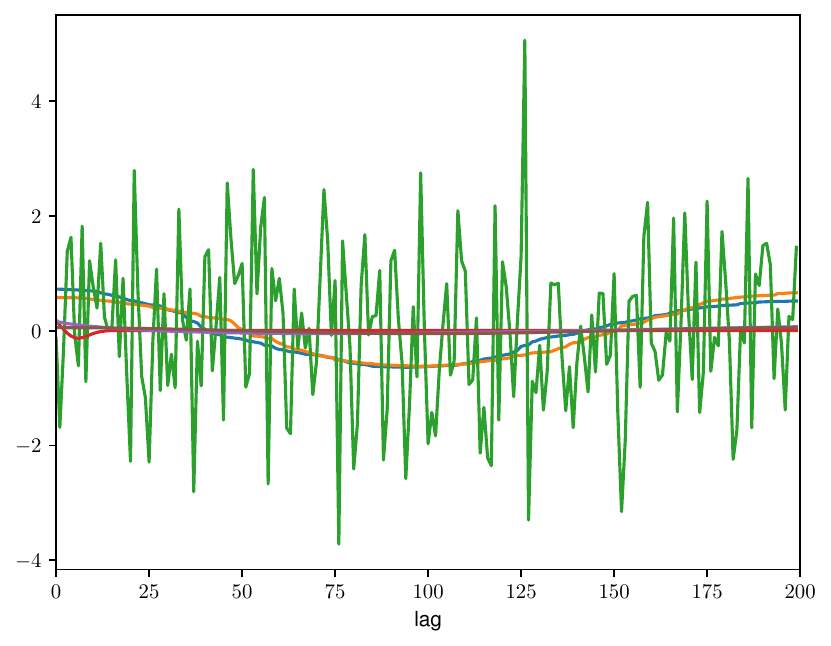}
    \caption{SVHN}
  \end{subfigure}%
  \begin{subfigure}[t]{0.25\textwidth}
    \centering
    \includegraphics[width=1\linewidth]{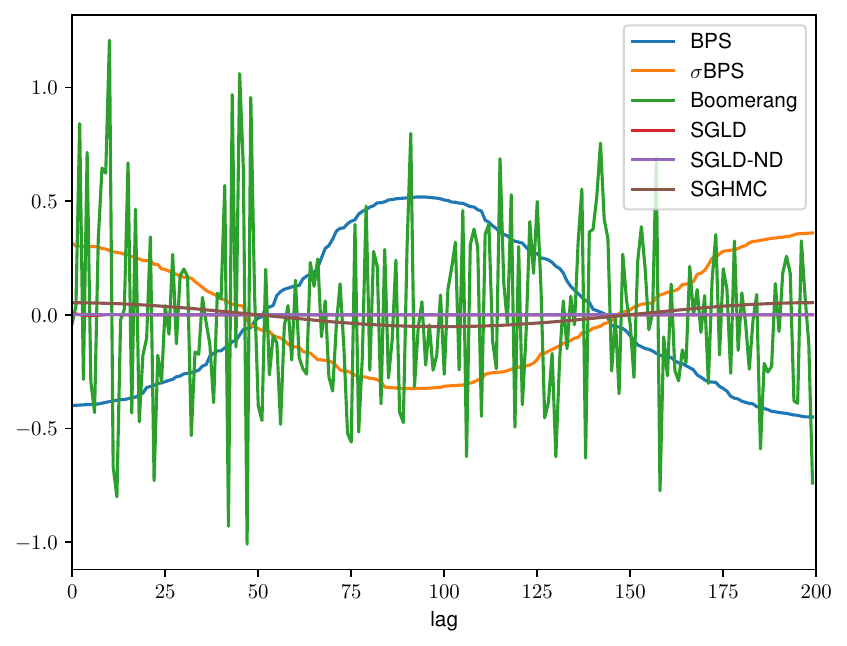}
    \caption{CIFAR-10}
  \end{subfigure}
  \caption{Plots summarising samples from tested samples projected onto second principal component. Top row represents the ACF plot, and the bottom shows the coordinate trace plot for the first principal component. Best viewed on a computer screen.}
  \label{fig:supp_pca_second}
\end{figure*}

\begin{figure*}[!hbt]
  \begin{subfigure}[t]{0.25\textwidth}
    \centering
    \includegraphics[width=1\linewidth]{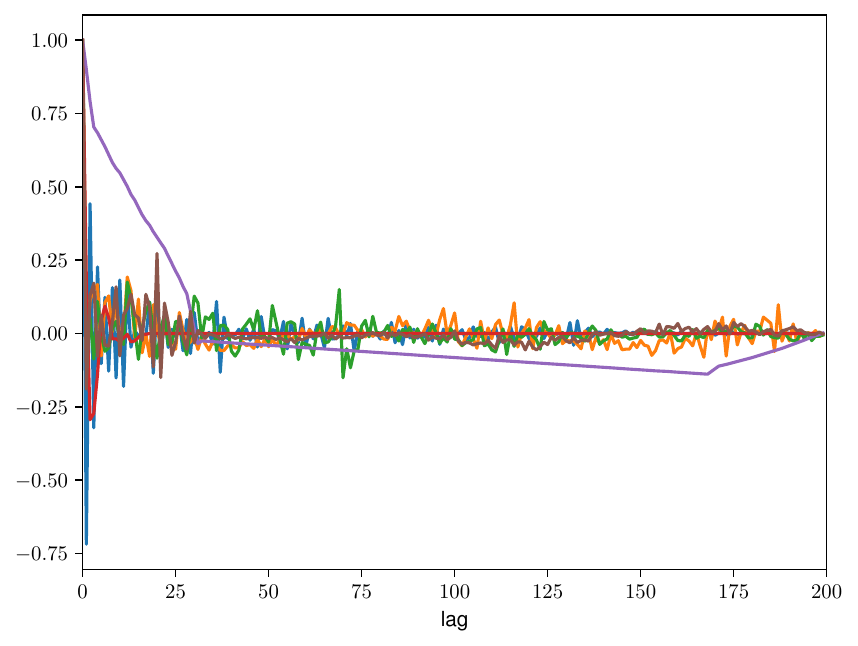}
  \end{subfigure}%
  \begin{subfigure}[t]{0.25\textwidth}
    \centering
    \includegraphics[width=1\linewidth]{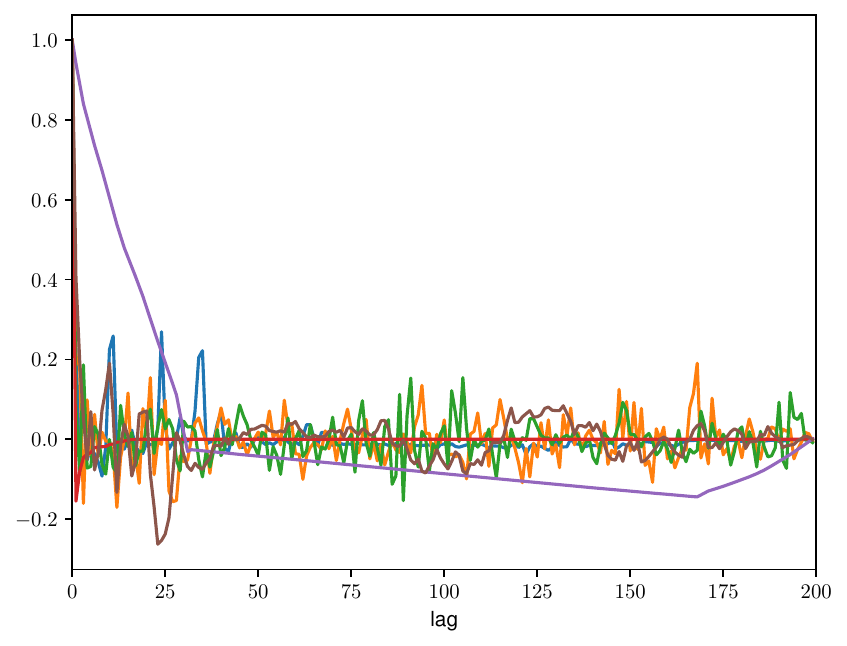}
  \end{subfigure}
  \begin{subfigure}[t]{0.25\textwidth}
    \centering
    \includegraphics[width=1\linewidth]{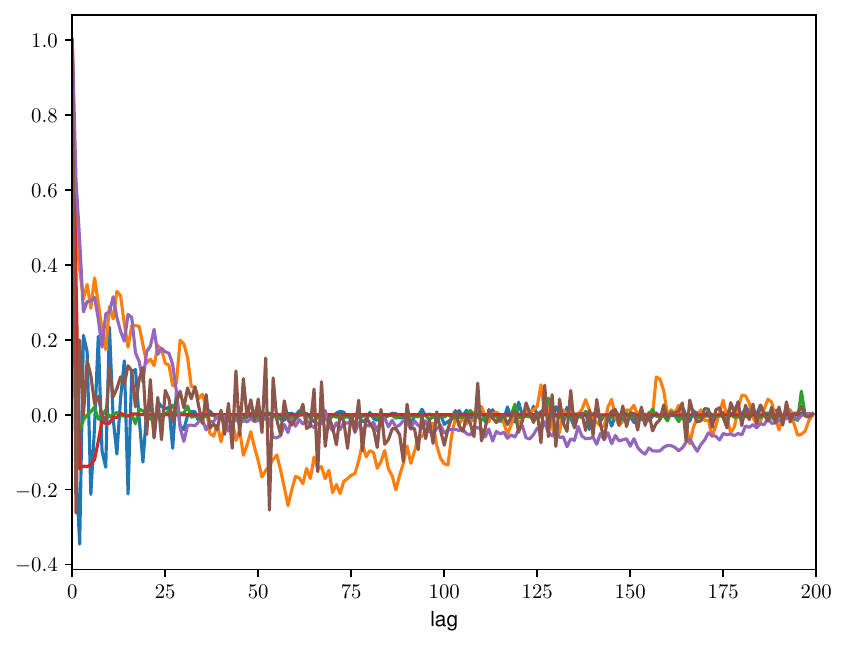}
  \end{subfigure}%
  \begin{subfigure}[t]{0.25\textwidth}
    \centering
    \includegraphics[width=1\linewidth]{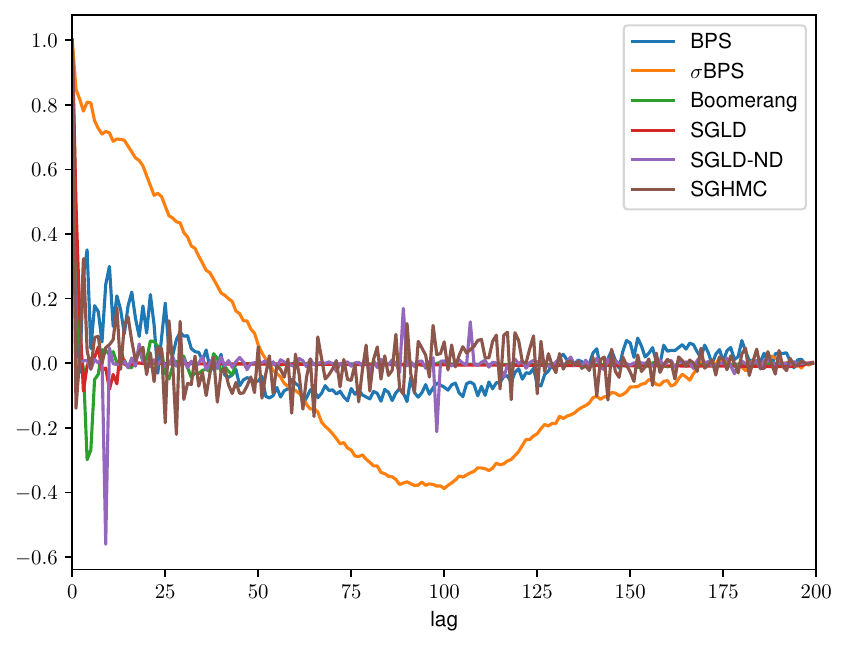}
  \end{subfigure}
  \\
  \begin{subfigure}[t]{0.25\textwidth}
    \centering
    \includegraphics[width=1\linewidth]{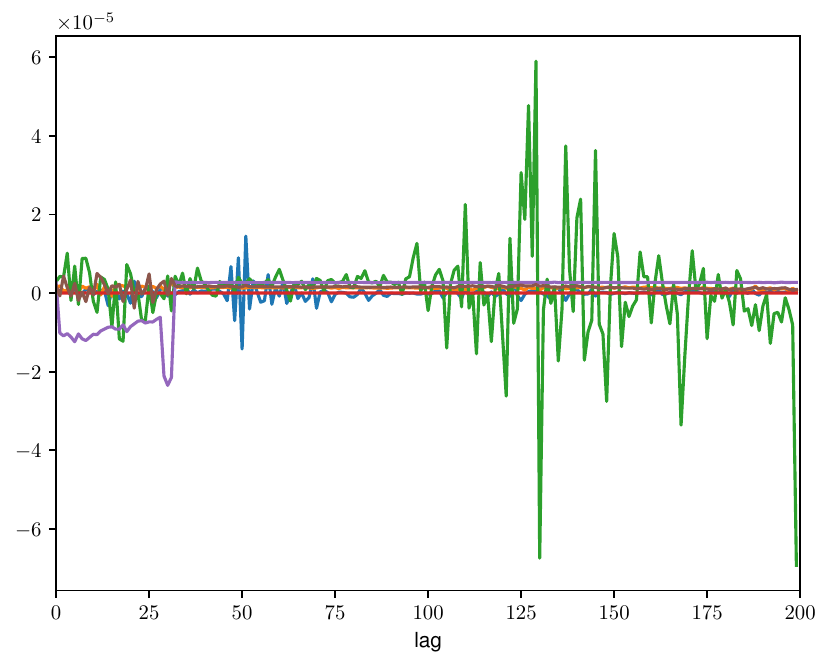}
    \caption{MNIST}
  \end{subfigure}%
  \begin{subfigure}[t]{0.25\textwidth}
    \centering
    \includegraphics[width=1\linewidth]{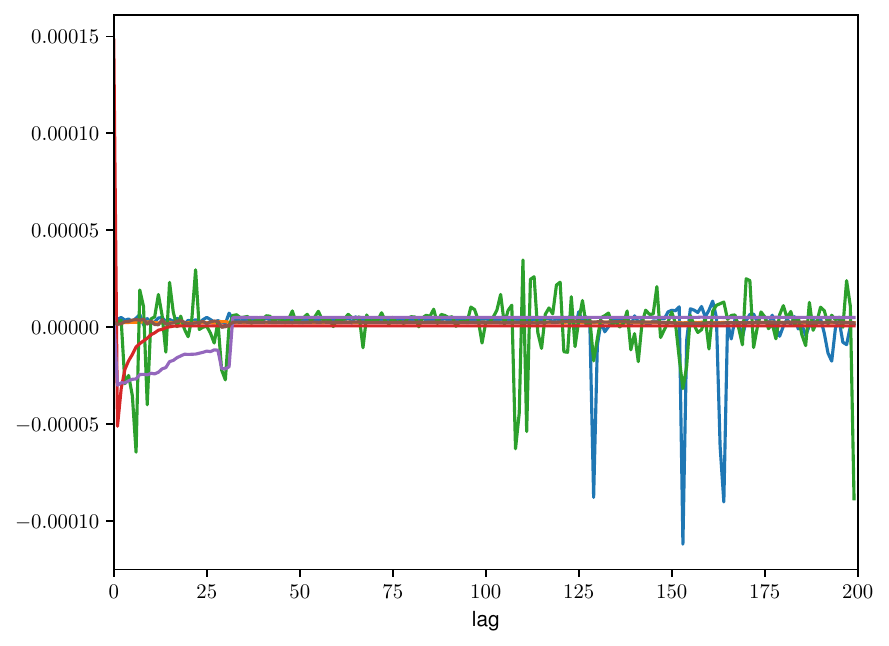}
    \caption{Fashion-MNIST}
  \end{subfigure}
  \begin{subfigure}[t]{0.25\textwidth}
    \centering
    \includegraphics[width=1\linewidth]{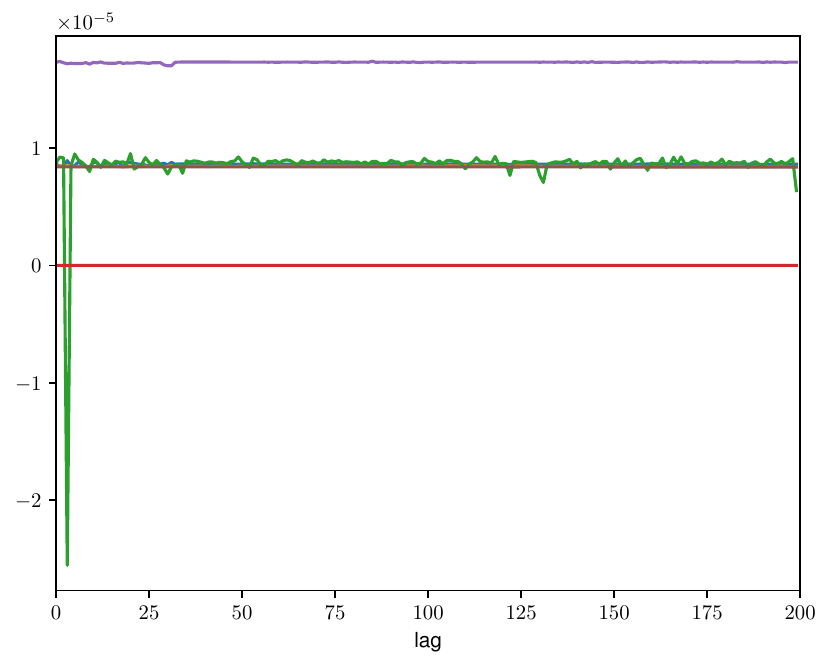}
    \caption{SVHN}
  \end{subfigure}%
  \begin{subfigure}[t]{0.25\textwidth}
    \centering
    \includegraphics[width=1\linewidth]{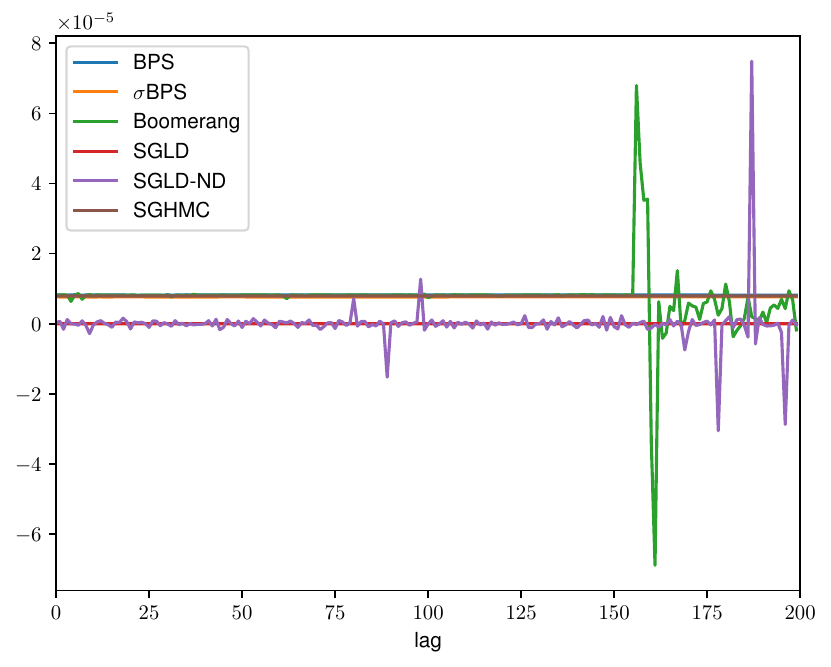}
    \caption{CIFAR-10}
  \end{subfigure}
  \caption{Plots summarising samples from tested samples projected onto last principal component. Top row represents the ACF plot, and the bottom shows the coordinate trace plot for the last principal component. Best viewed on a computer screen.}
  \label{fig:supp_least}
\end{figure*}

\begin{figure}[!h]
  \centering
  \begin{subfigure}[t]{0.45\textwidth}
    \centering \includegraphics[width=1\linewidth]{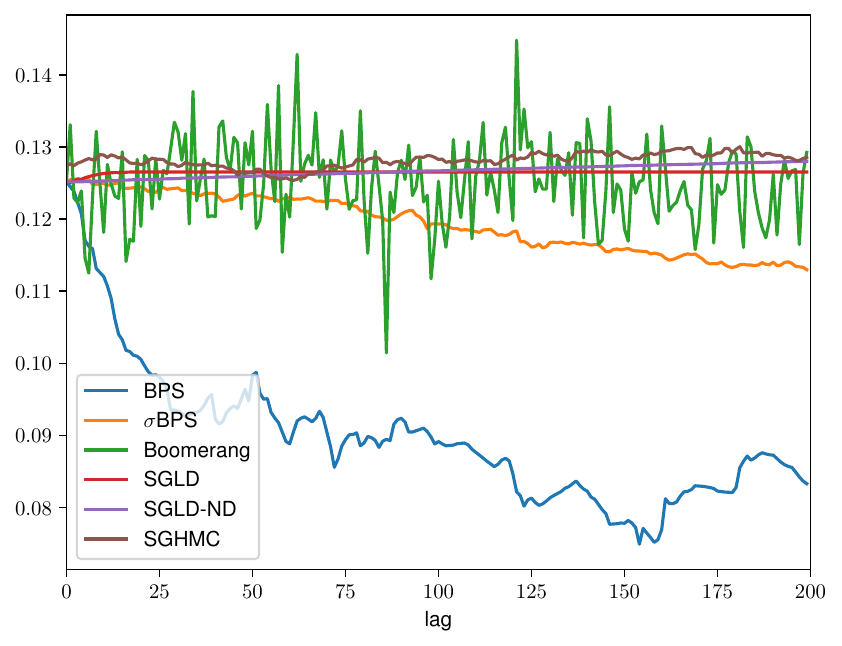}
    \caption{Parameter first layer MNIST}
  \end{subfigure}%
  ~
  \begin{subfigure}[t]{0.45\textwidth}
    \centering \includegraphics[width=1\linewidth]{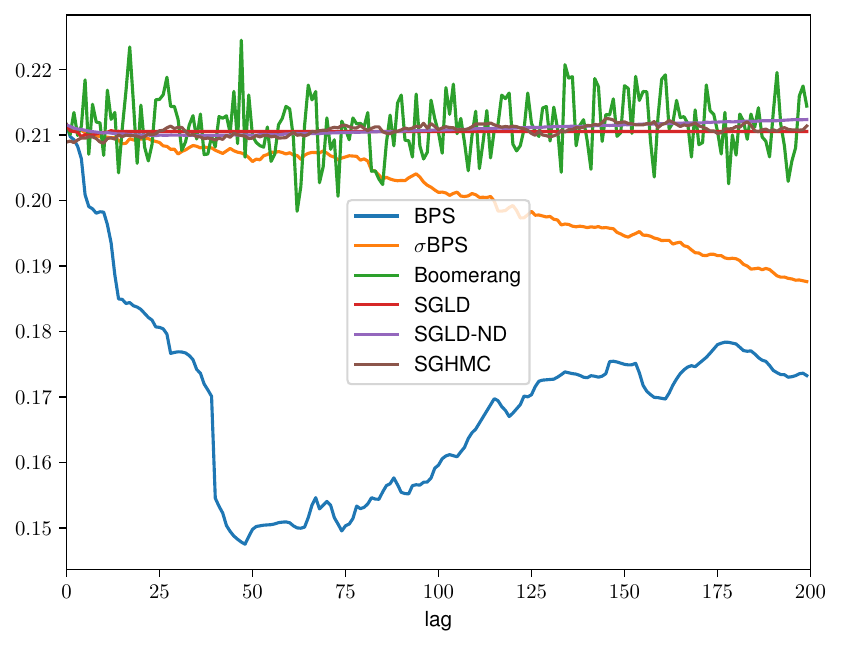}
    \caption{Parameter last layer MNIST}
  \end{subfigure}%
  \\
  \begin{subfigure}[t]{0.45\textwidth}
    \centering \includegraphics[width=1\linewidth]{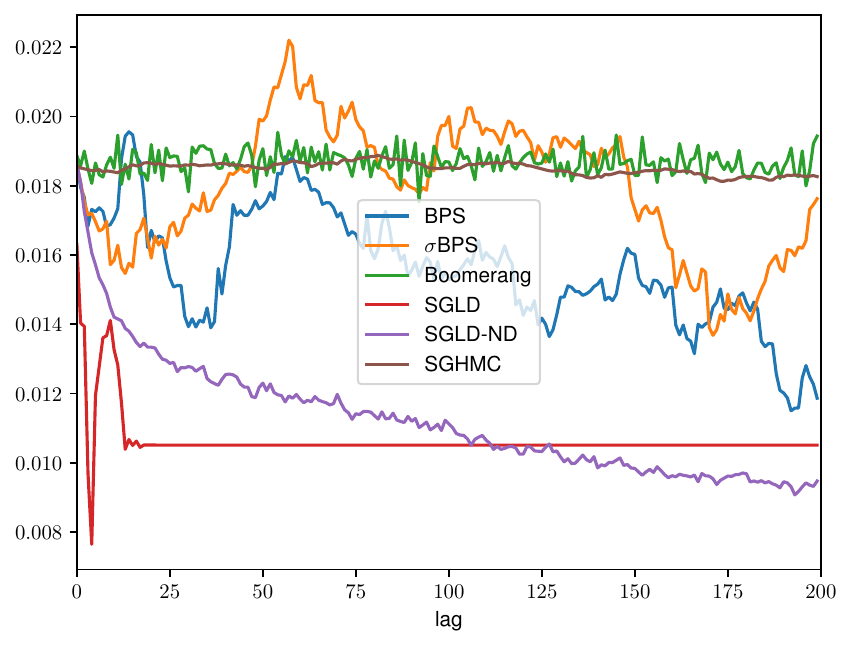}
    \caption{Parameter first layer SVHN}
  \end{subfigure}%
  ~
  \begin{subfigure}[t]{0.45\textwidth}
    \centering \includegraphics[width=1\linewidth]{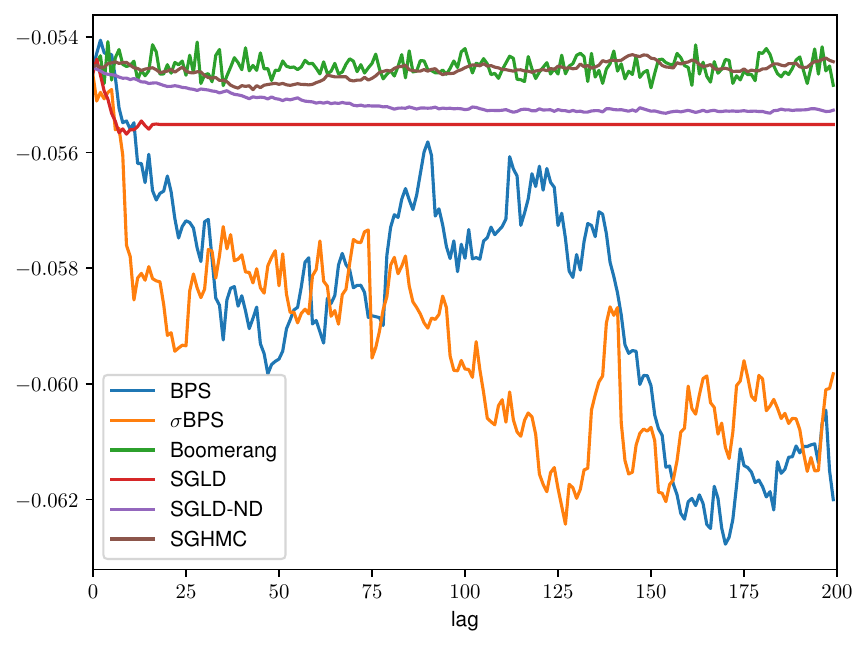}
    \caption{Parameter last layer SVHN}
  \end{subfigure}%
  \caption{Trace plots comparing mixing of SGLD and the Boomerang sampler for individual weight parameters within different networks at different locations.}
  \label{fig:param_trace}
\end{figure}

In Section 5.2, experiments to investigate the mixing capabilities of the PDMP
samplers were conducted using PCA to reduce the dimensionality of the samples
generated from the different samplers for a single network. We extend this
analysis here for all models in Figures \ref{fig:supp_pca_most}, \ref{fig:supp_pca_second} and
\ref{fig:supp_least} for the first, second, and last principal components respectively.
From these figures we can verify that the Boomerang sampler provided the
greatest overall
mixing across the different models and datasets, whilst SGLD consistently converges to a single solution. We
further investigate this here by comparing raw
parameter traces within different parts of the networks used for the MNIST and SVHN datasets. These results are shown
in Figure \ref{fig:param_trace}, and confirms the pathology of SGLD quickly
converging to a single steady-state solution, whilst the PDMP samplers are able
to able to explore the posterior at all stages in the networks. We also note
that the parameter traces for the Boomerang sampler a potential mode-seeking
behaviour, and that the trace plots for the BPS and \sigbps explore a greater
span of the posterior space. This can be attributed to the use of the reference
measure within the Boomerang sampler. Whilst it was able to consistently
generate more independent samples in terms of ESS, the BPS methods may be able
to traverse a greater span of the posterior space. We highlight this as an
direction for future research.
From this, we can verify that SGLD is converging to a steady-state solution,
whilst the Boomerang sampler consistently explores the posterior space and
provide improved mixing. Given the requirement for SGLD to maintain a small
learning rate that approaches 0 to target the posterior
\cite{nagapetyan2017true, brosse2018promises,welling2011bayesian}, these results
are expected. The theoretic ability of SGLD to maintain the posterior as its invariant distribution comes at the expense mixing efficiency.
\section{Sensitivity to Hyper-Parameters}
\label{sec:sensitivity}

\subsection{Sensitivity to Velocity Distribution}
As noted in Section 6, we discuss the sensitivity of these
PDMP Samplers for BNNs with respect to the distribution assigned to the
auxiliary velocity variable. Given that the aim of this velocity variable is to guide
the dynamics of the system to efficiently explore the parameter space, it needs
to be set appropriately. We demonstrate this here through experimentation to
highlight how poorly specified velocity distribution can corrupt inference.
\par
\begin{figure}[!h]
  \centering
  \begin{subfigure}[t]{0.45\textwidth}
    \centering
    \includegraphics[width=1\linewidth]{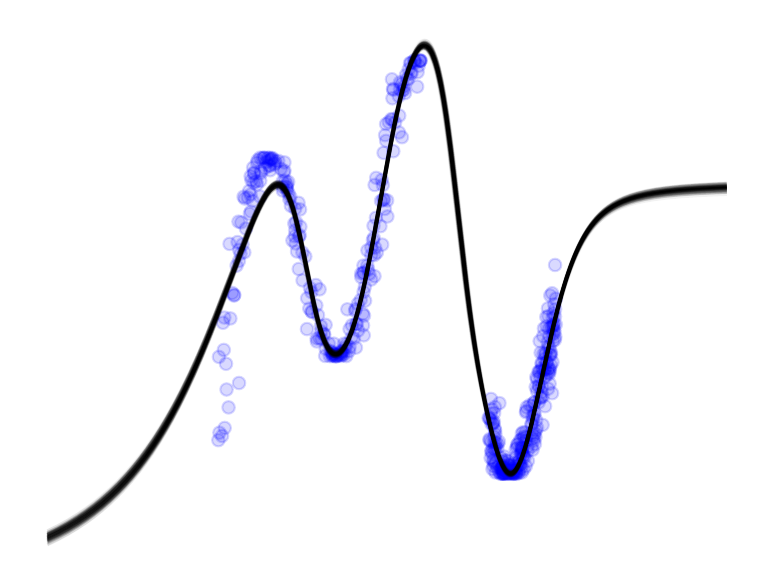}
    \caption{Small velocity distribution}
  \end{subfigure}%
  ~
  \begin{subfigure}[t]{0.45\textwidth}
    \centering
    \includegraphics[width=1\linewidth]{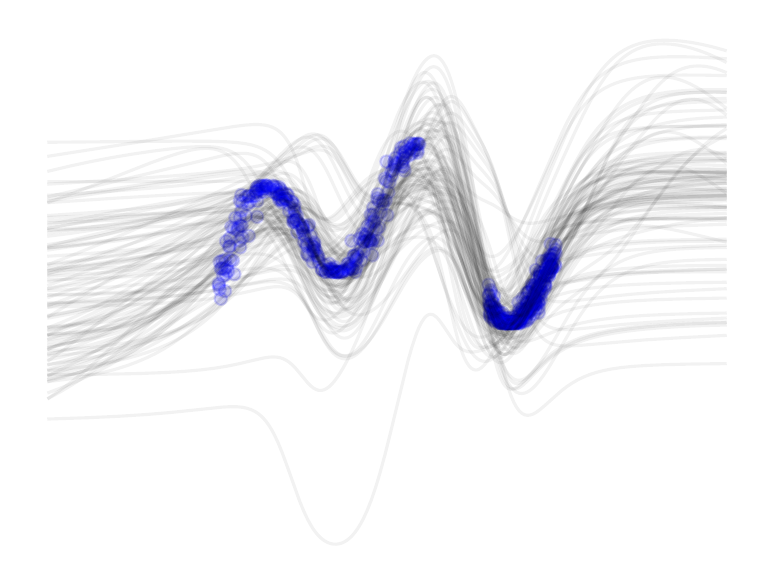}
    \caption{Large velocity distribution}
  \end{subfigure}%
  \caption{Effect of scale in velocity reference measure for PDMP samplers
    applied to BNNs.}
  \label{fig:velocity}
\end{figure}

Figure \ref{fig:velocity} illustrates the predictive distribution for poorly
specified velocity distributions for the Boomerang sampler, though similar
effects are seen amoungst the other PDMP samplers when the variance for the
velocity distribution is incorrectly specified. We see that the scale of the
velocity component proportionately controls the mixing capabilities of the model.
When variance is too low, the sampler is unable to explore beyond the MAP
solution, and when too large the predictive performance can suffer. With better
approximations to the diagonal of Hessian of the negative log-likelihood, the
effects of this may be mitigated for the Boomerang Sampler. We highlight these
behaviours of PDMP samplers applied to BNNs to show the limitations and to
provide insight into the importance of setting these parameters correctly, and
areas of future research.
\subsection{Sensitivity to Refresh Event Rate}
MCMC samplers such as HMC \cite{neal2011mcmc} and NUTS \cite{hoffman2014no} have
step size parameters that can be adjusted and tuned for individual models. With
a small step size, exploration of the posterior can be limited, and if too large
then divergences in the posterior trajectory can be encountered and corrupting
inference \cite{betancourt2017conceptual}. The step size parameter is typically
tuned during a warm-up phase before sampling is commenced to find an optimal
value to maximise exploration and minimise the risk of encountering these
divergences.
\par
The PDMP samplers within here do not have an equivalent parameter that can be
tuned to guide simulation. The trajectory of these samplers is defined solely on
the transition kernel to update velocity parameters and the event rate that
determines when these events occur. We can however yield a similar effect to
adjusting the step size of a traditional MCMC model through our choice of event
rate for our refreshment process $\text{PP}(\lambda_{ref})$.
\par
\begin{figure}[!h]
  \centering
  \begin{subfigure}[t]{0.45\textwidth}
    \centering \includegraphics[width=1\linewidth]{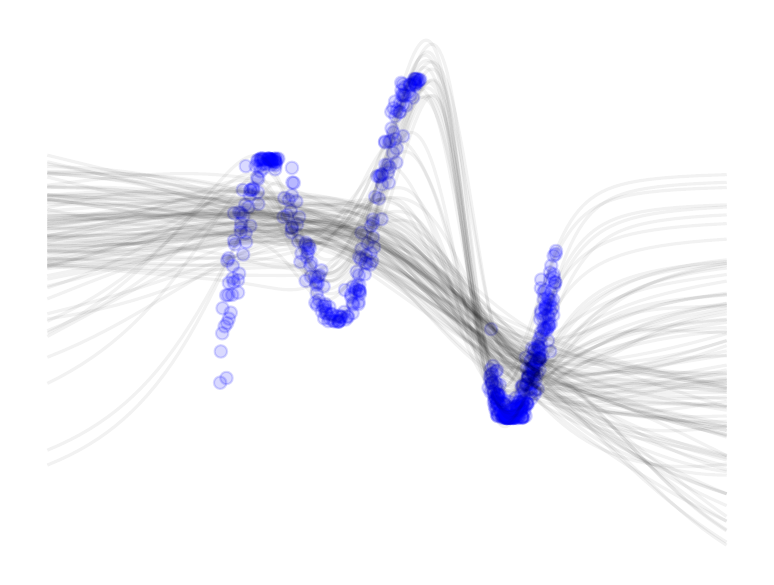}
    \caption{$\lambda_{ref}=0.01$}
  \end{subfigure}%
  ~
  \begin{subfigure}[t]{0.45\textwidth}
    \centering \includegraphics[width=1\linewidth]{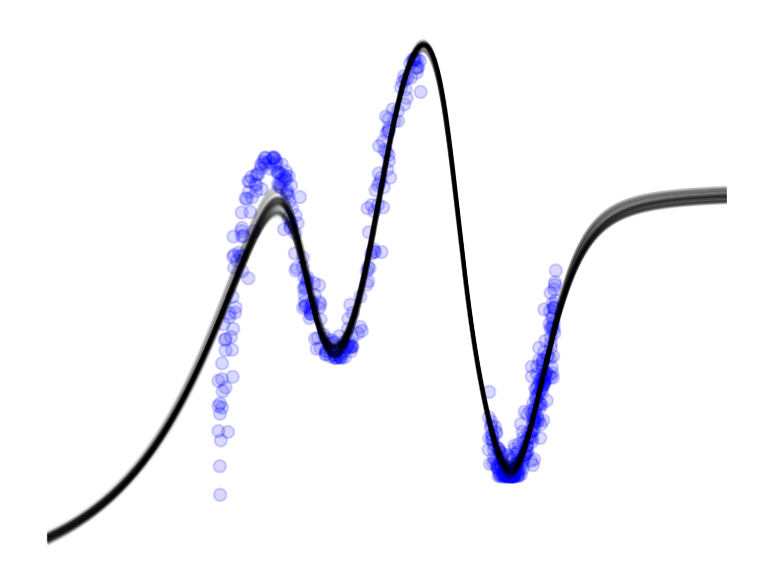}
    \caption{$\lambda_{ref}=10.0$}
  \end{subfigure}%
  \caption{Effect of $\lambda_{ref}$ on PDMP models applied to BNNs. Shown here is the predictive distribution found with the BPS using the proposed event rate thinning method.}
  \label{fig:refresh}
\end{figure}

Recall from Section 2.4 that the final event time is given by,
\begin{equation}
  \label{eq:1}
  \tau_{event} = \text{min}(\tau, \tau_{ref})
\end{equation}
Where \(\tau \sim \text{PP}(\lambda(\myw(t), \myv(t)))\),
and $\tau_{ref} \sim \text{PP}(\lambda_{ref})$. Setting the value
for $\tau_{ref}$ can implicitly control the level of exploration within our
samplers. For large $\lambda_{ref}$, we will encounter smaller proposed refresh
times and thus will refresh more frequently. Similarly, for
larger $\lambda_{ref}$, our sampled refresh times will be larger,
and $\tau_{event}$ will equal $\tau$ more frequently, and further exploration of
the posterior space with these dynamics will be possible. We illustrate this in
Figure \ref{fig:refresh}, where we show the effects for large and smaller values
of $\lambda_{ref}$.

We can see that the refresh rate can have a considerable impact on the inference quality of our model. With $\lambda_{ref}$ too large, our exploration is limited and we perform excessive refreshments instead of accepting those provided by the PDMP kernel. When is too small, we can accept larger event times as specified by the PDMP sampler and can diverge away from meaningful inferences.

\section{Summary of models used}
\subsection{Regression and Binary Classification Models for Synthetic Data}
\label{sec:synth}
Regression models used within this work consist of fully-connected networks with two hidden layers, each with 25 and  10 units respectively. Tanh non-linear activations are applied after each hidden layer, and a Normal likelihood with a variance of $\sigma^{2} = 0.01$ is used. MAP estimates for these networks are found with 10,000 iterations using the Adam optimiser \cite{kingma2015adam}, with each sampler initialised from the same MAP estimate.
\par
For binary classification models, the networks consist of a fully-connected network with three hidden layers, each with 100 units. ReLU non-linear activations are applied within the network, and a Bernoulli likelihood is used. Similar to the regression tasks, MAP estimate is found with Adam.
\subsection{Additonal UCI-Dataset Results}
We provide here additional results on datasets from the UCI
repository\cite{ucidatasets}. For each dataset, a simple MLP network with a
three hidden layers with 512, 256, and 128 hidden units is used, along with a ReLU
activation. MAP estimates are found similar to \ref{sec:synth}, followed by
1,000 samples generated by each method. Each experiment is run 5 times
with mean results and standard deviations reported. We further include ESS
as measured on from the smallest principle component of samples. Results from these
experiments reflect that seen in Section 5.1.1; where the PDMP methods
show comparable or improved predictive performance. The Boomerang sampler is able to consistently outperform
other sampling methods in terms of ESS, with other samplers only able to match sample efficiency
for the smallest principal components where exploration is smallest.
\begin{table}[h!]
  \caption{Results on UCI-Naval Dataset}
  \label{tab:uci-naval}
  \begin{center}
    \begin{small}
      \scalebox{0.85}{
        \begin{tabular}{l l l l l l}
          \toprule
          {\bfseries Inference} & {\bfseries NLL}
                                & {\bfseries RMSE}
                                & {\bfseries ESS-First}
                                & {\bfseries ESS-Second}
                                & {\bfseries ESS-Last}
          \\
          \midrule\midrule[.1em]
          BPS                   & \textbf{0.96 $\pm$ 0.00} & 2.76 $\pm$ 0.11          & 2.71 $\pm$ 0.02              & 4.09 $\pm$ 0.05             &
          412.68 $\pm$ 536.28                                                                                                                                                    \\
          \sigbps               & \textbf{0.96 $\pm$ 0.01} & 2.73 $\pm$ 0.12          & 2.70 $\pm$ 0.01              & 4.14 $\pm$ 0.08             & \textbf{1000.00 $\pm$ 0.00} \\
          Boomerang             & \textbf{0.96 $\pm$ 0.01} & 2.77 $\pm$ 0.13          & \textbf{901.07 $\pm$ 105.03} & \textbf{937.24 $\pm$ 72.62} &
          871.02 $\pm$ 288.40                                                                                                                                                    \\
          SGLD                  & \textbf{0.96 $\pm$ 0.00} & \textbf{2.68 $\pm$ 0.01} & 2.89 $\pm$ 0.00              & 4.45 $\pm$ 0.05             & \textbf{1000.00 $\pm$ 0.00} \\
          SGHMC                 & \textbf{0.96 $\pm$ 0.00} & 2.69 $\pm$ 0.05          & 2.71 $\pm$ 0.00              & 4.10 $\pm$ 0.00             & \textbf{1000.00 $\pm$ 0.00} \\
          \bottomrule
        \end{tabular}
      }
    \end{small}
  \end{center}
\end{table}

\begin{table}[h!]
  \caption{Results on UCI Energy Dataset}
  \label{tab:uci-energy}
  \begin{center}
    \begin{small}
      \scalebox{0.85}{
        \begin{tabular}{l l l l l l}
          \toprule
          {\bfseries Inference} & {\bfseries NLL}
                                & {\bfseries RMSE}
                                & {\bfseries ESS-First}
                                & {\bfseries ESS-Second}
                                & {\bfseries ESS-Last}
          \\
          \midrule\midrule[.1em]
          BPS                   & 0.95 $\pm$ 0.00          & \textbf{0.00 $\pm$ 0.00} & 2.71 $\pm$ 0.02              & 4.13 $\pm$ 0.07              & 392.42 $\pm$ 529.69         \\
          \sigbps               & 0.95 $\pm$ 0.00          & \textbf{0.00 $\pm$ 0.00} & 2.73 $\pm$ 0.03              & 4.15 $\pm$ 0.06              & 800.66 $\pm$ 445.74         \\
          Boomerang             & 0.97 $\pm$ 0.02          & \textbf{0.00 $\pm$ 0.00} & \textbf{837.54 $\pm$ 181.39} & \textbf{824.77 $\pm$ 139.27} & 790.48 $\pm$ 276.02         \\
          SGLD                  & 1.60 $\pm$ 0.19          & 0.01 $\pm$ 0.00          & 2.96 $\pm$ 0.01              & 5.74 $\pm$ 0.14              & \textbf{1000.00 $\pm$ 0.00} \\
          SGHMC                 & \textbf{0.92 $\pm$ 0.00} & \textbf{0.00 $\pm$ 0.00} & 2.71 $\pm$ 0.00              & 4.10 $\pm$ 0.00              & 195.98 $\pm$ 106.67         \\
          \bottomrule
        \end{tabular}
      }
    \end{small}
  \end{center}
\end{table}

\begin{table}[h!]
  \caption{Results on UCI Yacht Dataset}
  \label{tab:uci-yacht}
  \begin{center}
    \begin{small}
      \scalebox{0.85}{
        \begin{tabular}{l l l l l l}
          \toprule
          {\bfseries Inference} & {\bfseries NLL}
                                & {\bfseries RMSE}
                                & {\bfseries ESS-First}
                                & {\bfseries ESS-Second}
                                & {\bfseries ESS-Last}
          \\
          \midrule\midrule[.1em]
          BPS                   & \textbf{ 0.92 $\pm$ 0.00} & 0.97 $\pm$ 0.31          & 2.75 $\pm$ 0.05              & 4.11 $\pm$ 0.11              & 15.27 $\pm$ 10.84           \\
          \sigbps               & \textbf{0.92 $\pm$ 0.00}  & \textbf{0.75 $\pm$ 0.01} & 2.72 $\pm$ 0.03              & 4.24 $\pm$ 0.24              & 1000.00 $\pm$ 0.00          \\
          Boomerang             & \textbf{0.92 $\pm$ 0.00}  & 1.00 $\pm$ 0.33          & \textbf{732.04 $\pm$ 276.22} & \textbf{754.01 $\pm$ 156.27} & 777.24 $\pm$ 142.72         \\
          SGLD                  & \textbf{0.92 $\pm$ 0.00}  & 0.85 $\pm$ 0.10          & 3.07 $\pm$ 0.00              & 4.96 $\pm$ 0.05              & \textbf{1000.00 $\pm$ 0.00} \\
          SGHMC                 & \textbf{0.92 $\pm$ 0.00}  & 1.01 $\pm$ 0.13          & 2.71 $\pm$ 0.00              & 4.10 $\pm$ 0.00              & 224.12 $\pm$ 62.13          \\
          \bottomrule
        \end{tabular}
      }
    \end{small}
  \end{center}
\end{table}

\begin{table}[h!]
  \caption{Results on UCI Concrete Dataset}
  \label{tab:uci-concrete}
  \begin{center}
    \begin{small}
      \scalebox{0.85}{
        \begin{tabular}{l l l l l l}
          \toprule
          {\bfseries Inference} & {\bfseries NLL}
                                & {\bfseries RMSE}
                                & {\bfseries ESS-First}
                                & {\bfseries ESS-Second}
                                & {\bfseries ESS-Last}
          \\
          \midrule\midrule[.1em]
          BPS                   & \textbf{0.93 $\pm$ 0.00} & 1.85 $\pm$ 0.03          & 2.72 $\pm$ 0.03              & 4.10 $\pm$ 0.06              & 592.92 $\pm$ 531.11         \\
          \sigbps               & \textbf{0.93 $\pm$ 0.00} & \textbf{1.84 $\pm$ 0.03} & 2.72 $\pm$ 0.01              & 4.15 $\pm$ 0.07              & 1000.00 $\pm$ 0.00          \\
          Boomerang             & \textbf{0.93 $\pm$ 0.00} & 1.92 $\pm$ 0.05          & \textbf{888.04 $\pm$ 250.34} & \textbf{821.95 $\pm$ 336.23} & 757.61 $\pm$ 235.83         \\
          SGLD                  & \textbf{0.93 $\pm$ 0.00} & 1.87 $\pm$ 0.00          & 3.01 $\pm$ 0.00              & 4.97 $\pm$ 0.02              & \textbf{1000.00 $\pm$ 0.00} \\
          SGHMC                 & \textbf{0.93 $\pm$ 0.00} & 1.97 $\pm$ 0.28          & 2.71 $\pm$ 0.00              & 4.10 $\pm$ 0.00              & 245.39 $\pm$ 49.05          \\
          \bottomrule
        \end{tabular}
      }
    \end{small}
  \end{center}
\end{table}

\subsection{Convolutional Models}
For the \sigbps, an initial warm-up stage is again required, which
is identical to that in Section 5.1. For MNIST and Fashion-MNIST,
a batch size of 1024 is used, whilst a batch size of 512 is used for the
remaining models. MAP estimates for MNIST and Fashion-MNIST datasets were found
with the Adam optimiser \cite{kingma2015adam} for 10,000 iterations. SVHN and
CIFAR-10 used SGD with momentum of 0.1 and 0.9 respectively for 25,000
iterations, where for CIFAR-100, required 128,000 iterations and a momentum of 0.2.
\\
With the potential sensitivities to both refreshment rates and choice of velocity distribution $\Phi(\myv)$ identified in \ref{sec:sensitivity}, we deem it important to report the values used for fitting each model. We report these values in Table \ref{tab:conv_results} alongside full predictive performance measurements and sample efficiency metrics. Within Table \ref{tab:conv_results}, we represent the choice of velocity distribution with the $\gamma$ parameter. For \Gls{bps}, $\gamma$ describes the standard deviation of the velocity distribution chosen such that,
\begin{equation}
  \label{eq:2}
  \Phi(\myv) = \mathcal{N}(0, \gamma^{2}).
\end{equation}
Similarly, an initial velocity distribution is set for the \sigbps during
initial warmup phase, afterwards the velocity distribution is set
to $\mathcal{N}(0, \sigma^{2})$.  For the Boomerang sampler, $\gamma$ represents the scaling factor as found in
Equation 6 from the body of the paper.
\begin{table}[t!]
  \caption{Summary of hyperparameters used for samplers within this work.}
  \label{tab:conv_results}
  \begin{center}
    \scalebox{0.85}{
      \begin{tabular}{l l l l l}
        \toprule
        {\bfseries Dataset } & {\bfseries Inference} & {\bfseries $\lambda_{ref}$} & {\bfseries $\gamma$} & {\bfseries Time}
        \\
        \midrule\midrule[.1em]
        \multirow{5}{1.5cm}{MNIST}
                             & SGD                   & -                           & -                    & 74               \\
                             & SGLD                  & -                           & -                    & 87               \\
                             & SGLD-ND               & -                           & -                    & 87               \\
                             & BPS                   & 1.0                         & 0.001                & 145              \\
                             & \sigbps               & 1.0                         & 0.0005               & 197              \\
                             & Boomerang             & 1.0                         & 0.1                  & 151              \\
        \midrule
        \multirow{5}{1.5cm}{Fashion-MNIST}
                             & SGD                   & -                           & -                    & 74               \\
                             & SGLD                  & -                           & -                    & 87               \\
                             & SGLD-ND               & -                           & -                    & 87               \\
                             & BPS                   & 1.0                         & 0.001                & 144              \\
                             & \sigbps               & 0.1                         & 0.001                & 192              \\
                             & Boomerang             & 1.0                         & 0.1                  & 156              \\
        \midrule
        \multirow{5}{1.5cm}{SVHN}
                             & SGD                   & -                           & -                    & 3465             \\
                             & SGLD                  & -                           & -                    & 3653             \\
                             & SGLD-ND               & -                           & -                    & 3653             \\
                             & BPS                   & 0.5                         & 0.00005              & 4125             \\
                             & \sigbps               & 0.5                         & 0.00005              & 4535             \\
                             & Boomerang             & 1.0                         & 0.1                  & 4375             \\
        \midrule
        \multirow{5}{1.5cm}{CIFAR 10}
                             & SGD                   & -                           & -                    & 4905             \\
                             & SGLD                  & -                           & -                    & 5075             \\
                             & SGLD-ND               & -                           & -                    & 5074             \\
                             & BPS                   & 0.5                         & 0.00005              & 5614             \\
                             & \sigbps               & 1.0                         & 0.00005              & 6053             \\
                             & Boomerang             & 0.01                        & 0.01                 & 5868             \\
        \midrule
        \multirow{5}{1.5cm}{CIFAR 100}
                             & SGD                   & -                           & -                    & 9811             \\
                             & SGLD                  & -                           & -                    & 9985             \\
                             & SGLD-ND               & -                           & -                    & 9985             \\
                             & BPS                   & 0.70                        & 0.00005              & 10478            \\
                             & \sigbps               & 1.5                         & 0.000025             & 10808            \\
                             & Boomerang             & 2.0                         & 0.01                 & 10783            \\
        \bottomrule
      \end{tabular}
    }
  \end{center}
\end{table}
\section{How well are we really exploring the posterior?}
In Radford Neals influential thesis \cite{neal2012bayesian}, he states that
``Bayesian neural network users may have difficulty claiming with a straight
face that their models and priors are selected because they are just what is
needed to capture their prior beliefs about the problem'' \footnote{Although
  much important work has been conducted to establish suitable priors and model
  design \cite{hafner2018reliable, sun2019functional,
    vladimirova2019understanding}, this statement largely remains true today.}.
In a similar vein, we would state that any Bayesian neural network user would
have a difficult time honestly saying their inference strategy has sufficiently
explored the posterior, including the work proposed here. Previous research has
investigated gold-standard MCMC methods for larger networks
\cite{izmailov2021bayesian}, though were unable to obtain a sufficient number of
samples to maintain confidence in the levels of exploration. Although the
metrics in the previous section may show sufficient results for a machine
learning application, from a statistical perspective we need to further
investigate the quality of our inference to justify whether we have satisfied
our goal of sampling from the posterior distribution.
\par
Previous papers for PDMP methods for MCMC have shown favourable performance in
terms of mixing and sampling efficiency \cite{bouchard2018bouncy,
  bierkens2019zig, wu2017generalized, bierkens2020boomerang} and has similarly
outperformed methods such as \Gls{sgld}. Most studies have been restricted to
well-defined models; where prior information can be suitably provided and
sufficient prior studies with gold standard methods such as HMC and NUTS have
confirmed the general geometry of the posteriors in question and the existence
of a central limit theorem. Inference in BNNs is challenged by a posterior with
strong multi-modality, making exploration of any sampler more difficult. This is
further challenged by the comparatively large dimensional space over which we
need to explore. The favourable Gaussian Process and functional properties seen
by networks with infinite width \cite{neal2012bayesian} encourage the use of
large models, whilst also narrowing the typical set in which we wish to explore
\cite{betancourt2017conceptual}.
\par
Another limitation is the computational complexity added with sampling-based
schemes. This complexity not only includes the cost of sampling, but the
increase in memory consumption. The popular ResNet-50 model contains more than
23 million parameters. To perform inference on ImageNet with a ResNet50 model
using a mini-batch size of 100 samples, more than 10 thousand samples would be
needed to iterate over the entire data set of over one million
images\footnotetext{The commonly used variant of ImageNet is from the 2012 Large
  Scale Visual Recognition Challenge, which contains 1,281,167 samples
  \cite{russakovsky2015imagenet}}. With single-point precision, these samples
for a single complete iteration of the dataset would require more than 9.2GB of
memory. These constraints currently limit the applicability of such methods, as
evaluating predictive posteriors will require a large number of samples and many
operations to read sampled values from non-volatile storage.
\par
These limitations offer insights into areas of future research relating to
sampling schemes for BNNs. The geometry of the joint posterior distribution
could be improved by investigating non-local methods for preconditioning the
gradients, similar to that done in Riemannian HMC \cite{girolami2011riemann}. As
seen in this work through the efficacy of the Boomerang sampler, exploitation of
this geometry can considerably improve exploration of the posterior space.
Finally and perhaps most importantly, bespoke design of model
architecture that respects the data and includes priors that appropriately
reflect domain expertise could yield posteriors that are more easily traversed
and explored.

\section{In and Out of Distribution Data}
\begin{figure}[!h]
  \centering
  \begin{subfigure}[t]{0.2\textwidth}
    \centering
    \includegraphics[width=1\linewidth]{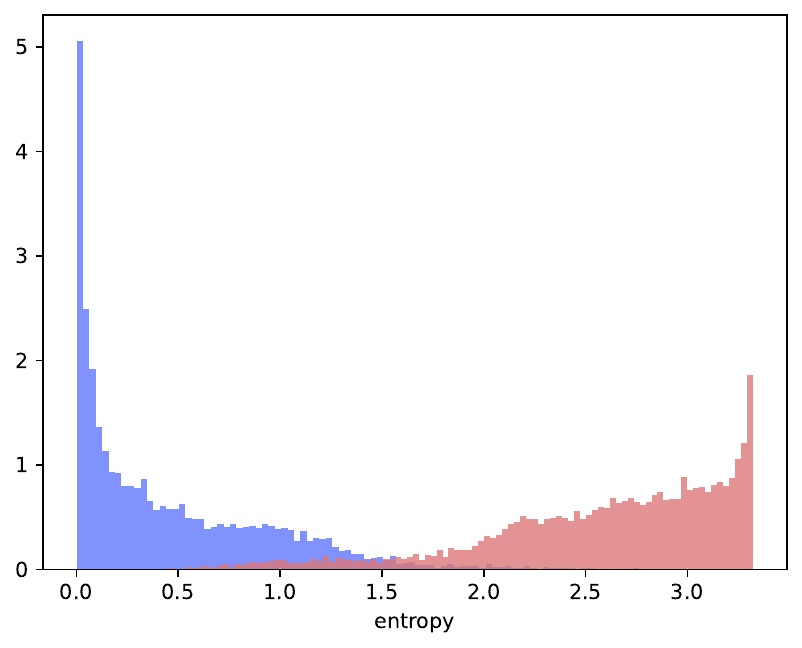}
  \end{subfigure}%
  ~
  \begin{subfigure}[t]{0.2\textwidth}
    \centering
    \includegraphics[width=1\linewidth]{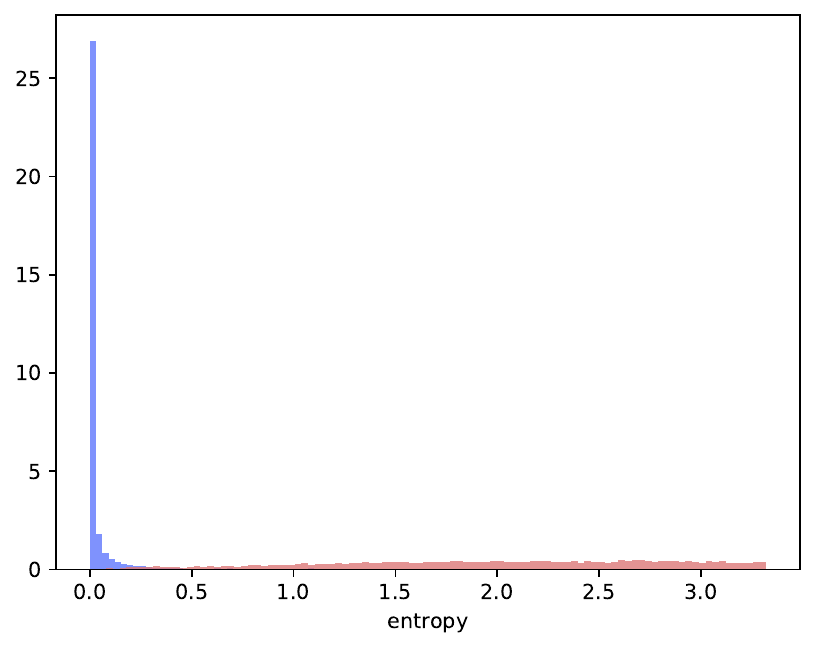}
  \end{subfigure}%
  ~
  \begin{subfigure}[t]{0.2\textwidth}
    \centering
    \includegraphics[width=1\linewidth]{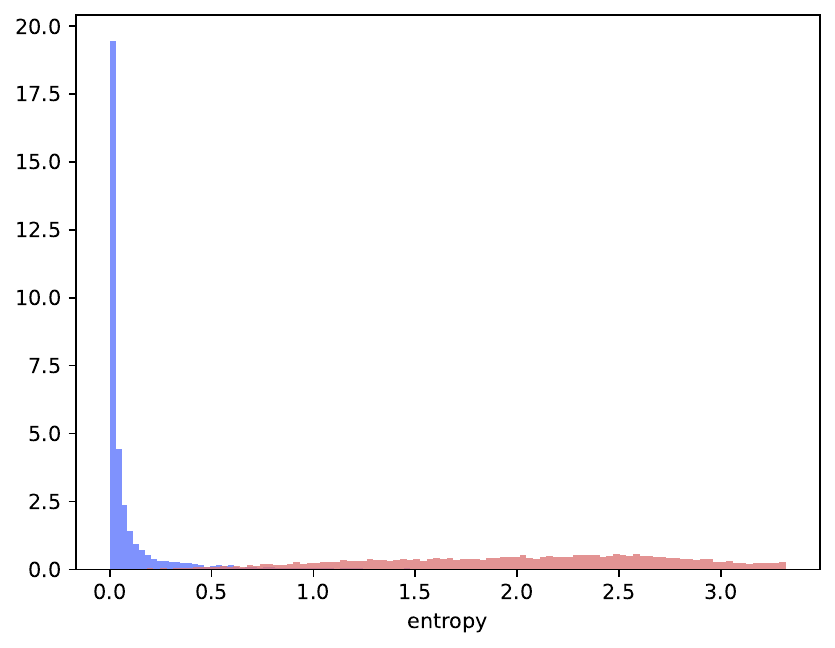}
  \end{subfigure}%
  ~
  \begin{subfigure}[t]{0.2\textwidth}
    \centering
    \includegraphics[width=1\linewidth]{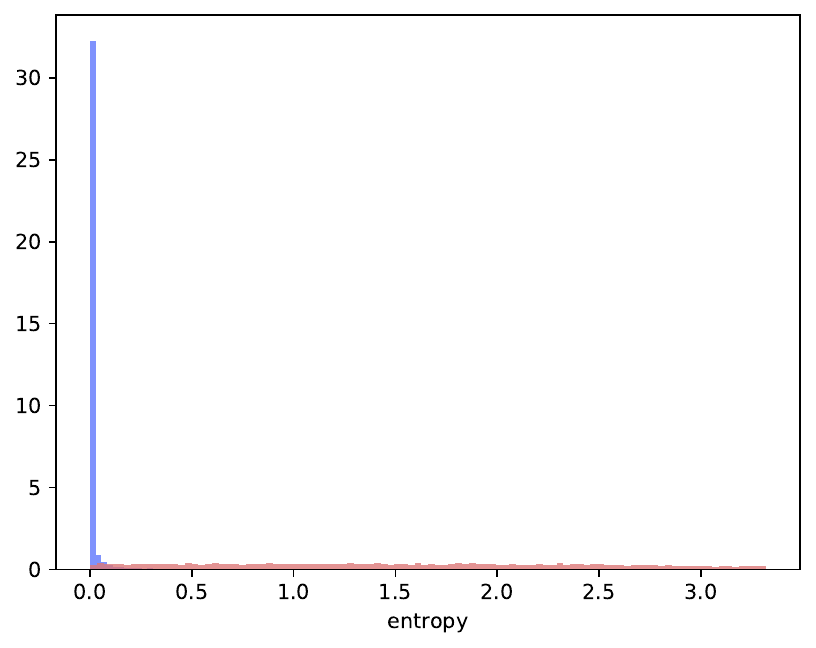}
  \end{subfigure}%
  \\
  \begin{subfigure}[t]{0.2\textwidth}
    \centering
    \includegraphics[width=1\linewidth]{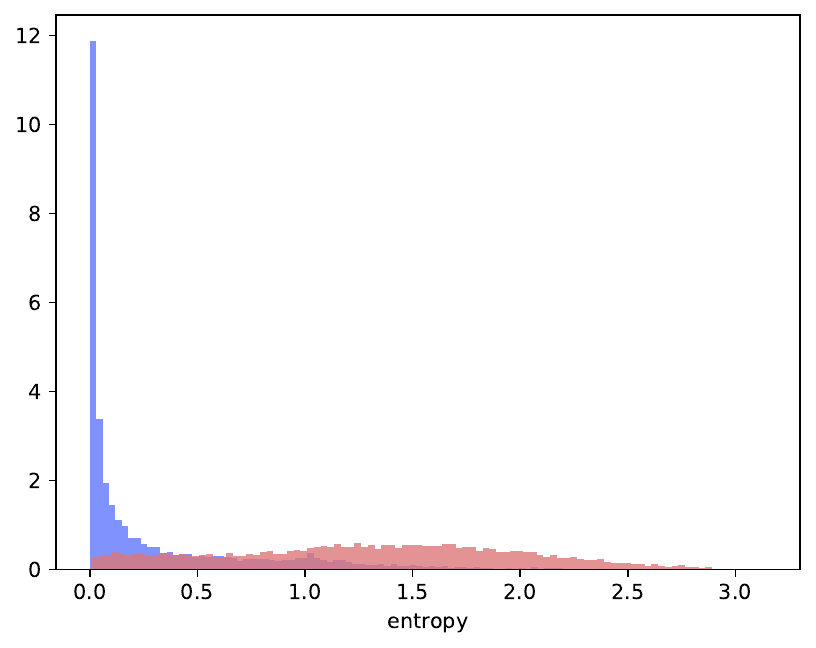}
  \end{subfigure}%
  ~
  \begin{subfigure}[t]{0.2\textwidth}
    \centering
    \includegraphics[width=1\linewidth]{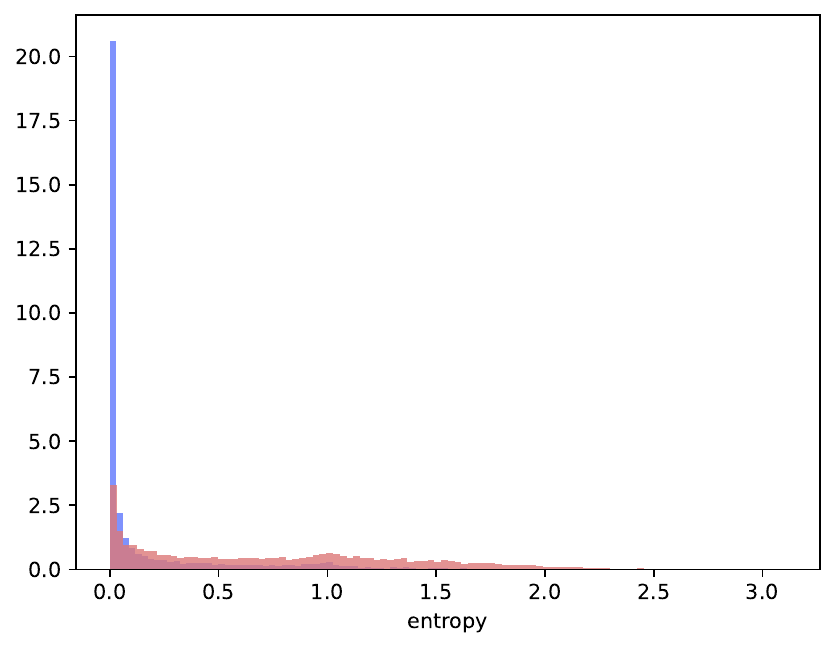}
  \end{subfigure}%
  ~
  \begin{subfigure}[t]{0.2\textwidth}
    \centering
    \includegraphics[width=1\linewidth]{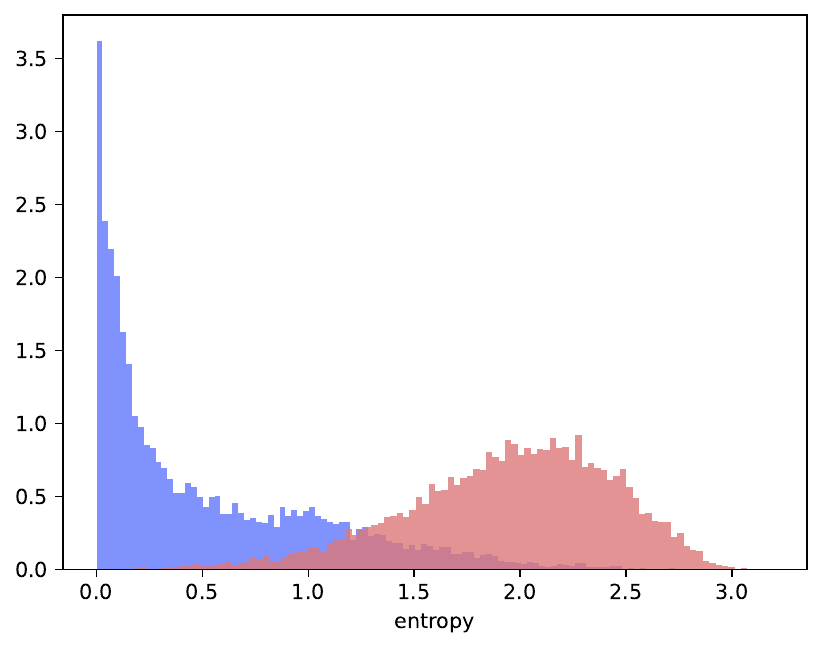}
  \end{subfigure}%
  ~
  \begin{subfigure}[t]{0.2\textwidth}
    \centering
    \includegraphics[width=1\linewidth]{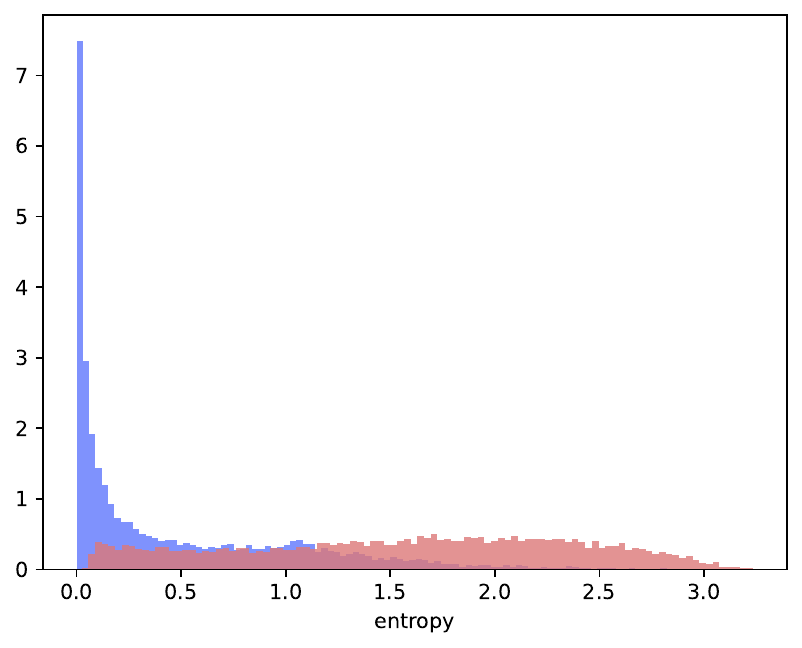}
  \end{subfigure}%
  \\
  \begin{subfigure}[t]{0.2\textwidth}
    \centering
    \includegraphics[width=1\linewidth]{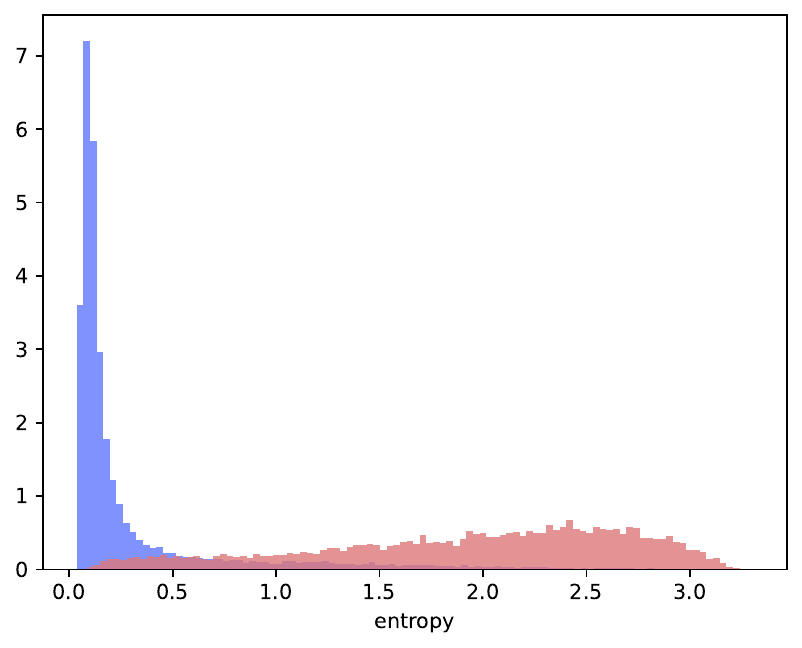}
  \end{subfigure}%
  ~
  \begin{subfigure}[t]{0.2\textwidth}
    \centering
    \includegraphics[width=1\linewidth]{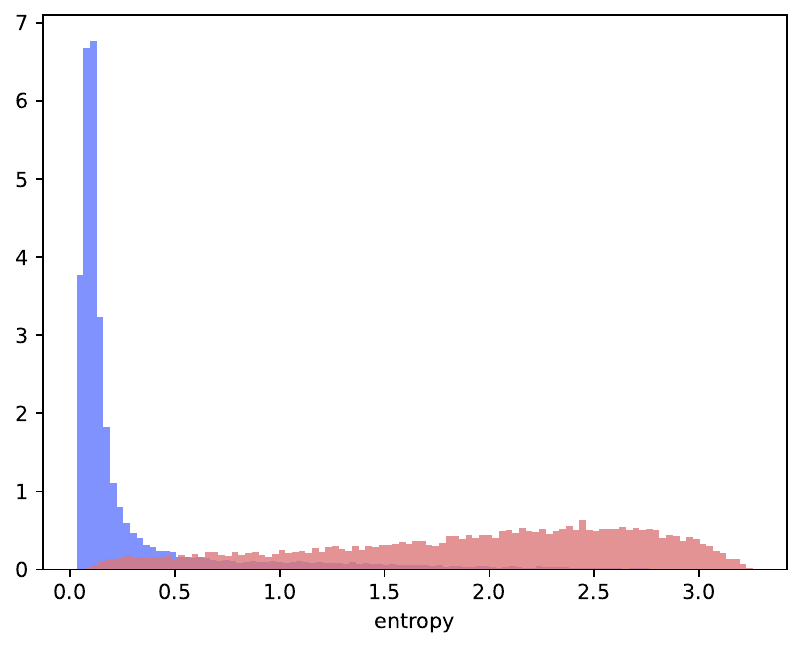}
  \end{subfigure}%
  ~
  \begin{subfigure}[t]{0.2\textwidth}
    \centering
    \includegraphics[width=1\linewidth]{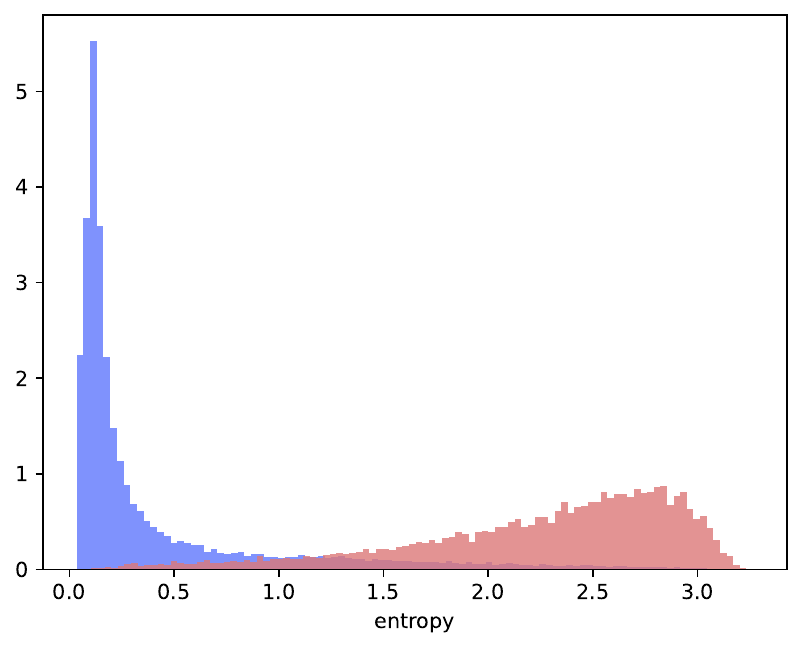}
  \end{subfigure}%
  ~
  \begin{subfigure}[t]{0.2\textwidth}
    \centering
    \includegraphics[width=1\linewidth]{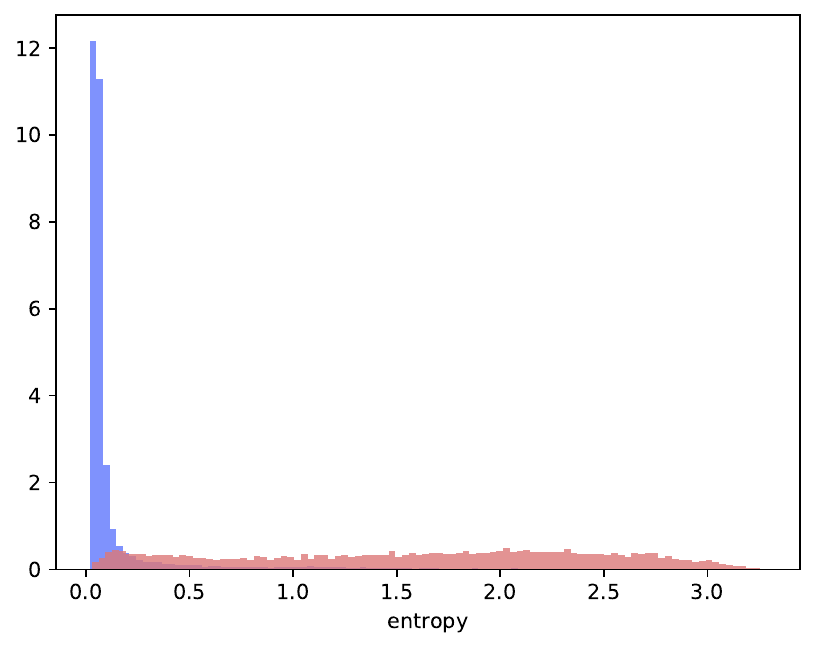}
  \end{subfigure}%
  \\
  \begin{subfigure}[t]{0.2\textwidth}
    \centering
    \includegraphics[width=1\linewidth]{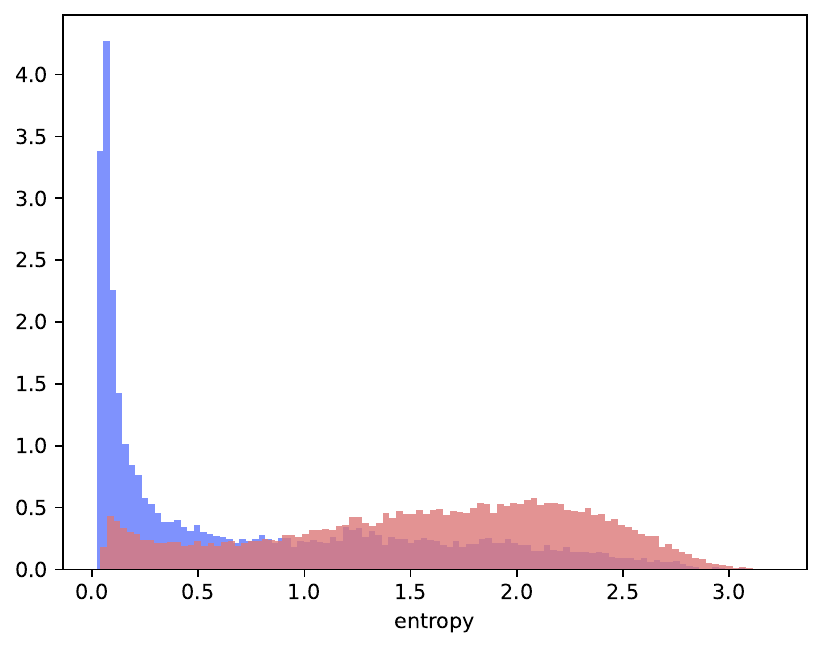}
    \caption{BPS}
  \end{subfigure}
  ~
  \begin{subfigure}[t]{0.2\textwidth}
    \centering
    \includegraphics[width=1\linewidth]{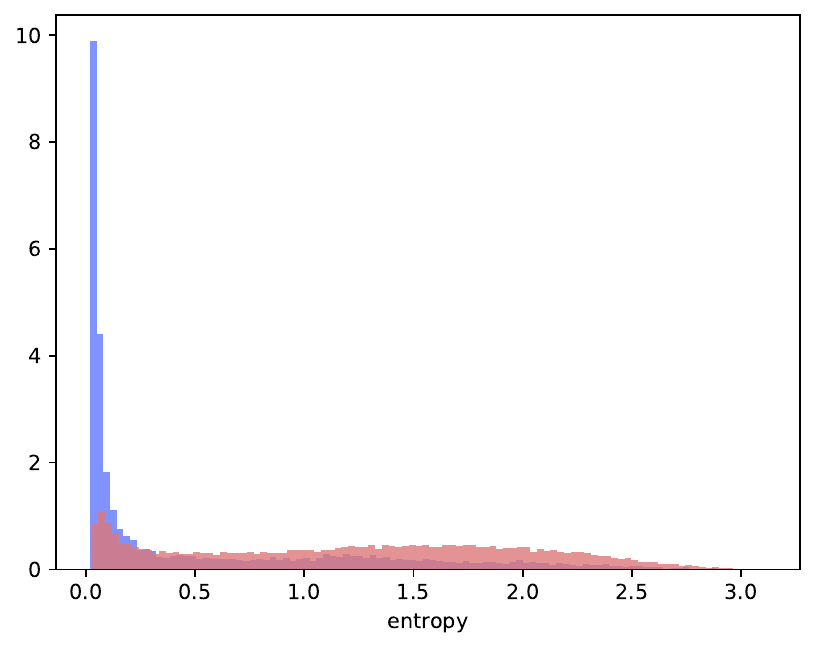}
    \caption{\sigbps}
  \end{subfigure}
  ~
  \begin{subfigure}[t]{0.2\textwidth}
    \centering
    \includegraphics[width=1\linewidth]{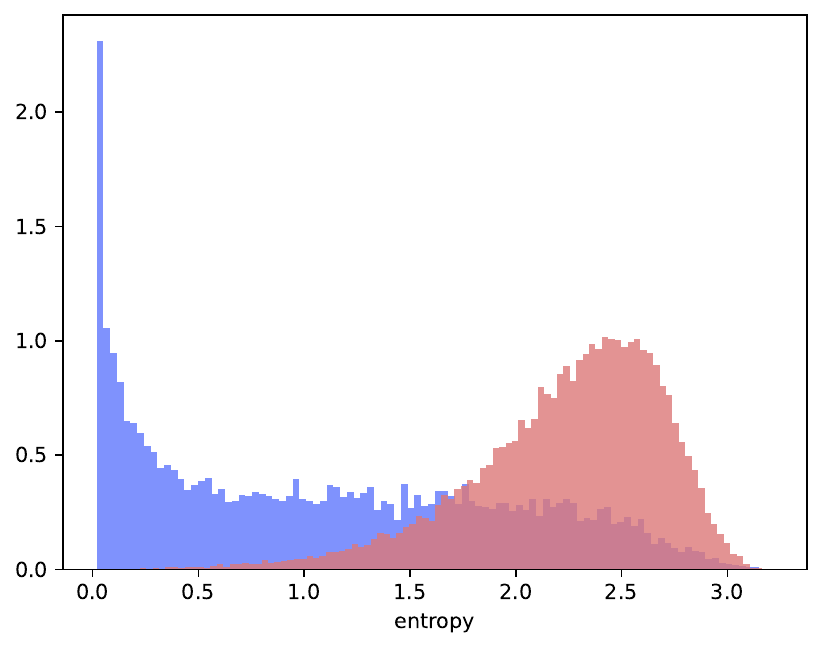}
    \caption{Boomerang}
  \end{subfigure}
  ~
  \begin{subfigure}[t]{0.2\textwidth}
    \centering
    \includegraphics[width=1\linewidth]{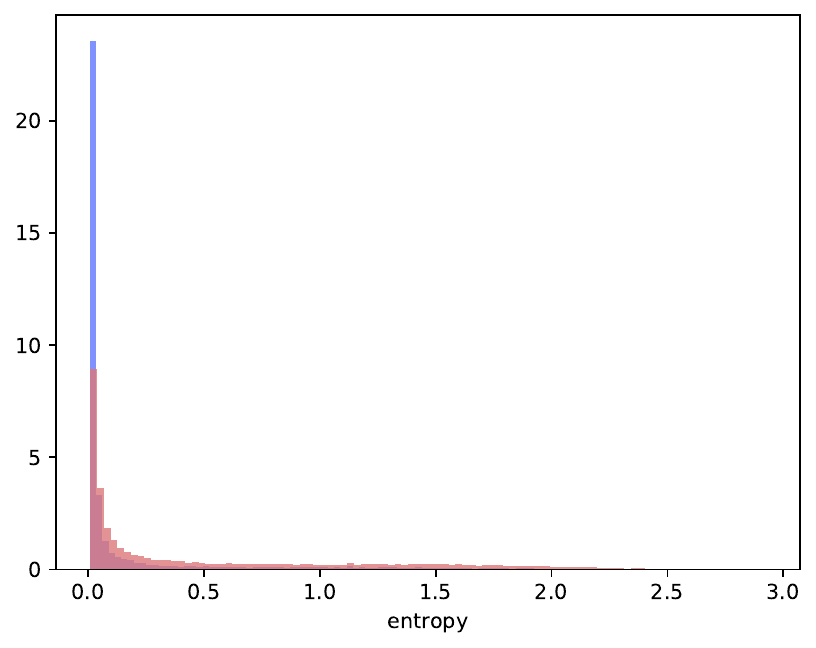}
    \caption{SGLD}
  \end{subfigure}
  \caption{Entropy within the final predictive categorical vector obtained from
    the tested sampling methods for the different datasets used. Blue histograms
    indicate in-data-distribution entropy and red for OOD data. Each column
    represents the predictive entropy for the corresponding labelled sampler and
    each row for a different dataset. From top to bottom, each row is for models
    fit on the MNIST, Fashion MNIST, SVHN, and CIFAR-10 data respectively. MNIST
    and Fashion MNIST datasets are used to model in and OOD datasets for the
    applicable models, and similarly SVHN and CIFAR-10 to model in and OOD for
    respective models.}
  \label{fig:supp_entropy}
\end{figure}

We investigate here the performance of the different sampling methods for in and
out-of-distribution (OOD) data in terms of predictive classification entropy. We
have demonstrated that PDMP sampling methods present meaningful
epistemic uncertainty in predictions. It is important to identify uncertainty in
the final predictions that are made. Within this work, predictions are made by
taking the $\argmax$ of the mean for the predictive posterior,
\begin{equation}
  \label{eq:pred_mean}
  \mathbf{t} = \argmax_{y^{*} \in \mathcal{Y}} \  \mathop{\mathbb{E}_{y^{*}}} \Big[ p(y^* | x^*, \mathcal{D}) \Big]
\end{equation}
Where $p(y^* | x^*, \mathcal{D})$ is our predictive posterior. Entropy within
this categorical probability vector given by this expectation can be viewed as
an approximate measure for aleatoric uncertainty within our model
\cite{smith2018understanding} to accompany the epistemic uncertainty given by
our Bayesian models. To assess this, we compute the entropy of the expectation
within Equation \ref{eq:pred_mean} for in-distribution data and OOD data. It is
desirable to have a lower entropy for in-distribution data indicating lower
predictive aleatoric uncertainty, and a larger entropy for OOD data to represent
an increase in uncertainty. Figure \ref{fig:supp_entropy} illustrates this for
the models used within this work.
From Figure \ref{fig:supp_entropy}, we see the SGLD provides a reduced entropy
for in-distribution data, but provides reduced levels of entropy for OOD data
when compared with PDMP samplers. This highlights a favourable property for
results from PDMP applications for neural networks for detection and
communication of additional uncertainty for OOD data.
\section{Examples of Difficult to Classify Samples}
\begin{figure}[!h]
  \centering
  \begin{subfigure}[t]{0.4\textwidth}
    \centering \includegraphics[width=1\linewidth]{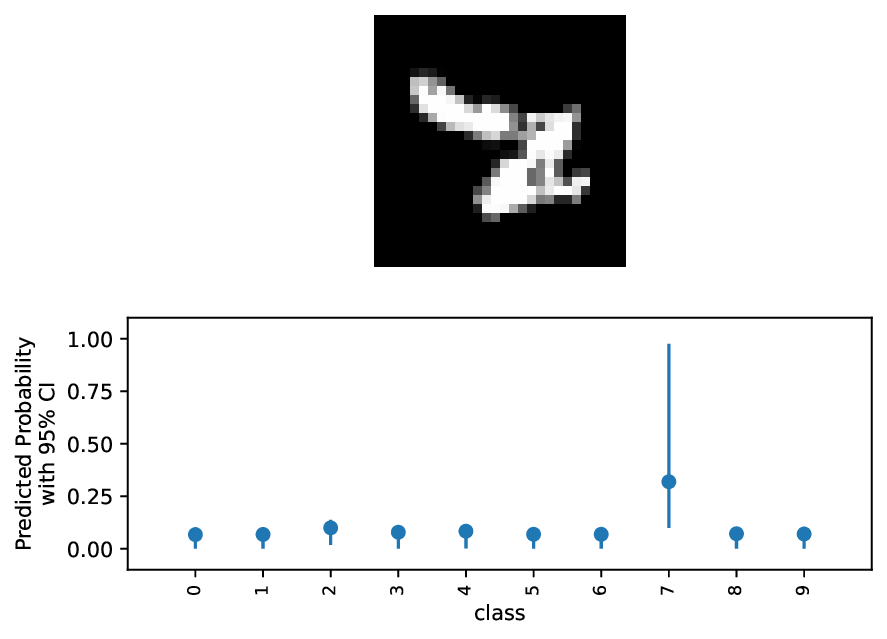}
    \caption{`Two` from MNIST misclassified as `Seven`}
  \end{subfigure}%
  ~
  \begin{subfigure}[t]{0.4\textwidth}
    \centering
    \includegraphics[width=1\linewidth]{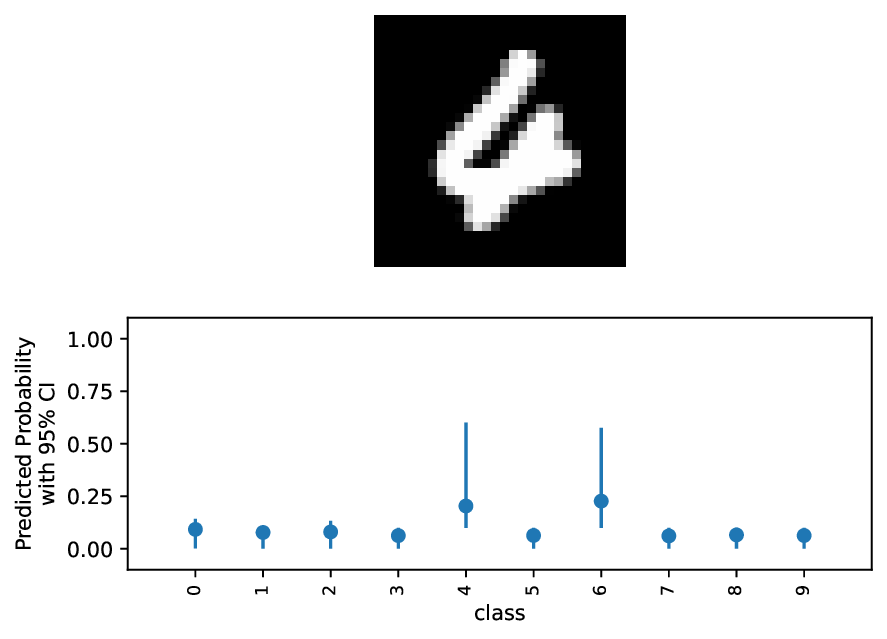}
    \caption{`Four` from MNIST missclassified as `Six`}
  \end{subfigure}
  \\
  \begin{subfigure}[t]{0.4\textwidth}
    \centering
    \includegraphics[width=1\linewidth]{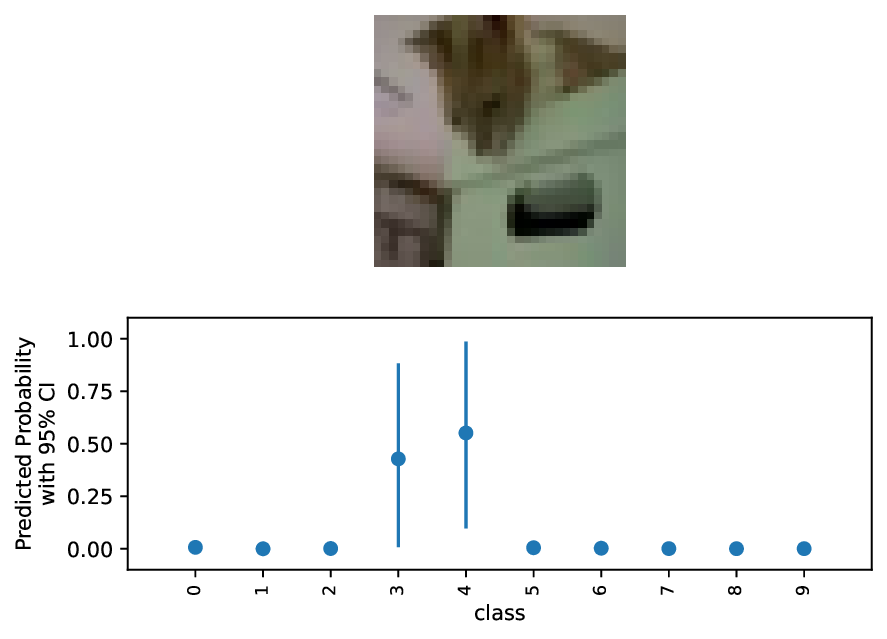}
    \caption{`Cat` from CIFAR-10 misclassified as `Deer`}
  \end{subfigure}%
  ~
  \begin{subfigure}[t]{0.4\textwidth}
    \centering
    \includegraphics[width=1\linewidth]{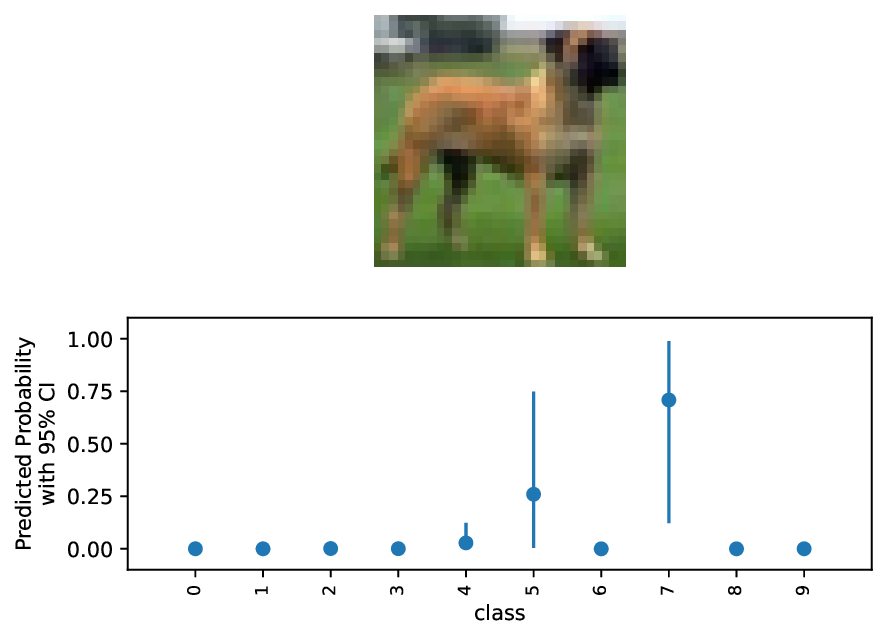}
    \caption{`Dog` from CIFAR-10 misclassified as `Horse`}
  \end{subfigure}
  \\
  \begin{subfigure}[t]{0.4\textwidth}
    \centering
    \includegraphics[width=1\linewidth]{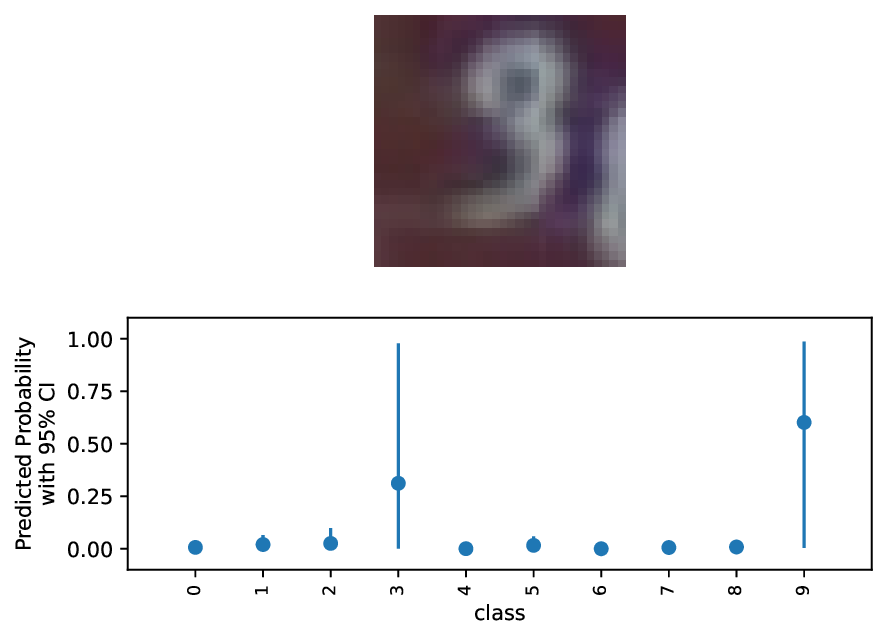}
    \caption{`Three` from SVHN missclassified as `Nine`}
  \end{subfigure}%
  ~
  \begin{subfigure}[t]{0.4\textwidth}
    \centering
    \includegraphics[width=1\linewidth]{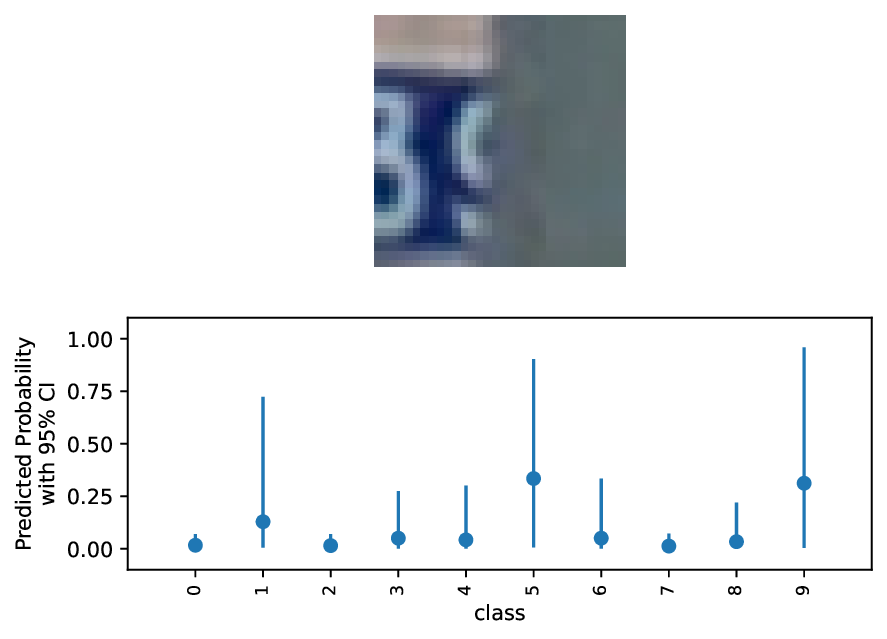}
    \caption{`Nine` from SVHN missclassified as `five`}
  \end{subfigure}
  \caption{Examples of difficult-to-classify images from the different image
    data sets used. Below each image is the predictive mean for each class
    represented by the dot, and error bars to represent the 95\% credible
    intervals. MNIST results fit with BPS, CIFAR-10 with \sigbps, and SVHN with
    Boomerang sampler using the proposed event thinning method. Best viewed on a
    computer screen.}
  \label{fig:miss}
\end{figure}

Given the increasing desire to apply deep learning models in practice, the
ability for them to reliably communicate uncertainty information is crucial. We
expect our models to encounter difficult-to-understand scenarios. We need to
be able to identify when these challenging scenarios occur and to incorporate the
uncertainty encountered into final decisions. Figure \ref{fig:miss} illustrates
examples of misclassified samples from the datasets evaluated within this work,
and illustrates the predictive probabilities of these models. We see that all
PDMP samplers provide meaningful uncertainty information for
difficult-to-classify instances within each data set.

\clearpage
\bibliography{ref.bib}